%% file: main.tex
\documentclass[runningheads]{llncs}

\usepackage{eccv}

\usepackage{eccvabbrv}

\usepackage{fontawesome}
\usepackage{graphicx}
\usepackage{booktabs}
\usepackage{graphicx}
\usepackage{subcaption}
\usepackage{booktabs}
\usepackage{xcolor}
\usepackage{wrapfig}
\usepackage{multirow}
\usepackage{tabularx}
\usepackage{array}
\usepackage{authblk}
\usepackage{amsmath}
\usepackage{bbm}
\usepackage{booktabs}
\usepackage{multirow}
\usepackage{pifont}
\usepackage{amssymb}

\newcommand{\cmark}{\ding{51}}
\newcommand{\xmark}{\ding{55}}
\usepackage[accsupp]{axessibility}  

\usepackage{hyperref}

\usepackage{orcidlink}

\usepackage[normalem]{ulem}
\usepackage{algorithm}
\usepackage{algpseudocode} 
\usepackage[table]{xcolor}  
\begin{document}


\newcommand{\methodfullname}{{\ttfamily\scshape Preservation} {\ttfamily\upshape aware} {\ttfamily\scshape Adaptive} {\ttfamily\scshape Ranked} {\ttfamily\scshape Subspace} {\ttfamily\scshape Expansion} {\ttfamily\scshape (PARSE)}}
\newcommand{\method}{{\ttfamily\scshape PARSE}}
\newcommand{\evalmetricfullname}{{\ttfamily\scshape Balanced} {\ttfamily\scshape Erasure} {\ttfamily\scshape Utility} {\ttfamily\scshape Score} {\ttfamily\scshape (BEUS)}}
\newcommand{\evalmetric}{{\ttfamily\scshape BEUS}}


\newcommand{\BGOPA}{50} 
\definecolor{nsfwcol}{HTML}{0F766E}   
\definecolor{stylecol}{HTML}{14B8A6}   
\definecolor{objcol}{HTML}{2DD4BF}   
\definecolor{utilitygreen}{HTML}{306223}

\newcommand{\backgroundnsfwcol}[1]{\cellcolor{nsfwcol!50!white}{#1}}   
\newcommand{\backgroundstylecol}[1]{\cellcolor{stylecol!40!white}{#1}} 
\newcommand{\backgroundobjcol}[1]{\cellcolor{objcol!25!white}{#1}}     
\newcommand{\utilitycol}[1]{\textcolor{utilitygreen}{#1}}     
\newcommand{\tbackgroundnsfwcol}[1]{\textcolor{nsfwcol!100!white}{#1}}   
\newcommand{\tbackgroundstylecol}[1]{\textcolor{stylecol!100!white}{#1}} 
\newcommand{\tbackgroundobjcol}[1]{\textcolor{objcol!100!white}{#1}}     

\title{\faBalanceScale{} To Erase, or Not to Erase: Robust Training-Free Concept Erasure with Preservation aware Adaptive Ranked Subspace Expansion} 

\titlerunning{PARSE}

\author{Shaswati Saha\inst{1}\orcidlink{0000-0002-3995-7235} \and
Rajasekhar Anguluri\inst{1}\orcidlink{0000-0003-2537-2778} \and
Manas Gaur\inst{1}\orcidlink{0000-0002-5411-2230}}

\authorrunning{S. Saha, R. Anguluri, M. Gaur}

\institute{University of Maryland Baltimore County, USA\\
\email{\{ssaha3, rajangul, manas\}@umbc.edu}}

\maketitle
\begin{figure}[ht]
  \centering
  \includegraphics[width=\linewidth]{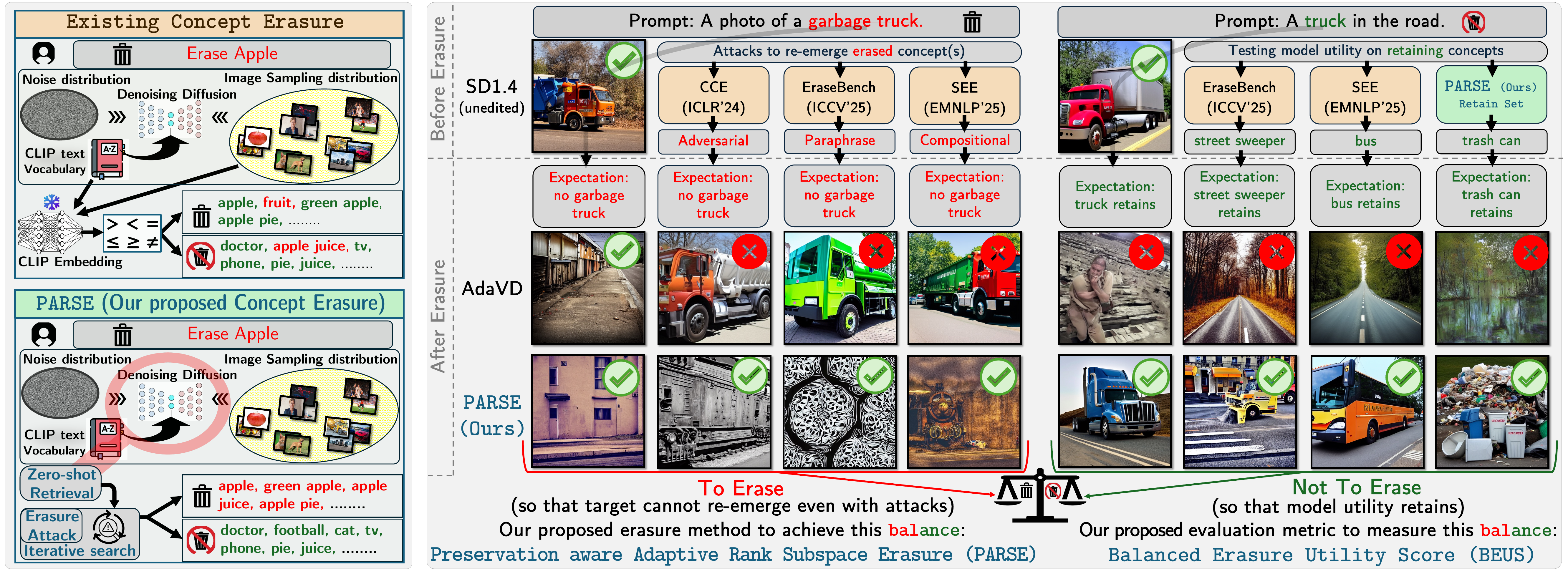}
\caption{
  Existing CETs often trade erasure robustness for utility: they either fail to prevent target re-emergence under attacks or over-suppress and harm benign/adjacent concepts.
  \method{} reframes \emph{what to erase vs.\ what to retain} as a \emph{diffusion-grounded retrieval problem}: given a target, we retrieve target-adjacent concepts by directly querying text-conditioned denoising/steering guidance to build diffusion-grounded erase/retain banks.
  We then apply preservation-aware, training-free model editing and adaptively expand the erased subspace to minimize re-emergence while preserving utility, and evaluate this balance \faBalanceScale{} with \evalmetric{}.
  \textcolor{red}{\texttt{\small {Warning:} This paper contains NSFW contents.}}
}
\label{fig:teaser}
\end{figure}

\begin{abstract} 
Concept erasure techniques (CETs) edit text-to-image diffusion models to erase undesired targets (e.g., NSFW content or copyrighted styles) while preserving model utility on image generation for benign concepts. 
Current CETs often face a trade-off between erasure robustness and model utility: stronger edits erase target more reliably, but degrade model utility on non-target concepts and vice-versa.
A key source of this trade-off is how existing methods define what to erase and what to preserve. 
Many CETs rely on static concept banks, where erase or retain concepts are manually specified, generated by Large Language Models (LLMs), or selected using CLIP-based image--text similarity. 
However, such static banks do not explicitly model how prompts steer the diffusion model during denoising, leaving edited models vulnerable to triggers that can reintroduce the target while unintentionally suppressing nearby benign concepts.
We present \methodfullname{}, a training-free framework for robust concept erasure in latent diffusion models.
Given a target, \method{} queries the diffusion model using classifier-free guidance to dynamically discover target-inducing erase concepts and nearby retain concepts from the model vocabulary.
\method{} then edits the cross-attention value space with a preservation-aware projection that removes target directions while leaving retain directions intact.
To address triggers beyond the vocabulary-indexed search space, \method{} iteratively searches for re-emergence triggers using textual inversion and adaptively expands the erased subspace only when a new trigger direction does not conflict with retain semantics.
We further introduce \evalmetricfullname{}, a balanced erasure utility score that combines robustness (ASR under multiple attacks) and utility preservation (FID) via bounded monotone transforms and harmonic mean aggregation. 
Extensive experiments across NSFW, artistic style, and object erasure, together with a large-scale robustness utility analysis over many CET baselines, show that \method{} achieves robust erasure across multiple concepts without sacrificing post-edit model utility. {Project page: {\url{https://shaswati1.github.io/parse/}}}
\keywords{Machine Unlearning \and Concept Erasure \and Responsible and Safe Image Generation \and Reliable T2I Diffusion Models \and Responsible AI}
\end{abstract}

\input{sections/introduction}

\input{sections/prelim}
\input{sections/method}

\input{sections/eval_metric}

\input{sections/experiments}

\input{sections/conclusion}
\input{sections/limitations}

\section*{Acknowledgements}
We thank ECCV reviewers for their constructive feedback. This work was supported in part by a gift award from NeuralNest LLC and USISTEF: US-India Technology Endowment Fund. All the opinions are those of the authors and not of the sponsoring organizations.

\clearpage
\bibliographystyle{splncs04}
\bibliography{main}

\clearpage
\appendix
\section*{Appendix}
\input{appendix/appendix}

\end{document}

%% file: sections/introduction.tex
\section{Introduction}
Text-to-image diffusion models are widely deployed for image generation from text prompts \cite{lu2022dpm, rombach2022high}, yet their ability to faithfully reproduce undesirable concepts (e.g., NSFW content or copyrighted styles) raises pressing safety and compliance concerns~\cite{saha2023rfc, ravin2022mitigating, saha2022pairwise, saha2018total, saha2018efficient, kamran2019optic, saha2018lightning, saha2025improving, saha2023seebel}. 
Concept Erasure Techniques (CETs), a form of \emph{machine unlearning} \cite{joshi2024towards, gupta2021adaptive, ullah2023adaptive, cha2024learning, tarun2023fast} address this by editing a pretrained model to suppress or erase user-specified targets while preserving model utility to generate benign non-target concepts. 
In practice, concept erasure demands a careful \faBalanceScale{} balance: robust erasure often degrades model utility \cite{saha-etal-2025-side, joshi2024towards}, whereas with inferior erasure \cite{tarun2023fast, gandikota2024unified}, the target can still re-emerge with semantically similar \cite{saha-etal-2025-side, thakral2025fine}, paraphrased \cite{amara2025erasing}, or adversarial prompts \cite{zhang2024defensive} (\cref{fig:teaser}). 

First, many CETs \footnote{ Detailed literature review and related works are provided in supplementary material. } rely on static target concept sets \cite{gandikota2024unified,kumari2023ablating,gandikota2023erasing} either pre-defined by humans, LLMs or retrieved using image-text similarity in proxy CLIP \cite{radford2021learning} embedding space.
Since these sets are curated without explicitly modeling how prompts steer the denoising trajectory \cite{luo2022understanding, rombach2022high, zhang2023adding} in diffusion models, they fail to capture diffusion-grounded triggers \cite{gandikota2024unified,kumari2023ablating,gandikota2023erasing} and leave the erased model vulnerable to re-emergence attacks \cite{pham2023circumventing}.
Second, adversarially robust CETs \cite{zhang2024defensive,huang2024receler,kim2024race} suppress often over-suppress and inadvertently erase semantically adjacent concepts (e.g., removing ``garbage truck'' also suppresses ``tow truck''), compromising generation quality and diversity \cite{amara2025erasing, thakral2025genm}. 
Finally, existing CETs typically report robustness and utility separately \cite{ gong2024reliable, wang2025precise, srivatsan2025stereo}, making it difficult to evaluate whether a CET truly achieves an effective erasure-utility \faBalanceScale{} balance.

We introduce \method{}, a preservation-aware and trigger-adaptive concept erasure framework for latent diffusion models.
Unlike existing methods that rely on predefined or proxy concept pools \cite{bui2025fantastic, lee2025localized, li2025speed}, project target directions without a target-conditioned retain bank \cite{wang2025precise}, or
finetune on discovered adversarial prompts without enforcing a retain-orthogonal residual
expansion \cite{srivatsan2025stereo, gong2024reliable, kim2024race}, our method
(i) dynamically discovers target-adjacent concepts by querying the diffusion model using
classifier-free guidance \cite{ho2021classifier} rather than proxy embeddings,
(ii) replaces human/LLM-defined anchor sets with diffusion-vocabulary knowledge-bank
retrieval (zero-shot), treating the target as a query to retrieve target-adjacent tokens
that define erase/retain banks, (iii) performs preservation-aware value-subspace projection
that explicitly protects retain semantics while removing target directions, and
(iv) adaptively expands the erased subspace by iteratively discovering triggers that can reintroduce the target, while preserving model utility.
Therefore, the key idea of \method{} is an end-to-end zero-shot method that defines what to erase, what to retain, and when to expand, using priors from the diffusion model.
Together, these design principles yield robust concept erasure for latent diffusion models across diverse concept categories without sacrificing post-edit model utility.
Our contributions are threefold:
\begin{itemize}
\item We propose \method{}, a robust training-free concept erasure framework that unifies diffusion-grounded ranked subspace construction, preservation-aware projection, and trigger-guided adaptive subspace expansion.
    \item We introduce \evalmetric{}, an erasure utility balance score that jointly measures erasure robustness (ASR) and model utility preservation (FID) via bounded monotone transforms and harmonic-mean aggregation. 
    \item We conduct extensive experiments across NSFW, artistic style, and object erasure under multiple attacks and datasets, including a large-scale robustness-utility balance analysis over many CET baselines with \evalmetric{}, showing that \method{} improves erasure robustness, preserving model utility. 
\end{itemize}

%% file: sections/prelim.tex
\section{Background} 
\label{sec:prelime}
The goal of concept erasure in latent diffusion models is to prohibit image generation of undesired \texttt{\normalsize target} concept(s) specified by the user {\small(e.g.,
\emph{nudity})}.
In this section, we discuss how text-to-image latent diffusion models generate images from text prompts, and how value-space projection in diffusion cross-attention layers can be used for Concept Erasure. 

\subsection{Text-to-Image Latent Diffusion Models}
\label{sec:prelim_ldm}
Text-to-Image Latent Diffusion Models (LDMs)~\cite{rombach2022high} generate \texttt{\normalsize images} from text \texttt{\normalsize prompts} by performing \emph{noise-to-image} diffusion in the low-dimensional latent space of a pretrained Variational Auto Encoder (VAE).
The VAE $\mathrm{\textit{\footnotesize Encoder}}$ maps an \texttt{\normalsize image} to a latent $z{=}\mathrm{\textit{\footnotesize Encoder}}(\texttt{\footnotesize image})$, and the VAE $\mathrm{\textit{\footnotesize Decoder}}$ reconstructs image using $\mathrm{\textit{\footnotesize Decoder}}(z)$.
For generation, LDMs sample Gaussian noise $z_T {\sim} \mathcal{N}(0,I)$ and iteratively denoise $z_t$ over timesteps $t {\in} \{T,\dots,1\}$ with a UNet denoiser $\epsilon_\theta$ to obtain a clean latent $z_0$, which is decoded as $\mathrm{\textit{\footnotesize Decoder}}(z_0)$. 
LDMs condition denoising on the \texttt{\normalsize prompt} via cross-attention (CA) layers inside the UNet: queries $Q$ are computed from noisy image latents, while keys $K$ and values $V$ are computed from prompt token embeddings.
This yields an attention map $\mathcal{A} {=} \mathrm{\textit{\footnotesize softmax}} ( QK^\top {/} \sqrt{d} )$, which selects relevant tokens per spatial location, and the weighted sum of $\mathcal{A}$ and $V$ injects token information back into UNet features (e.g., via an MLP).
Repeating this conditioning across timesteps steers denoising toward a noise-free, \texttt{\normalsize prompt}-aligned \texttt{\normalsize image}. 
Additionally, during inference, classifier-free guidance (CFG) \cite{ho2021classifier, shen2024rethinking, shenoy2026gradient} amplifies \texttt{\normalsize prompt}-aligned generation by combining unconditional denoisers ($\epsilon_u$) and conditional denoisers ($\epsilon_c$):
\begin{equation}
    \epsilon_u(z_t,t) {=} \epsilon_\theta(z_t,t,\varnothing),
    \qquad
    \epsilon_c(z_t,t;\texttt{\footnotesize prompt}) {=} \epsilon_\theta(z_t,t;\texttt{\footnotesize prompt}).
    \label{eq:conditional_and_unconditional_denoiser_original}
\end{equation}
CFG then forms a guided noise prediction by moving from $\epsilon_u$ in the \texttt{\normalsize prompt}-induced direction $(\epsilon_c{-}\epsilon_u)$ with guidance scale $s {\geq} 1$:
\begin{equation}
  \epsilon_{\mathrm{cfg}}(z_t,t;\texttt{\footnotesize prompt})=\epsilon_u(z_t,t)+s\big(\epsilon_c(z_t,t;\texttt{\footnotesize prompt})-\epsilon_u(z_t,t)\big).
  \label{eq:cfg_original}
\end{equation}

\subsection{Value-Space Projection for Concept Erasure}
\label{sec:prelim_valuespaceproj}
From CA layer discussion, the value matrix $V$ determines what \texttt{\normalsize prompt} information is injected into UNet features $\theta$ during generation.
Thus, concept erasure can be implemented as a low-rank linear edit on values: $V^\texttt{\scriptsize edited} {=} PV^\texttt{\scriptsize prompt}$, where $P$ removes directions induced by the \texttt{\normalsize target prompt}.
For a \texttt{\normalsize prompt}, let $V^\texttt{\scriptsize prompt}$ denote the value matrix with token-wise columns $\mathrm{v}_{\,j}$.
For a \texttt{\normalsize target prompt}, let $V^\texttt{\scriptsize target prompt}$ denote the corresponding value matrix with columns $\mathrm{v}^{\,t}_{\,j}$.
ADaVD~\cite{wang2025precise} performs token-wise erasure by projecting each $\mathrm{v}_{\,j}$ onto the orthogonal complement of the span of $\mathrm{v}^{\,t}_{\,j}$, i.e., $\mathrm{v}^r_{\,j} {=} \bigl(I {-} P\bigr)\,\mathrm{v}_{\,j}$.
Stacking $\{\mathrm{v}^r_{\,j}\}$ column-wise yields $V^\texttt{\scriptsize edited}$, used for erasure via $V^\texttt{\scriptsize edited} {=} PV^\texttt{\scriptsize prompt}$.

Now, the rank-$1$ projector onto the target span is -
\begin{equation}
    P_{\texttt{\tiny span} (\texttt{\tiny tokens} {\in} V^\texttt{\tiny target prompt})}
    \;=\;
    {\mathrm{v}^t_{\,j}(\mathrm{v}^t_{\,j})^\top} / {(\mathrm{v}^t_{\,j})^\top \mathrm{v}^t_{\,j}}
    \;=\;
    {\mathrm{v}^t_{\,j}(\mathrm{v}^t_{\,j})^\top} / {\|\mathrm{v}^t_{\,j}\|_2^2}.
    \label{eq:rank1_adavd_projector}
\end{equation}
From unit norm definition $e_j {=} {\mathrm{v}^t_{\,j}}/{\|\mathrm{v}^t_{\,j}\|_2}$, we have
$P_{\texttt{\tiny span} (\texttt{\tiny tokens} {\in} V^\texttt{\tiny target prompt})} {=} e_j e_j^\top$.
Letting $\mathcal{E}_j {=} [e_j]$, the ADaVD projector can be written as -
\begin{equation}
    P_{\texttt{\tiny span} \perp (\texttt{\tiny tokens} {\in} V^\texttt{\tiny target prompt})}
    \;{=}\;
    I {-} P_{\texttt{\tiny span} (\texttt{\tiny tokens} {\in} V^\texttt{\tiny target prompt})}
    \;{=}\;
    I - e_j e_j^\top
    \;=\; 
    I {-} \mathcal{E}_j \mathcal{E}_j^\top.
    \label{eq:adavd_projector}
\end{equation}

%% file: sections/method.tex
\section{Preservation aware Adaptive Ranked Subspace Expansion \textcolor{blue}{\texttt{(\method{})}} for balanced \faBalanceScale{} Concept Erasure}
\label{sec:method}
\begin{figure}[ht]
  \centering
  \includegraphics[width=\linewidth]{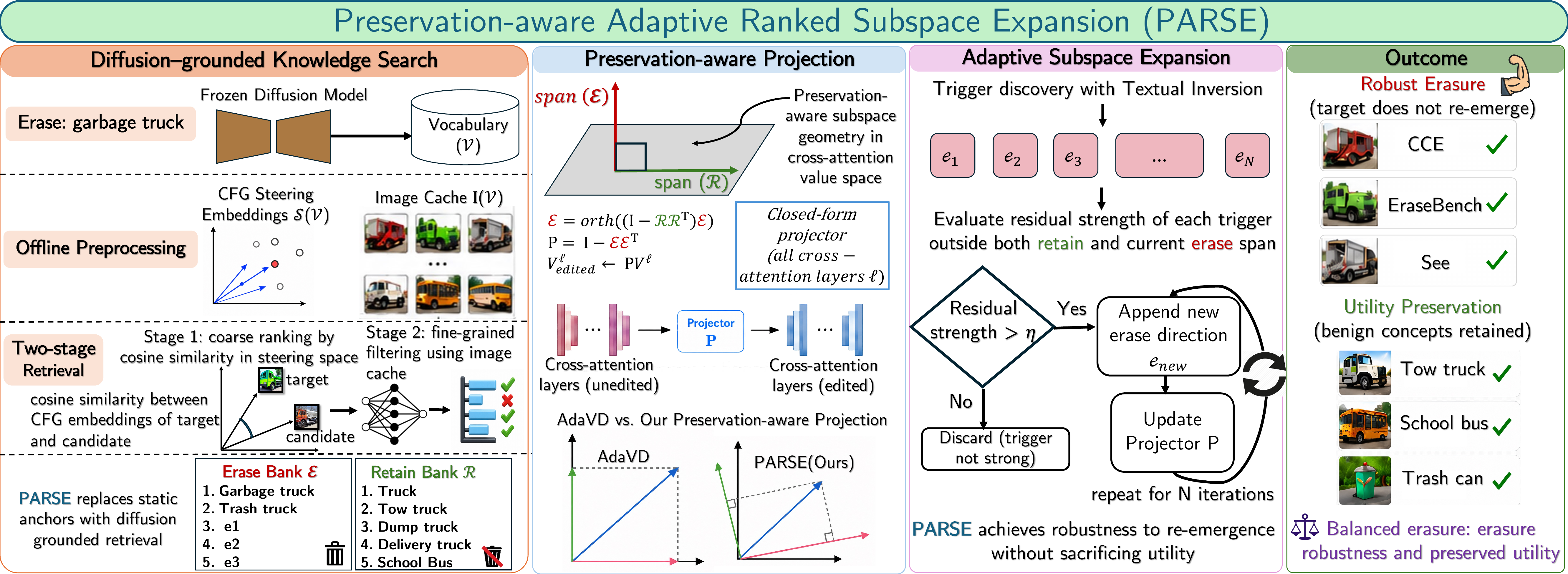}
  \caption{\method{} builds diffusion-grounded erase and retain banks via two-stage Knowledge Search, applies preservation-aware value-space projection to remove target directions while preserving retain concepts, and uses textual-inversion triggers to adaptively expand the erased subspace for stronger robustness without sacrificing utility.}
  \label{fig:method}
\end{figure}

Given a latent diffusion model $M_0$ and \texttt{\normalsize target} concept(s), CET returns an edited model $M_\texttt{\normalsize target}$ that suppresses \texttt{\normalsize target} generation while preserving utility on non-target concepts.
To balance erasure and utility, \method{} has three components:
(i) {Ranked Subspace (\textsc{Knowledge Search})},
(ii) {Preservation-aware Subspace Projection}, and
(iii) {Adaptive Subspace Expansion}.
We describe each component below and how they compose into \method{} (\cref{fig:method}).

\subsection{Ranked Subspace for Concept Erasure: \textsc{Knowledge Search}}
\label{sec:utility_discovery}
A core design choice for balanced erasure is defining a concept bank to \emph{erase} and a concept bank to \emph{retain}.
Prior CETs~\cite{gandikota2024unified,kumari2023ablating} often use limited, static anchor sets (human/LLM-curated) and proxy similarity (e.g., CLIP~\cite{radford2021learning}), which limits coverage and admit triggers for post-erasure re-emergence.
We instead: (i) search over the frozen text-encoder vocabulary $\mathcal{V}$, (ii) build the erase bank \emph{dynamically} from a user target by treating the target as a retrieval query over $\mathcal{V}$, and (iii) rank tokens using diffusion-grounded CFG, specifically the prompt-guided denoising direction $(\epsilon_c-\epsilon_u)$, rather than proxy embeddings.

We construct an erase bank by retrieving tokens that best represent the erasure \texttt{\normalsize target}; the retain bank is its complement.
Since $\mathcal{V}$ is the diffusion model's ``knowledge bank'' for text-to-image generation, we refer to this retrieval-based bank construction as \textsc{Knowledge Search}.
Our key novelty is a zero-shot two-stage retrieval pipeline with offline indexing \cite{saha2026zero}, where tokens are ranked by how similarly they steer UNet denoising compared to the \texttt{\normalsize target}.

\paragraph{\textbf{Offline Vocabulary Preprocessing - Computing Tokenwise CFG Steering Embeddings and Image Cache:}}
\label{sec:offline}
From \cref{eq:cfg_original}, $(\epsilon_c {-} \epsilon_u)$ is a vector field that steers noisy latents toward a prompt-aligned image.
For a \texttt{\normalsize prompt}, we write its CFG steering field as $\Delta \epsilon (\texttt{\normalsize prompt}) {=} \epsilon_c {-} \epsilon_u$.
Let $\mathcal{V}$ denote the frozen text-encoder vocabulary.
For each token $v {\in} \mathcal{V}$, we form a templated \texttt{\normalsize prompt} containing $v$ (e.g., \texttt{\normalsize ``a photo of $v$''}) and compute its steering field $\Delta \epsilon (v)$ at timestep $t$.
At UNet layer $\ell {\in} \{1 \dots \mathcal{L}\}$, $\Delta \epsilon (v^\ell) {\in} \mathbb{R}^{c^\ell {\times} h^\ell {\times} w^\ell}$, where $c^\ell$ is channel dimension and $h^\ell{\times}w^\ell$ is spatial size.
Applying spatial RMS pooling yields a token embedding at layer $\ell$:
$\mathcal{S}^\ell (v) {=} \mathrm{RMS}_{h^\ell, w^\ell} (\Delta \epsilon (v^\ell)) \in \mathbb{R}^{c^\ell}$.
We concatenate layer-wise vectors and average over seeds and timesteps to obtain a net steering embedding $\mathcal{S}(v) ~\forall v{\in}\mathcal{V}$:

\begin{equation}
    \mathcal{S} (v)
    \;=\;
    \underset{\texttt{\scriptsize seeds} (s), \texttt{\scriptsize steps} (t) }{\mathbb{E}} \Big[ \mathrm{Normalize} \Big( \big[ \mathcal{S}^1 (v) ; \mathcal{S}^2 (v); \dots ; \mathcal{S}^\mathcal{L} (v) \big] \Big) \Big] \in \mathbb{R}^{\sum_{\ell {\in} \mathcal{L}} c^\ell}
    \label{eq:steering_vector}
\end{equation}
Treating $\mathcal{V}$ as a database, we store $\mathcal{S}(v)$ for each token $v$.
While Eq.~(\ref{eq:steering_vector}) is defined per vocabulary token, the retrieval query
is not restricted to one: both multi-token targets and candidates can be encoded as full-prompt CFG steering embeddings via the same construction; therefore, our Knowledge Search applies to multi-word concepts.
Additionally, for each $v{\in}\mathcal{V}$ we generate and cache $n_\mathcal{I}$ images $\{\mathcal{I} (v)_i \}_{i=1}^{n_\mathcal{I}}$ using the same prompt template as above with $n_\mathcal{I}$ different noise seeds.

\paragraph{\textbf{Stage 1 - Coarse Ranking by CFG Steering Embeddings:}}
\label{sec:stage1}
Given an erasure \texttt{\normalsize target}, we compute its steering embedding $\mathcal{S}(\texttt{\normalsize target})$ via \cref{eq:steering_vector}, then score every token $v{\in}\mathcal{V}$ by cosine similarity:

\begin{equation}
    \texttt{\normalsize cosine}(v, \texttt{\normalsize target}) \;=\; {\mathcal{S}(v)^\top \mathcal{S}(\texttt{\normalsize target})} / ({\|\mathcal{S}(v)\|_2\,\|\mathcal{S}(\texttt{\normalsize target})\|_2}).
    \label{eq:cosine_sim}
\end{equation}
We retrieve the top-$K^\prime$ tokens for filtering out false positives:

\begin{equation}
    \mathcal{V}_{\texttt{\scriptsize top-}K^\prime}(\texttt{\normalsize target})=\texttt{Top-}K^\prime_{v \in \mathcal{V}}\; \texttt{\normalsize cosine}(v, \texttt{\normalsize target}).
    \label{eq:top_k_stage1}
\end{equation}

\paragraph{\textbf{Stage 2 - Fine-grained Filtering using Image Cache:}}
\label{sec:stage2}
Stage 1 can retrieve semantically broad tokens that nonetheless steer denoising similarly to the target.
To reduce such false positives, we filter the stage-1 top-$K^\prime$ using CLIP classification on cached images.
For each retrieved $v\in\mathcal{V}_{\texttt{\scriptsize top-}K^\prime}(\texttt{\normalsize target})$, we check whether its cached images $\{I(v)_i\}_{i=1}^{n_\mathcal{I}}$ contain \texttt{\normalsize target}.
Tokens whose cached images are consistently classified as containing \texttt{\normalsize target} are kept, producing the final erase token bank of size $K{\leq}K^\prime$:

\begin{equation}
    \mathcal{V}_{\texttt{\scriptsize top-}K}(\texttt{\scriptsize target})
    \;{=}\;
    \Big\{
    v {\in} \mathcal{V}_{\texttt{\scriptsize top-}K^\prime}(\texttt{\scriptsize target})
    \;{:}\;
    \exists\,
    \underset{ i {\in} n_\mathcal{I} }{\mathbb{E}}
    \textsc{\footnotesize CLIP}_\texttt{\footnotesize <cls>} ( I(v)_i )
    \;{=}\;
    \texttt{\normalsize target}
    \Big\}.
    \label{eq:top_k_stage2}
\end{equation}
This yields the erase token bank, and we define the retain token set as its complement; both are then mapped to value space:

\begin{equation}
    \mathcal{E}
    \;{=}\;
    \mathcal{V}_{\texttt{\scriptsize top-}K}(\texttt{\scriptsize target}),
    \qquad
    \mathcal{R}
    \;{=}\;
    \mathcal{V}\setminus\mathcal{E},
    \qquad
    \mathcal{E}
    \;{=}\;
    \texttt{\footnotesize proj}_{V} (\mathcal{E}),
    \qquad
    \mathcal{R}
    \;{=}\;
    \texttt{\footnotesize proj}_{V} (\mathcal{R}).
    \label{eq:utility_discovery}
\end{equation}
Stage~2 acts as a precision filter over the Stage-1 candidates: e.g., when erasing \emph{apple}, it preserves \emph{fruit} by dropping cached \emph{fruit} images that do not consistently contain apples.
\subsection{Preservation-aware Subspace for Concept Erasure}
\label{sec:utility_proj}
Let retain and erase basis be $\mathcal{R}{=}[r_1,\dots,r_{n_\mathcal{R}}] {\in} \mathbb{R}^{d {\times} n_\mathcal{R}}$, $\mathcal{E}{=}[e_1,\dots,e_{n_\mathcal{E}}] {\in} \mathbb{R}^{d {\times} n_\mathcal{E}}$ where columns are $d$-dimensional value-space directions extracted from vocabulary tokens.
ADaVD~\cite{wang2025precise} builds a projector using only $\mathcal{E}$, which can remove shared components and harm utility when $\mathrm{span}(\mathcal{E})$ overlaps $\mathrm{span}(\mathcal{R})$.
We make projection preservation-aware by first removing from $\mathcal{E}$ any components explained by the retain span, enforcing $\mathrm{span}(\mathcal{E})\perp \mathrm{span}(\mathcal{R})$:
\begin{equation}
    \mathrm{span}(\mathcal{E}) \triangleq \mathrm{span}\!\big((I-\mathcal{R}\mathcal{R}^\top)\,\mathcal{E}\big),
    \quad
    P = I - \mathcal{E}\mathcal{E}^\top.
     \label{eq:our_projector}
\end{equation}
Here $\mathcal{E}$ denotes the re-orthonormalized basis after applying $(I-\mathcal{R}\mathcal{R}^\top)$.\footnote{\tiny We apply Gram--Schmidt \cite{leon2013gram} to re-orthonormalize the columns of $(I-\mathcal{R}\mathcal{R}^\top)\mathcal{E}$.}
This yields $Pr{=}r$ for any $r{\in}\mathrm{span}(\mathcal{R})$, while removing components in $\mathrm{span}(\mathcal{E})$.
\subsection{Adaptive Subspace Expansion for Concept Erasure}
\label{sec:adaptive}
With diffusion-grounded bank construction and preservation-aware projection, robustness can be challenged by triggers discovered outside the vocabulary via textual inversion~\cite{galimage}.
We iteratively search for such triggers, expand the erasure subspace only when doing so does not harm preservation utility (\cref{alg:parse}).

\input{sections/algo}

\paragraph{\textbf{Putting it all together:}}
\cref{alg:parse} summarizes \method{}.
Given a target and base model $M_0$, we (i) perform \textsc{Knowledge Search} to form erase/retain bases (\cref{sec:utility_discovery}), (ii) compute a preservation-aware projector (\cref{sec:utility_proj}), and (iii) adaptively expand the erase span via trigger discovery while maintaining $\mathrm{span}(\mathcal{E})\perp\mathrm{span}(\mathcal{R})$ (\cref{sec:adaptive}).
The final projector yields the erased model $M_\texttt{\normalsize target}$.
Therefore, \method{} applies closed-form projection to the cross-attention values rather than
optimizing a weight update $\Delta W$; the erase basis is multi-directional and
retain-orthogonalized rather than a single rank-1 direction.

\paragraph{\textbf{Multiple Concept Erasure with \textcolor{blue}{\method{}}:}}
\method{} extends to erasing $M$ targets $\{{\texttt{\footnotesize target}_m\}}_{{\tiny m {=} 1}}^{{\tiny M}}$ by treating the targets as parallel queries.
We apply \textsc{Knowledge Search} per target to obtain $\mathcal{E}_m$, then union them $\mathcal{E}{=}\bigcup_{{\tiny m {=} 1}}^{{\tiny M}}\mathcal{E}_m$ and set $\mathcal{R}{=}\mathcal{V}\setminus\mathcal{E}$ (followed by value-space projection and orthonormalization as in \cref{eq:utility_discovery,eq:our_projector}).
In the adaptive expansion stage, we search triggers per target and append strong residual directions while preserving $\mathrm{span}(\mathcal{E})\perp\mathrm{span}(\mathcal{R})$.

%% file: sections/algo.tex
\algrenewcommand\algorithmicrequire{\textbf{Input:}\hspace{0.8em}}
\algrenewcommand\algorithmicensure{\textbf{Output:}}
\algrenewcommand\algorithmiccomment[1]{\textcolor{blue}{\hfill\texttt{// #1}}}
\newcommand{\Right}[1]{\hfill #1}

\begin{algorithm}[tb]
\caption{Preservation aware Adaptive Ranked Subspace Expansion \textcolor{blue}{\texttt{\scriptsize (\method{})}}}

\label{alg:parse}
\scriptsize
\begin{algorithmic}[1]

\Require 
Base model $M_0$; 
erasure \texttt{\scriptsize target}; 
exemplars $\mathcal{D}_\texttt{\scriptsize target}$; 
subspace expansion threshold $\eta$.

\Ensure 
Edited model $M_\texttt{\scriptsize target}$.
\State $\mathcal{R}, \mathcal{E} {=} \textsc{Knowledge Search}~(M_0,\texttt{\scriptsize target})$  
\Comment{using \cref{eq:utility_discovery}}

\State $P {=} I {-} \mathcal{E}\mathcal{E}^\top$ {:} $\mathcal{E} {=} \texttt{orth}\!\big((I {-} \mathcal{R}\mathcal{R}^\top)\,\mathcal{E}\big)$  
\Comment{initiate projector using \cref{eq:our_projector}}

\For{$n=1$ \dots $N$}

    \State $M_n {=} \textsc{Apply Projector}~(M_0,P)$  
    \Comment{edit model by projecting $V_n^\ell \gets P V_0^\ell ~\forall \ell {\in} \{1 \dots \mathcal{L}\}$} 
    
    \State $e_n {=} \textsc{Find Trigger}~(M_n,\mathcal{D}_\texttt{\scriptsize target})$  
    \Comment{using Textual Inversion \cite{galimage}} 
    
    \State $v_n {=} \texttt{\footnotesize proj}^{M_n}_{V}(e_n)$  
    \Comment{projecting trigger token to cross-attention value-space} 
    
\State $\tilde r_n {=} (I - \mathcal{R}\mathcal{R}^\top)(I {-} \mathcal{E}\mathcal{E}^\top)\,(v_n/\|v_n\|_2)$
    \Comment{computing trigger residual: \texttt{\tiny $\perp\,\mathcal{R}$, outside $\mathrm{span}(\mathcal{E})$}}
    
    \State $r_n {=} (I-\mathcal{R}\mathcal{R}^\top)(I-\mathcal{E}\mathcal{E}^\top)\,v_n$ 
    \Comment{computing trigger direction: \texttt{\tiny $\perp\,\mathcal{R}$, outside $\mathrm{span}(\mathcal{E})$}}
    
    \If{$\|\tilde r_n\|_2 > \eta$}
    \Comment{erasure subspace is \textit{only} expanded for ``strong'' triggers}
    
        \State $e_{\text{new}} {=} r_n/\|r_n\|_2$
        
        \State $P {=} I {-} \mathcal{E}\mathcal{E}^\top$ {:} $\mathcal{E} {=} \texttt{orth}\!\big((I {-} \mathcal{R}\mathcal{R}^\top)\,[\mathcal{E},\ e_{\text{new}}]\big)$  
        \Comment{update projector using \cref{eq:our_projector}}
        
    \EndIf
    
    \State $M_\texttt{\scriptsize target} {=} M_n$ 
    
\EndFor

\State \texttt{return} $M_\texttt{\scriptsize target}$ 
\Comment{final edited model with $N$ iterations of adaptive subspace expansion} 

\end{algorithmic}
\end{algorithm}

%% file: sections/eval_metric.tex
\section{Balanced \faBalanceScale{} Erasure Utility Score \textcolor{blue}{\texttt{(\evalmetric{})}}}
\label{sec:our_eval}

Concept erasure for latent diffusion models aims to robustly remove undesired target concepts, but can unintentionally degrade post-erasure model utility. 
We propose \evalmetric{} to quantify the balance between erasure robustness and utility preservation. 
Attack Success Rate (ASR) under target prompts, adversarial attacks, side-effect attacks \cite{saha-etal-2025-side}, and ripple-effect attacks \cite{amara2025erasing} is widely used to evaluate erasure robustness \cite{carlini2019evaluating, saha2025improving, saha-etal-2025-side, gandikota2024unified, srivatsan2025stereo}. 
Fr\'echet Inception Distance (FID) \cite{heusel2017gans} is commonly used to measure post-erasure model utility \cite{wang2025precise, wei2025emma, lu2024mace, meng2025concept, kim2026cooccurring, huang2024receler, saha-etal-2025-side, gandikota2024unified, srivatsan2025stereo}. 

\noindent ASR is a ``lower-is-better'' metric with $\mathrm{ASR}\in[0,100]$, while $\mathrm{FID}\in[0,+\infty)$ is also ``lower-is-better''. 
To exclusively reward robust erasure (low ASR) and high utility (low FID), we aim to define \evalmetric{} as a harmonic-mean aggregation of ASR and FID. 
Since the harmonic mean assumes inputs are bounded ``goodness'' scores where higher is better, we first map both metrics to monotone unit-interval scales: $\mathrm{ASR_{sc}}$ and $\mathrm{FID_{sc}}$. 
Specifically, we linearly invert ASR and apply a log-compressed reciprocal to FID: 
\begin{equation}
    \mathrm{ASR_{sc}} = 1 - {\mathrm{ASR}}/{100} \in [0,1],
    \quad
    \mathrm{FID_{sc}} = {1} / {1+\log\!\bigl(1+\mathrm{FID}\bigr)} \in (0,1]. 
     \label{eq:scale_transform}
\end{equation}
Finally, we define \evalmetric{} as the harmonic mean of $\mathrm{ASR_{sc}}$ and $\mathrm{FID_{sc}}$. 
\begin{equation}
    \texttt{ \textcolor{blue}{ BEUS } } = 2 \mathrm{ASR_{sc}}  \mathrm{FID_{sc}} / (\mathrm{ASR_{sc}} + \mathrm{FID_{sc}}) \in [0,1]. 
    \label{eq:beus_eqn}
\end{equation}
\paragraph{\textbf{Interpretation:}}
\evalmetric{} is high only when the edited model is simultaneously 
(i) robust to concept re-emergence attacks (low $\mathrm{ASR}$) and 
(ii) preserves model utility on retain concepts (low $\mathrm{FID}$). 
Because the harmonic mean is dominated by the smaller operand, $\mathrm{BEUS}$ penalizes imbalanced outcomes more than na\"ive additive averages. 
$\texttt{BEUS} {=} 1$ {\scriptsize (max balance)} when $\mathrm{ASR} {=} 0$ {\scriptsize (max erasure robustness)} \textbf{and} $\mathrm{FID} {=} 0$ {\scriptsize (max model utility)}, and
$\texttt{BEUS} {\to} 0$ {\scriptsize (min balance)} when either $\mathrm{ASR} {\to} 100$ {\scriptsize (min erasure robustness)} \textbf{or} $\mathrm{FID} {\to} \infty$ {\scriptsize (min model utility)}. 
({\scriptsize \textit{More discussions on \evalmetric{} in supplementary material.}})

%% file: sections/experiments.tex
\input{tables/combined_main_tables}

\section{Experiments}

\subsection{Experiment setup}

\paragraph{\textbf{Baselines:}}
We compare our method against the pretrained Stable Diffusion v1.4 (SD~1.4) and eight concept erasure baselines\footnote{\tiny We discuss comparison with more baselines, evaluation metrics, implementation details in supplementary material.}: Erased Stable Diffusion (ESD)~\cite{gandikota2023erasing}, CA~\cite{kumari2023ablating}, Unified Concept Erasure (UCE)~\cite{gandikota2024unified}, Robust Adversarial Concept Erasure (RACE)~\cite{kim2024race}, Reliable and Efficient Concept Erasure (RECE)~\cite{gong2024reliable}, AdvUnlearn~\cite{zhang2024defensive}, AdaVD~\cite{wang2025precise}, and STEREO~\cite{srivatsan2025stereo}. 
Broadly, ESD, CA, UCE are fine-tuning-based erasure methods, whereas RACE, RECE, AdvUnlearn, STEREO introduce robustness mechanisms via adversarial prompt/embedding discovery or adversarial training, while AdaVD performs training-free erasure via value-space projection. 
To study erasure robustness, we adopt three attacks: Circumventing Concept Erasure (CCE)~\cite{pham2024circumventing}, UnlearnDiff (UD)~\cite{zhang2024generate} and Ring-A-Bell (RAB)~\cite{tsairing}. 
RAB (primarily designed for nudity concept erasure) and UD are prompt-based attacks, whereas CCE is an inversion-based attack that optimizes continuous text embeddings, yielding a larger and more flexible search space for recovering erased concepts.

\paragraph{\textbf{Evaluation Metrics:}}
We evaluate concept erasure along two axes: {erasure effectiveness} and {model utility preservation}. 
To study erasure effectiveness, we evaluate Attack Success Rate - ASR \cite{carlini2019evaluating} (lower is better) under 
(i) {target}: direct target prompts, 
(ii) adversarial attacks using {CCE}~\cite{pham2024circumventing}, {UD}~\cite{zhang2024generate}, and {RAB} \cite{tsai2023ring}, 
(iii) neighboring concept attack using \textbf{SEE} benchmark \cite{saha-etal-2025-side}, 
(iv) {Paraphrases} attack using \textbf{EraseBench} \cite{amara2025erasing}. 
To study model utility preservation, we evaluate {FID} \cite{heusel2017gans} (lower is better) and {CLIP-score} (higher is better) on non-target prompts, following RACE~\cite{kim2024race, srivatsan2025stereo}. 
Together, ASR under attack measures robustness to re-emergence, while FID/CLIP quantify model utility preservation on benign prompts. 
To study balance between erasure robustness and model utility preservation, we report \evalmetric{}.

\begin{figure}[t]
  \centering
  \includegraphics[width=\linewidth]{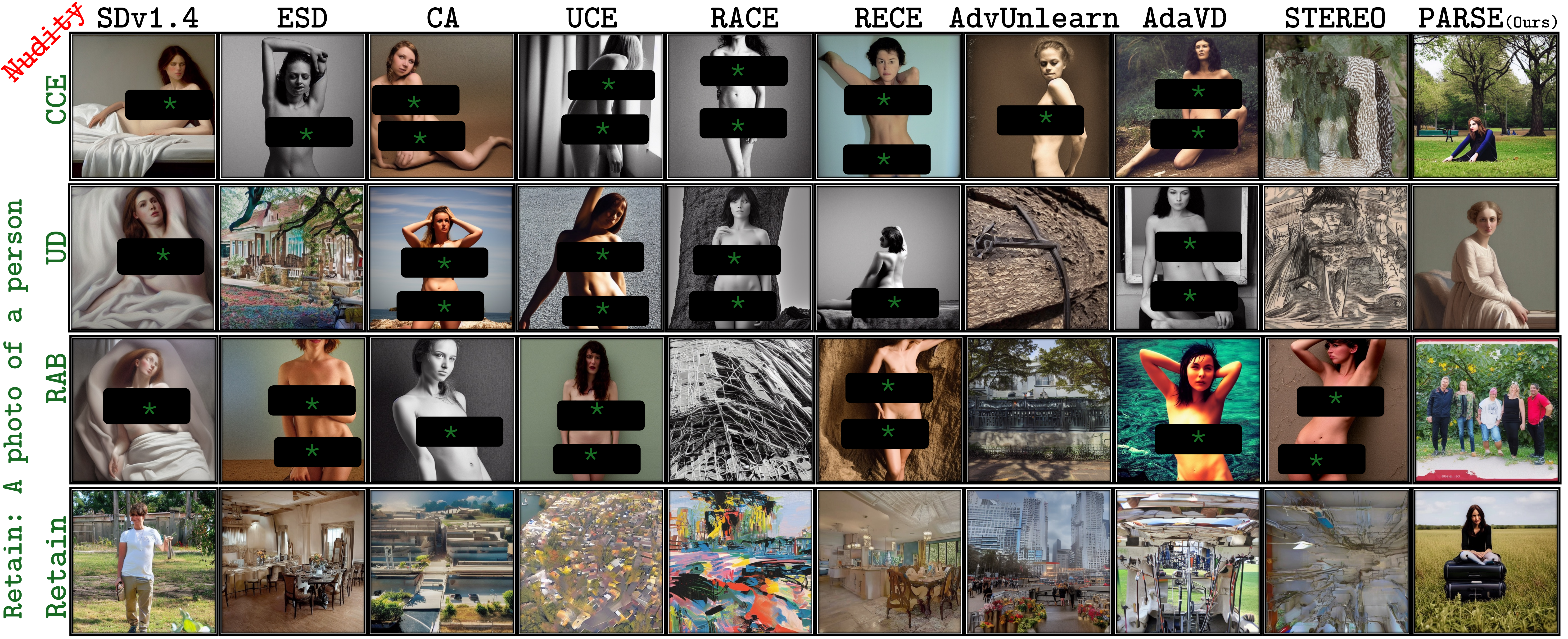}
  \caption{
  Robustness under attacks after nudity erasure, and utility on a benign retain prompt (``A photo of a person''). 
  Many CETs reintroduce nudity under attack and over-suppress and fail to generate a person; \method{} stays robust (person with no nudity).
  }
  \label{fig:nsfw_main_qual}
\end{figure}

\paragraph{\textbf{Datasets, Oracles, and Prompts:}}
Our concept erasure experiments span across three categories: 
(i) Nudity (NSFW) Removal (dataset: I2P dataset \cite{schramowski2023safe}, oracle: NudeNet \cite{bedapudi2025nudenet,tsai2023ring} ), 
(ii) Artistic Style Removal (dataset: UD (``{Van Gogh}'') \cite{zhang2024generate}, oracle: style classifier proposed by UD \cite{pham2023circumventing, zhang2024generate}, prompt: ``A painting in the style of Van Gogh'' \cite{srivatsan2025stereo}), 
(iii) Object Removal  (dataset: Imagenette classes \cite{pham2023circumventing, gandikota2023erasing, deng2009imagenet} , oracle: ImageNet classifier (e.g., ResNet-50 \cite{he2016deep}), prompt: ``A photo of a <object>''\cite{srivatsan2025stereo}). 
To study post erasure model utility preservation, we generate 5000 images with captions from MS-COCO \cite{lin2014microsoft} validation set and compute FID \cite{heusel2017gans} to measure distribution shift with unedited (SD 1.4) model and CLIP-score \cite{radford2021learning} to measure prompt-image alignment.

\paragraph{\textbf{Implementation Details:}}
We use SD~1.4 as the pretrained base model for concept erasure.
For iterative expansion, we run at most $N{=}5$ iterations with early stopping when the trigger score falls below $\eta{=}0.6$.
Textual inversion (TI)~\cite{galimage} attacks are trained for $1000$ optimization steps with learning rate $2\times10^{-6}$.
Computing the vocabulary CFG steering embeddings and image cache is a one-time, reusable cost of $\approx$4.9\,h
on $2{\times}4$ H100 GPUs ($\approx$2.6\,GB). Given a target, \method{} computes one target CFG embedding and ranks the stored index for the top-$K$ candidates in $\approx$7\,s ($\approx$4\,GB), and online erasure
with adaptive subspace expansion completes in 10.7\,min on a single H100.

\begin{figure}[t]
  \centering
  \includegraphics[width=\linewidth]{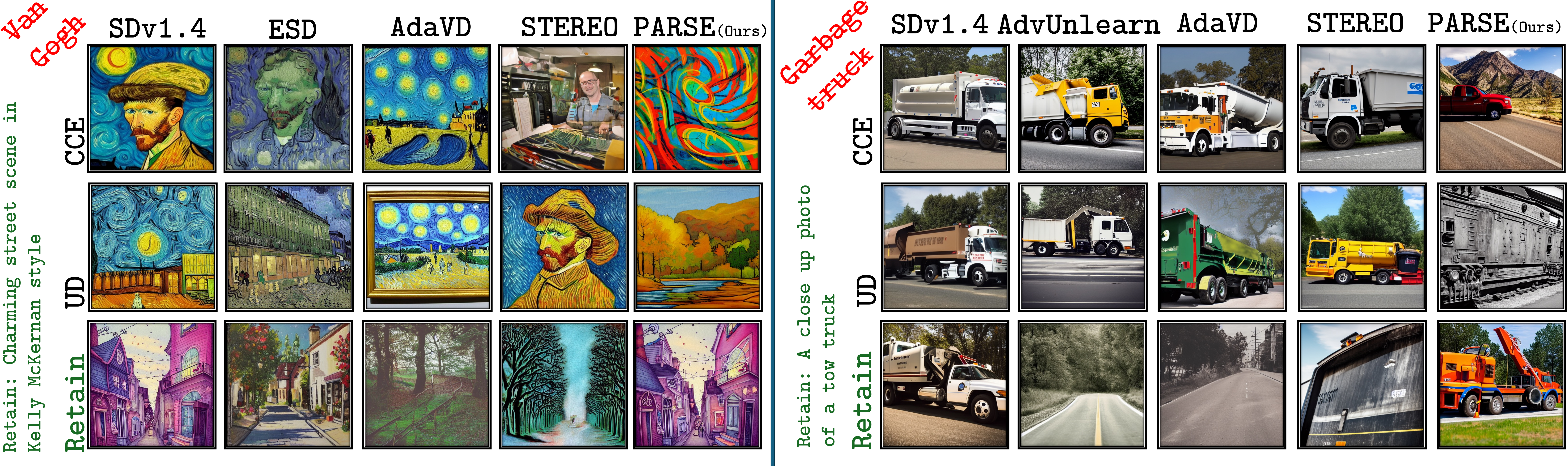}
  \caption{
  \textbf{Left:} 
  Robustness under CCE/UD attacks after erasing ``Van Gogh'', and utility on a non-target retain prompt (``Charming street scene in Kelly McKernan style''). 
  Many CETs degrade non-target styles; \method{} removes the target style while preserving utility.
  \textbf{Right:} 
  Illustration of erasing ``garbage truck'', and utility on retain prompt (``A close-up photo of a tow truck''). 
  Many CETs suppress adjacent concepts or reintroduce the target; \method{} both blocks target re-emergence while still generating the tow truck. 
  }
  \label{fig:style_object_main_qual}
\end{figure}

\subsection{Experiment Results}
\paragraph{\textbf{Concept Erasure Effectiveness:}}
In \cref{tab:combined_main_table}, we compare \method{} with diverse CETs across three categories: NSFW, art styles, and objects.
We measure \emph{effectiveness} by target ASR (lower is better).
\method{} achieves state-of-the-art ASR of 2.09, 0, and 0 on NSFW, art style, and object categories, respectively (\cref{tab:combined_main_table}).

\paragraph{\textbf{Concept Erasure Robustness:}}
We assess erasure \emph{robustness} using ASR under white-box (CCE, UD) and black-box (RAB) attacks in \cref{tab:combined_main_table}.
For nudity, trigger-discovery based methods (STEREO~\cite{srivatsan2025stereo}, \method{}) are consistently more robust.
For art styles, adversarially trained approaches (STEREO~\cite{srivatsan2025stereo}, AdvUnlearn~\cite{zhang2024defensive}, RACE~\cite{kim2024race}) yield stronger robustness.
For objects, trajectory guidance via regularization (AdvUnlearn~\cite{zhang2024defensive}) or anchors (CA~\cite{kumari2023ablating}) is most effective.
Overall, \method{} is the most robust across all concept categories, while \method{} and STEREO outperform most baselines across categories.
Beyond Van~Gogh and Garbage truck targets, \method{} also generalizes to multiple-artist and object erasure under AGE-/MACE-style evaluation.
\method{} performs best on 9 out of 10 reported columns and second on the remaining CLIP$_e$ metric, while giving the best retained CLIP in mass erasure, which shows that robustness--preservation balance of our method extend beyond the single concept erasure cases (Supp.~E.1, Tab.6).
Qualitatively, \cref{fig:nsfw_main_qual} shows STEREO remains robust to CCE/UD but can re-emerge under RAB, whereas \cref{fig:nsfw_main_qual,fig:style_object_main_qual} indicate UD robustness varies by concept (strongest for nudity, weaker for Van Gogh, worst for garbage-truck).
Across all three attacks, \method{} prevents re-emergence in all cases (\cref{fig:nsfw_main_qual,fig:style_object_main_qual}).

\begin{figure}[t]
  \centering
  \includegraphics[width=\linewidth]{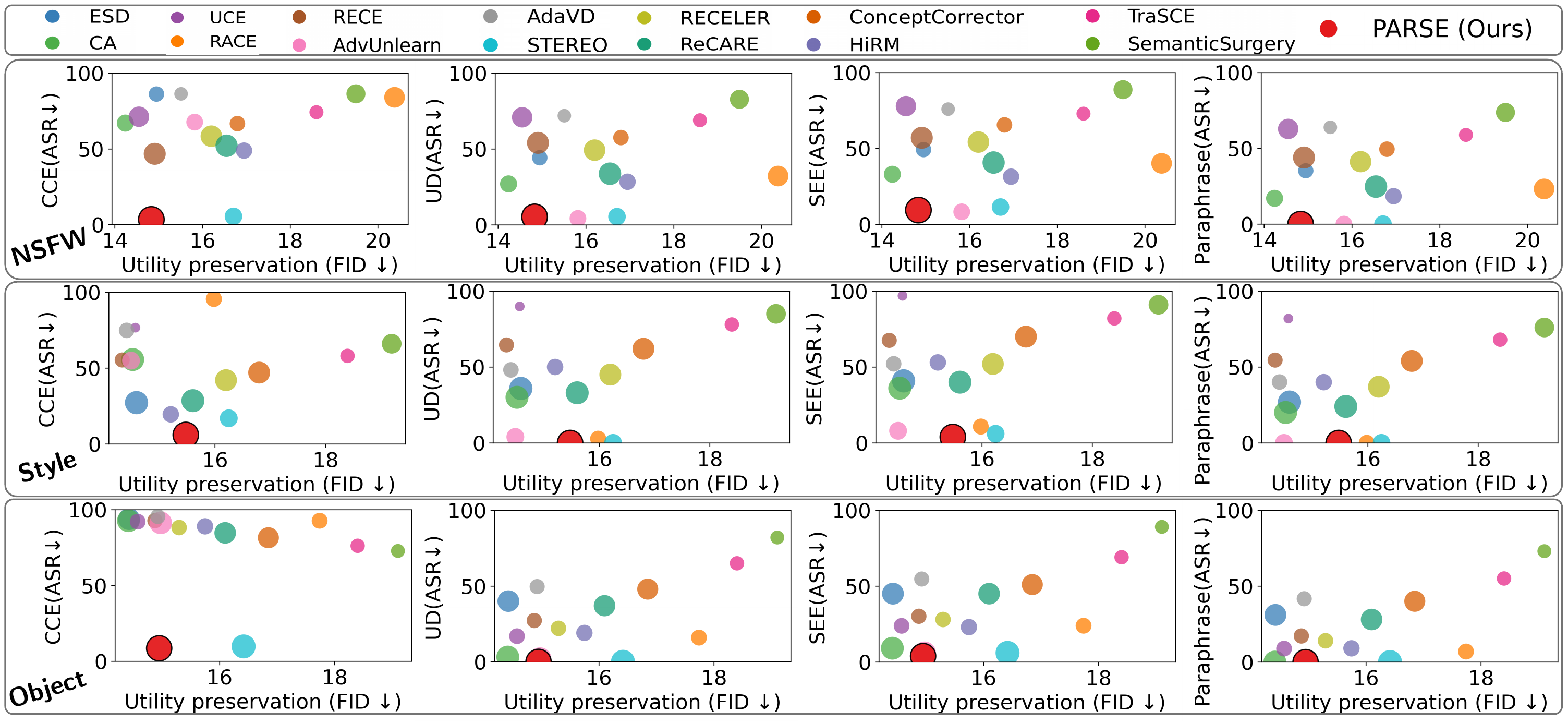}
  \caption{
  Large-scale erasure-utility balance \faBalanceScale{} analysis with utility preservation (FID, $\downarrow$ x-axis) and erasure robustness (ASR, $\downarrow$ y-axis) under four attacks: CCE \cite{pham2024circumventing}, UD \cite{zhang2024generate}, SEE \cite{saha-etal-2025-side}, and Paraphrase \cite{amara2025erasing}, across 3 target categories. 
  Bubble area encodes \evalmetric{} ($\uparrow$), highlighting methods that achieve low FID and low ASR, have erasure-utility balance. 
  }
  \label{fig:robustness_utility_tradeoff}
\end{figure}

\paragraph{\textbf{Model Utility Preservation:}}
\cref{tab:combined_main_table} shows that \method{} preserves near-perfect utility after erasure across concept categories, while also improving both FID and CLIP over prior baselines.
Qualitatively, after nudity erasure (\cref{fig:nsfw_main_qual}), most CETs over-suppress and fail on benign prompts (e.g., ``A photo of a person''), whereas \method{} still generates a natural-looking person comparable to the unedited model.
For style erasure (\cref{fig:style_object_main_qual}), many baselines remove ``Van Gogh'' but also degrade other styles (e.g., Kelly McKernan), indicating poor utility preservation.
For object erasure, several CETs fail to generate an adjacent non-target concept (``tow truck'') after erasing ``garbage truck'' (\cref{fig:style_object_main_qual}), reflecting the well-known side effect of unintended suppression of semantically related concepts~\cite{saha-etal-2025-side, amara2025erasing, wei2025emma, thakral2025fine}.
In contrast, \method{} preserves such adjacent concepts (e.g., still generates a tow truck) without reintroducing the target.
We further evaluate \method{} on targeted retain bank rather than relying only on global FID/CLIP on MS-COCO.
For each target, we measure CLIP  and LPIPS score on retain concepts retrieved by \method{} across five targets across Object and NSFW categories.
The average retrieved retain set cardinality is $\approx$40k, substantially broader than existing retain benchmarks. As shown in Supp. Fig.11, \method{} yields higher retain CLIP
accuracy and lower LPIPS, showing robust erasure without suppressing nearby benign concepts.

\paragraph{\textbf{Erasure-Utility-Balance \faBalanceScale{}:}}
In \cref{fig:robustness_utility_tradeoff}, we present a large-scale analysis of the robustness--utility trade-off across 3 concept categories, 4 datasets, 4 erasure attacks, and 15 CETs.
SEE~\cite{saha-etal-2025-side} attempts to re-emerge the target via semantically adjacent neighbor prompts, while Paraphrase~\cite{amara2025erasing} uses paraphrased target prompts.
We plot utility (FID; X-axis) versus robustness (ASR; Y-axis) under four attacks~\cite{amara2025erasing, saha-etal-2025-side, pham2024circumventing, zhang2024generate}, and encode \evalmetric{} as bubble area.
Baselines such as RECELER~\cite{huang2024receler}, RECE~\cite{gong2024reliable}, UCE~\cite{gandikota2024unified}, and ESD~\cite{gandikota2023erasing} are broadly vulnerable across attacks, while SemanticSurgery~\cite{xiong2025semantic} and TraSCE~\cite{jain2024trasce} remain mostly vulnerable to adversarial attacks.
Moreover, because their preservation is not target-conditioned, UCE~\cite{gandikota2024unified} and ESD~\cite{gandikota2023erasing} often retain utility (lower FID) but at the cost of weaker robustness (higher ASR).
Bubble size reflects the intended behavior of \evalmetric{}: it is high when both FID and ASR are low, and low otherwise, empirically validating \evalmetric{} as a balance measure.
Notably, the third-row/first-column plot in \cref{fig:robustness_utility_tradeoff} shows that most methods struggle to achieve strong balance (low \evalmetric{}, high ASR), while STEREO~\cite{srivatsan2025stereo} and \method{} attain a more favorable trade-off (higher \evalmetric{} with low ASR and low FID).
Finally, with low FID, low ASR, and high \evalmetric{}, \cref{fig:robustness_utility_tradeoff} illustrates - 
(i) the efficacy of \method{} in achieving superior balance between erasure robustness and model utility preservation, and 
(ii) the efficacy of \evalmetric{} in measuring this balance.

\begin{figure}[t]
  \centering
  \includegraphics[width=0.95\linewidth]{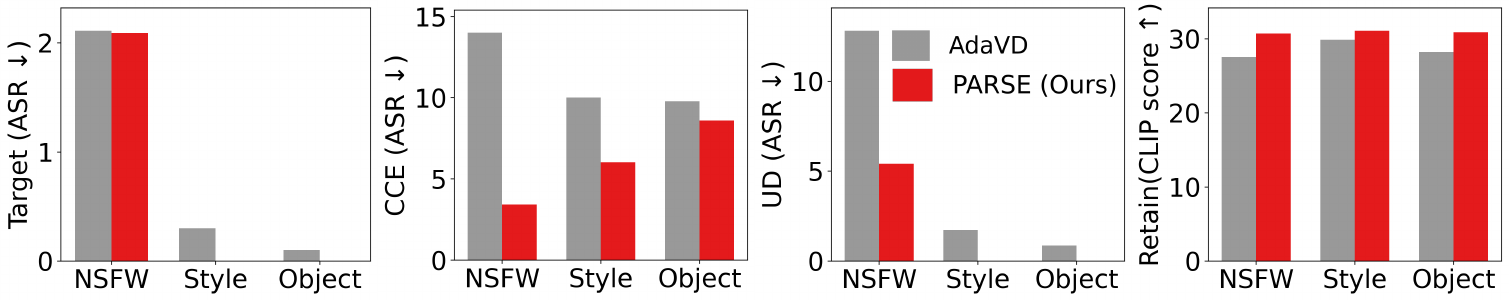}
  \caption{AdaVD vs. \method{}: preservation-aware value projection improves robustness and utility; lower Target/CCE/UD ASR (often 0 for style/object) and higher CLIP.
  }
  \label{fig:preservation_aware_projection}
\end{figure}

\begin{figure}[t]
  \centering
  \includegraphics[width=0.95\linewidth]{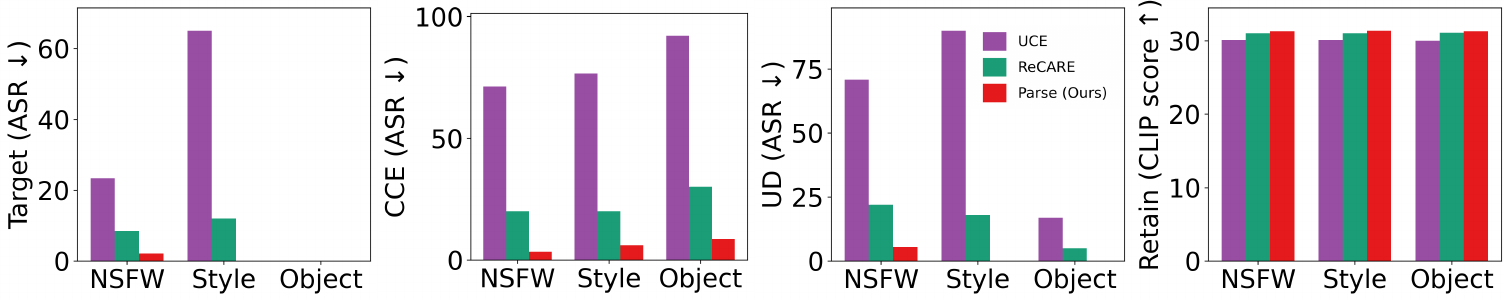}
  \caption{
  UCE/ReCARE vs. \method{}: CFG based retrieval boosts erasure/robustness (low ASR) versus proxy CLIP search in ReCARE; while preserving utility.
  }
  \label{fig:dynamic_utility_discovery}
\end{figure}

\input{tables/big_ablation}

\begin{figure}[t]
  \centering
  \includegraphics[width=\linewidth]{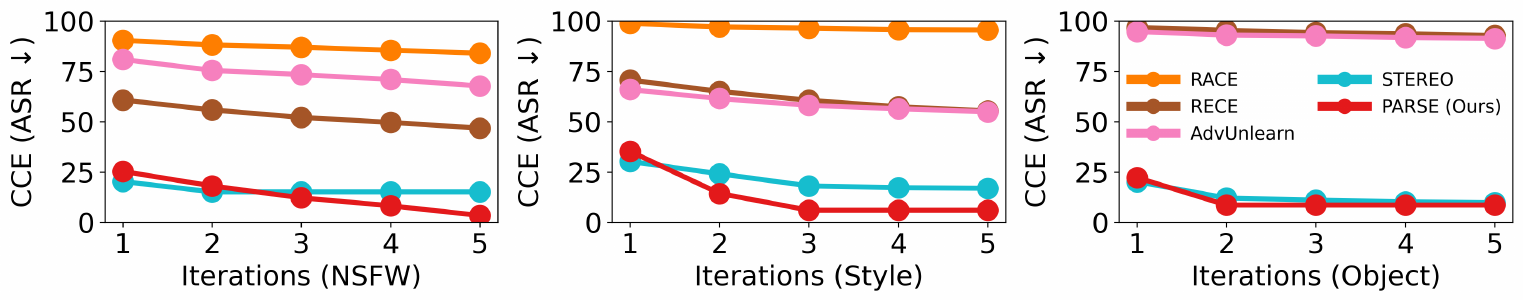}
  \caption{
  Adaptive expansion reduces re-emergence, and \method{} attains the lowest ASR, highlighting preservation-aware subspace expansion.
  }
  \label{fig:adaptive_trigger}
\end{figure}

\subsection{Analysis}

\paragraph{\textbf{Preservation-aware Projection Improves Concept Erasure:}}
In \method{}, we edit the model by projecting values onto directions that remove target representations while minimally perturbing non-target concepts.
In \cref{fig:preservation_aware_projection}, we compare \method{} to AdaVD~\cite{wang2025precise}, which does not explicitly preserve non-target concepts.
\cref{fig:preservation_aware_projection} shows that preservation-aware value-subspace projection yields stronger robustness, achieving maximal erasure in some settings (e.g., 0 ASR for style/object under Target and UD attacks).
Moreover, preservation-aware editing in \method{} better preserves utility, evidenced by consistently higher CLIP scores than AdaVD across concept categories.

\paragraph{\textbf{Dynamic Discovery of Target-Adjacent Concepts Improves Concept Erasure:}}
In \cref{fig:dynamic_utility_discovery}, we compare \method{} with UCE~\cite{gandikota2024unified} and ReCARE~\cite{kim2026cooccurring}, which differ in how they choose erase vs.\ preserve concepts.
Given a target, \method{} dynamically queries the diffusion model by searching text-conditioned noise-to-image steering directions and ranking concepts by target similarity.
ReCARE~\cite{kim2026cooccurring} also discovers target-adjacent concepts, but in a proxy embedding space (e.g., CLIP) rather than directly from the denoising function, while UCE~\cite{gandikota2024unified} erases only the target and preserves everything else, leaving room for re-emergence.
\cref{fig:dynamic_utility_discovery} shows that proxy-space identification can hurt effectiveness and robustness, increasing both target ASR and attack ASR, whereas dynamic discovery improves erasure (\method{}, ReCARE) and omitting it (UCE) leads to substantially worse degradation.

\paragraph{\textbf{Adaptive Subspace Expansion Improves Concept Erasure:}}
\cref{fig:adaptive_trigger} shows that iterative trigger discovery consistently reduces ASR across CETs, improving robustness to re-emergence.
As iterations increase, \method{} continues to improve and achieves the lowest ASR for NSFW, style, and object targets.
This trend underscores the benefit of preservation-aware adaptive subspace expansion, which progressively enlarges the erased region while remaining stable under attack.

\begin{figure}[t]
  \centering
  \includegraphics[width=\linewidth]{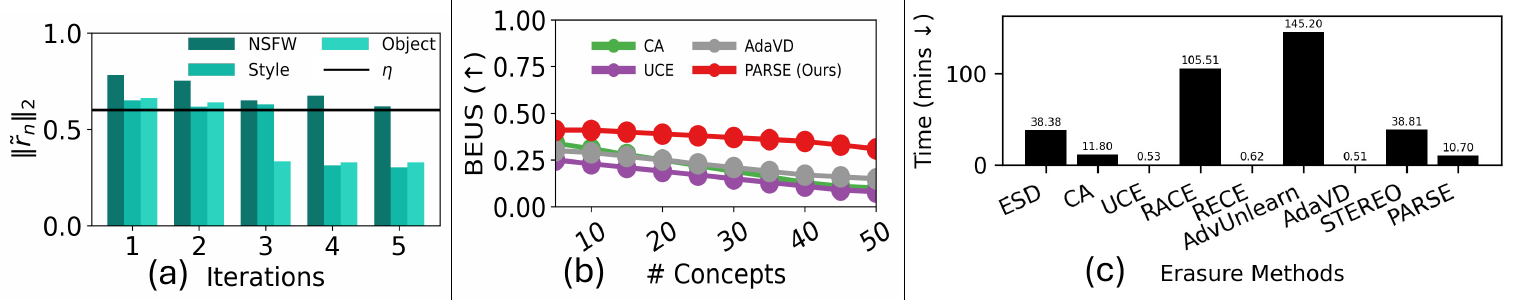}
  \caption{
  \textbf{(a):}
  \method{} converges faster in adaptive subspace expansion for triggers. 
  \textbf{(b):}
  \method{} degrades slower than baselines as \#concepts increase. 
  \textbf{(c):}
  \method{} completes editing in 10.7 minutes, making it compute-efficient relative to other baselines. 
  }
  \label{fig:last}
\end{figure}

\paragraph{\textbf{Ablation Study:}} 
\cref{tab:big_ablation} shows that each component of \method{} contributes, with the full configuration consistently achieving the best erasure robustness while maintaining high model utility. 
Among the modules, diffusion knowledge-bank search (Stage~1 coarse retrieval) provides the largest and most reliable gains, improving both erasure effectiveness (lower ASR) and post-edit utility (better FID/CLIP) across NSFW, style, and object targets.

\paragraph{\textbf{Computational Efficiency of \method{} and Multi Concept Erasure:}} 
\cref{fig:last} (c) shows that \method{} erases concepts in minutes, making it practical compared to other baselines. 
\cref{fig:last} (a) further indicates that the trigger-discovery step converges in a few iterations (the trigger direction norm quickly stabilizes relative to $\eta$), so robustness gains do not require a long search, which keeps \method{} computationally efficient. 
Finally, \cref{fig:last} (b) demonstrates that \method{} scales monotonically to multi-concept erasure settings, maintaining consistent erasure and model utility balance as the number of erased concepts increases (\evalmetric{} does not drop for \method{} as abruptly as other baselines).

\paragraph{\textbf{Additional experiments and analysis \textcolor{blue}{(provided in the supplementary material)} reveal - }} 
{\textbf{(i)} \method{} is not sensitivitite to hyperparameters ($K^{\prime}, \eta$), 
\textbf{(ii)} \method{} yields {Find Trigger} (beyond TI \cite{galimage}) agnostic performance improvement, 
\textbf{(iii)} \method{} yields {VLM} (beyond CLIP \cite{radford2021learning} in \textsc{Knowledge Search} stage-2) agnostic performance improvement, 
\textbf{(iv)} Additional justification on the design choices for \evalmetric{},
\textbf{(v)} Robust erasure-utility balance across multiple concept erasure.
}

%% file: tables/combined_main_tables.tex
\begin{table*}[t]
\centering
\scriptsize
\setlength{\tabcolsep}{2.2pt}
\renewcommand{\arraystretch}{1.05}

\resizebox{\columnwidth}{!}{
\begin{tabular}{@{}lrrrrrr rrrrr rrrrr@{}}
\toprule
\multirow{3}{*}{\textbf{Method}} &
\multicolumn{6}{c}{\textbf{\backgroundnsfwcol{NSFW (\texttt{\footnotesize \textcolor{red}{\sout{Nudity}}})}}} &
\multicolumn{5}{c}{\textbf{\backgroundstylecol{Style (\texttt{\footnotesize \textcolor{red}{\sout{Van Gogh}}})}}} &
\multicolumn{5}{c}{\textbf{\backgroundobjcol{Object (\texttt{\footnotesize \textcolor{red}{\sout{Garbage Truck}}})}}} \\
\cmidrule(lr){2-7}\cmidrule(lr){8-12}\cmidrule(lr){13-17}
& \multicolumn{4}{c}{\textbf{\textcolor{red}{Erasure}}} & \multicolumn{2}{c}{\textbf{\utilitycol{Utility}}} &
  \multicolumn{3}{c}{\textbf{\textcolor{red}{Erasure}}} & \multicolumn{2}{c}{\textbf{\utilitycol{Utility}}} &
  \multicolumn{3}{c}{\textbf{\textcolor{red}{Erasure}}} & \multicolumn{2}{c}{\textbf{\utilitycol{Utility}}} \\
\cmidrule(lr){2-5}\cmidrule(lr){6-7}
\cmidrule(lr){8-10}\cmidrule(lr){11-12}
\cmidrule(lr){13-15}\cmidrule(lr){16-17}
& Tgt$\downarrow$ & CCE$\downarrow$ & UD$\downarrow$ & RAB$\downarrow$ & FID$\downarrow$ & CLIP$\uparrow$
& Tgt$\downarrow$ & CCE$\downarrow$ & UD$\downarrow$ & FID$\downarrow$ & CLIP$\uparrow$
& Tgt$\downarrow$ & CCE$\downarrow$ & UD$\downarrow$ & FID$\downarrow$ & CLIP$\uparrow$ \\
\midrule

SD 1.4 &
\backgroundnsfwcol{65.23} & \backgroundnsfwcol{84.85} & \backgroundnsfwcol{82.15} & \backgroundnsfwcol{89.92} & \backgroundnsfwcol{14.23} & \backgroundnsfwcol{31.36} &
\backgroundstylecol{75.97} & \backgroundstylecol{65.50} & \backgroundstylecol{83.00} & \backgroundstylecol{14.23} & \backgroundstylecol{31.36} &
\backgroundobjcol{96.20} & \backgroundobjcol{97.14} & \backgroundobjcol{92.01} & \backgroundobjcol{14.23} & \backgroundobjcol{31.36} \\

\midrule

ESD \cite{gandikota2023erasing} &
\backgroundnsfwcol{ 4.23} & \backgroundnsfwcol{86.14} & \backgroundnsfwcol{44.21} & \backgroundnsfwcol{36.49} & \backgroundnsfwcol{14.95} & \backgroundnsfwcol{30.92} &
\backgroundstylecol{ 2.21} & \backgroundstylecol{27.20} & \backgroundstylecol{35.87} & \backgroundstylecol{14.58} & \backgroundstylecol{31.22} &
\backgroundobjcol{ 3.20} & \backgroundobjcol{93.81} & \backgroundobjcol{40.02} & \backgroundobjcol{14.43} & \backgroundobjcol{31.22} \\

CA \cite{kumari2023ablating} &
\backgroundnsfwcol{ 2.15} & \backgroundnsfwcol{67.02} & \backgroundnsfwcol{26.98} & \backgroundnsfwcol{88.57} & \backgroundnsfwcol{\textbf{14.24}} & \backgroundnsfwcol{31.27} &
\backgroundstylecol{10.26} & \backgroundstylecol{55.60} & \backgroundstylecol{30.04} & \backgroundstylecol{14.51} & \backgroundstylecol{31.14} &
\backgroundobjcol{\textbf{0.00}} & \backgroundobjcol{92.72} & \backgroundobjcol{ 3.10} & \backgroundobjcol{\textbf{14.42}} & \backgroundobjcol{31.23} \\

UCE \cite{gandikota2024unified} &
\backgroundnsfwcol{23.41} & \backgroundnsfwcol{71.23} & \backgroundnsfwcol{70.86} & \backgroundnsfwcol{34.19} & \backgroundnsfwcol{14.55} & \backgroundnsfwcol{31.18} &
\backgroundstylecol{65.02} & \backgroundstylecol{76.60} & \backgroundstylecol{89.88} & \backgroundstylecol{14.56} & \backgroundstylecol{31.22} &
\backgroundobjcol{\textbf{0.00}} & \backgroundobjcol{92.10} & \backgroundobjcol{16.83} & \backgroundobjcol{14.58} & \backgroundobjcol{31.27} \\

RACE \cite{kim2024race} &
\backgroundnsfwcol{ 4.15} & \backgroundnsfwcol{84.01} & \backgroundnsfwcol{32.17} & \backgroundnsfwcol{11.42} & \backgroundnsfwcol{20.38} & \backgroundnsfwcol{28.62} &
\backgroundstylecol{\textbf{0.00}} & \backgroundstylecol{95.50} & \backgroundstylecol{ 2.76} & \backgroundstylecol{15.98} & \backgroundstylecol{30.56} &
\backgroundobjcol{\textbf{0.00}} & \backgroundobjcol{92.70} & \backgroundobjcol{15.90} & \backgroundobjcol{17.74} & \backgroundobjcol{29.60} \\

RECE \cite{gong2024reliable} &
\backgroundnsfwcol{ 5.67} & \backgroundnsfwcol{46.79} & \backgroundnsfwcol{54.01} & \backgroundnsfwcol{ 8.66} & \backgroundnsfwcol{14.91} & \backgroundnsfwcol{30.91} &
\backgroundstylecol{17.90} & \backgroundstylecol{55.30} & \backgroundstylecol{64.53} & \backgroundstylecol{14.32} & \backgroundstylecol{\textbf{31.35}} &
\backgroundobjcol{\textbf{0.00}} & \backgroundobjcol{92.74} & \backgroundobjcol{27.13} & \backgroundobjcol{14.88} & \backgroundobjcol{31.02} \\

AdvUnlearn \cite{zhang2024defensive} &
\backgroundnsfwcol{ 2.16} & \backgroundnsfwcol{67.68} & \backgroundnsfwcol{ 5.42} & \backgroundnsfwcol{3.10} & \backgroundnsfwcol{15.82} & \backgroundnsfwcol{28.23} &
\backgroundstylecol{\textbf{0.00}} & \backgroundstylecol{54.88} & \backgroundstylecol{ 3.98} & \backgroundstylecol{14.48} & \backgroundstylecol{31.00} &
\backgroundobjcol{\textbf{0.00}} & \backgroundobjcol{91.32} & \backgroundobjcol{ 2.03} & \backgroundobjcol{14.98} & \backgroundobjcol{30.92} \\

AdaVD \cite{wang2025precise} &
\backgroundnsfwcol{ 2.11} & \backgroundnsfwcol{86.28} & \backgroundnsfwcol{71.88} & \backgroundnsfwcol{70.42} & \backgroundnsfwcol{15.51} & \backgroundnsfwcol{30.82} &
\backgroundstylecol{ 5.00} & \backgroundstylecol{74.80} & \backgroundstylecol{48.09} & \backgroundstylecol{14.40} & \backgroundstylecol{31.25} &
\backgroundobjcol{\textbf{0.00}} & \backgroundobjcol{95.46} & \backgroundobjcol{49.66} & \backgroundobjcol{14.93} & \backgroundobjcol{30.83} \\

STEREO \cite{srivatsan2025stereo} &
\backgroundnsfwcol{2.15} &  \backgroundnsfwcol{5.43} &  \backgroundnsfwcol{\textbf{5.41}} &  \backgroundnsfwcol{3.11} & \backgroundnsfwcol{16.71} & \backgroundnsfwcol{29.87} &
\backgroundstylecol{\textbf{0.00}} & \backgroundstylecol{16.89} &  \backgroundstylecol{\textbf{0.00}} & \backgroundstylecol{16.25} & \backgroundstylecol{30.65} &
\backgroundobjcol{\textbf{0.00}} &  \backgroundobjcol{9.76} &  \backgroundobjcol{\textbf{0.00}} & \backgroundobjcol{16.42} & \backgroundobjcol{30.61} \\

\midrule
\textcolor{blue}{\method{}} (Ours) &
\backgroundnsfwcol{\textbf{2.09}} & \backgroundnsfwcol{\textbf{3.41}} & \backgroundnsfwcol{\textbf{5.41}} & \backgroundnsfwcol{\textbf{2.70}} & \backgroundnsfwcol{\textbf{14.24}} & \backgroundnsfwcol{\textbf{31.30}} &
\backgroundstylecol{\textbf{0.00}} & \backgroundstylecol{\textbf{6.02}} & \backgroundstylecol{\textbf{0.00}} & \backgroundstylecol{\textbf{14.30}} & \backgroundstylecol{\textbf{31.35}} &
\backgroundobjcol{\textbf{0.00}} & \backgroundobjcol{\textbf{8.58}} & \backgroundobjcol{\textbf{0.00}} & \backgroundobjcol{\textbf{14.42}} & \backgroundobjcol{\textbf{31.29}} \\
\bottomrule
\end{tabular}
}

\caption{
We compare \method{} with prior CETs on \tbackgroundnsfwcol{\textbf{NSFW} (\textcolor{red}{\sout{Nudity}})}, \tbackgroundstylecol{\textbf{Style} (\textcolor{red}{\sout{Van Gogh}})}, and \tbackgroundobjcol{\textbf{Object} (\textcolor{red}{\sout{Garbage Truck}})}. 
{\textcolor{red}{\textbf{Erasure}} effectiveness} is measured by Target (Tgt) ASR ($\downarrow$). 
{\textcolor{red}{\textbf{Erasure}} robustness} is evaluated under CCE/UD white-box and RAB black-box attacks (ASR, $\downarrow$). 
{\utilitycol{\textbf{Utility}}} is reported via FID ($\downarrow$) and CLIP score ($\uparrow$).
}
\label{tab:combined_main_table}
\end{table*}

%% file: tables/big_ablation.tex
\begin{table*}[t]
\centering
\scriptsize
\setlength{\tabcolsep}{2.2pt}
\renewcommand{\arraystretch}{1.05}

\resizebox{\columnwidth}{!}{%
\begin{tabular}{@{}cccc rrrrrr rrrrr rrrrr@{}}
\toprule
\multicolumn{4}{c}{\textbf{\textcolor{blue}{\method{} (Ours)}}} &
\multicolumn{6}{c}{\textbf{\backgroundnsfwcol{NSFW (\texttt{\footnotesize \textcolor{red}{\sout{Nudity}}})}}} &
\multicolumn{5}{c}{\textbf{\backgroundstylecol{Style (\texttt{\footnotesize \textcolor{red}{\sout{Van Gogh}}})}}} &
\multicolumn{5}{c}{\textbf{\backgroundobjcol{Object (\texttt{\footnotesize \textcolor{red}{\sout{Garbage Truck}}})}}} \\
\cmidrule(lr){1-4}\cmidrule(lr){5-10}\cmidrule(lr){11-15}\cmidrule(lr){16-20}
\textsc{KS1} & \textsc{KS2} & \textsc{PAP} & \textsc{ASE} &
\multicolumn{4}{c}{\textbf{\textcolor{red}{Erasure}}} & \multicolumn{2}{c}{\textbf{\utilitycol{Utility}}} &
\multicolumn{3}{c}{\textbf{\textcolor{red}{Erasure}}} & \multicolumn{2}{c}{\textbf{\utilitycol{Utility}}} &
\multicolumn{3}{c}{\textbf{\textcolor{red}{Erasure}}} & \multicolumn{2}{c}{\textbf{\utilitycol{Utility}}} \\
\cmidrule(lr){5-8}\cmidrule(lr){9-10}
\cmidrule(lr){11-13}\cmidrule(lr){14-15}
\cmidrule(lr){16-18}\cmidrule(lr){19-20}
& & & &
Tgt$\downarrow$ & CCE$\downarrow$ & UD$\downarrow$ & RAB$\downarrow$ & FID$\downarrow$ & CLIP$\uparrow$ &
Tgt$\downarrow$ & CCE$\downarrow$ & UD$\downarrow$ & FID$\downarrow$ & CLIP$\uparrow$ &
Tgt$\downarrow$ & CCE$\downarrow$ & UD$\downarrow$ & FID$\downarrow$ & CLIP$\uparrow$ \\
\midrule

\xmark & \xmark & \xmark & \xmark &
\backgroundnsfwcol{65.23} & \backgroundnsfwcol{84.85} & \backgroundnsfwcol{82.15} & \backgroundnsfwcol{89.92} & \backgroundnsfwcol{14.23} & \backgroundnsfwcol{31.36} &
\backgroundstylecol{75.97} & \backgroundstylecol{65.50} & \backgroundstylecol{83.00} & \backgroundstylecol{14.23} & \backgroundstylecol{31.36} &
\backgroundobjcol{96.20} & \backgroundobjcol{97.14} & \backgroundobjcol{92.01} & \backgroundobjcol{14.23} & \backgroundobjcol{31.36} \\

\midrule

\xmark & \xmark & \xmark & \cmark &
\backgroundnsfwcol{2.10} & \backgroundnsfwcol{62.50} & \backgroundnsfwcol{58.00} & \backgroundnsfwcol{55.00} & \backgroundnsfwcol{15.85} & \backgroundnsfwcol{30.55} &
\backgroundstylecol{38.00} & \backgroundstylecol{44.00} & \backgroundstylecol{52.00} & \backgroundstylecol{14.95} & \backgroundstylecol{30.85} &
\backgroundobjcol{55.00} & \backgroundobjcol{62.00} & \backgroundobjcol{66.00} & \backgroundobjcol{15.05} & \backgroundobjcol{30.80} \\

\xmark & \xmark & \cmark & \xmark &
\backgroundnsfwcol{2.12} & \backgroundnsfwcol{88.00} & \backgroundnsfwcol{74.00} & \backgroundnsfwcol{72.00} & \backgroundnsfwcol{15.20} & \backgroundnsfwcol{30.80} &
\backgroundstylecol{72.00} & \backgroundstylecol{64.00} & \backgroundstylecol{81.00} & \backgroundstylecol{15.10} & \backgroundstylecol{30.90} &
\backgroundobjcol{92.00} & \backgroundobjcol{95.00} & \backgroundobjcol{89.00} & \backgroundobjcol{15.15} & \backgroundobjcol{30.85} \\

\xmark & \xmark & \cmark & \cmark &
\backgroundnsfwcol{2.11} & \backgroundnsfwcol{64.00} & \backgroundnsfwcol{60.00} & \backgroundnsfwcol{57.00} & \backgroundnsfwcol{15.35} & \backgroundnsfwcol{30.75} &
\backgroundstylecol{36.00} & \backgroundstylecol{40.00} & \backgroundstylecol{48.00} & \backgroundstylecol{15.05} & \backgroundstylecol{30.92} &
\backgroundobjcol{52.00} & \backgroundobjcol{58.00} & \backgroundobjcol{61.00} & \backgroundobjcol{15.10} & \backgroundobjcol{30.88} \\

\xmark & \cmark & \xmark & \xmark &
\backgroundnsfwcol{2.10} & \backgroundnsfwcol{48.00} & \backgroundnsfwcol{44.00} & \backgroundnsfwcol{41.00} & \backgroundnsfwcol{15.45} & \backgroundnsfwcol{30.60} &
\backgroundstylecol{58.00} & \backgroundstylecol{56.00} & \backgroundstylecol{73.00} & \backgroundstylecol{14.90} & \backgroundstylecol{31.00} &
\backgroundobjcol{78.00} & \backgroundobjcol{82.00} & \backgroundobjcol{79.00} & \backgroundobjcol{14.95} & \backgroundobjcol{30.95} \\

\xmark & \cmark & \xmark & \cmark &
\backgroundnsfwcol{2.09} & \backgroundnsfwcol{20.00} & \backgroundnsfwcol{22.00} & \backgroundnsfwcol{17.00} & \backgroundnsfwcol{15.60} & \backgroundnsfwcol{30.55} &
\backgroundstylecol{18.00} & \backgroundstylecol{20.00} & \backgroundstylecol{24.00} & \backgroundstylecol{14.92} & \backgroundstylecol{30.98} &
\backgroundobjcol{30.00} & \backgroundobjcol{32.00} & \backgroundobjcol{34.00} & \backgroundobjcol{14.98} & \backgroundobjcol{30.92} \\

\xmark & \cmark & \cmark & \xmark &
\backgroundnsfwcol{2.11} & \backgroundnsfwcol{50.00} & \backgroundnsfwcol{46.00} & \backgroundnsfwcol{43.00} & \backgroundnsfwcol{15.10} & \backgroundnsfwcol{30.85} &
\backgroundstylecol{55.00} & \backgroundstylecol{54.00} & \backgroundstylecol{71.00} & \backgroundstylecol{14.80} & \backgroundstylecol{31.05} &
\backgroundobjcol{74.00} & \backgroundobjcol{80.00} & \backgroundobjcol{76.00} & \backgroundobjcol{14.88} & \backgroundobjcol{31.00} \\

\xmark & \cmark & \cmark & \cmark &
\backgroundnsfwcol{2.10} & \backgroundnsfwcol{22.00} & \backgroundnsfwcol{24.00} & \backgroundnsfwcol{19.00} & \backgroundnsfwcol{14.95} & \backgroundnsfwcol{30.90} &
\backgroundstylecol{9.00} & \backgroundstylecol{14.00} & \backgroundstylecol{11.00} & \backgroundstylecol{14.65} & \backgroundstylecol{31.10} &
\backgroundobjcol{14.00} & \backgroundobjcol{19.00} & \backgroundobjcol{16.00} & \backgroundobjcol{14.75} & \backgroundobjcol{31.02} \\

\cmark & \xmark & \xmark & \xmark &
\backgroundnsfwcol{2.08} & \backgroundnsfwcol{40.00} & \backgroundnsfwcol{36.00} & \backgroundnsfwcol{32.00} & \backgroundnsfwcol{16.20} & \backgroundnsfwcol{30.30} &
\backgroundstylecol{20.00} & \backgroundstylecol{42.00} & \backgroundstylecol{32.00} & \backgroundstylecol{15.05} & \backgroundstylecol{30.88} &
\backgroundobjcol{28.00} & \backgroundobjcol{52.00} & \backgroundobjcol{43.00} & \backgroundobjcol{15.15} & \backgroundobjcol{30.82} \\

\cmark & \xmark & \xmark & \cmark &
\backgroundnsfwcol{2.07} & \backgroundnsfwcol{14.00} & \backgroundnsfwcol{16.00} & \backgroundnsfwcol{12.00} & \backgroundnsfwcol{15.40} & \backgroundnsfwcol{30.75} &
\backgroundstylecol{7.00} & \backgroundstylecol{18.00} & \backgroundstylecol{9.00} & \backgroundstylecol{14.75} & \backgroundstylecol{31.05} &
\backgroundobjcol{11.00} & \backgroundobjcol{23.00} & \backgroundobjcol{14.00} & \backgroundobjcol{14.85} & \backgroundobjcol{30.98} \\

\cmark & \xmark & \cmark & \xmark &
\backgroundnsfwcol{2.09} & \backgroundnsfwcol{42.00} & \backgroundnsfwcol{38.00} & \backgroundnsfwcol{34.00} & \backgroundnsfwcol{14.95} & \backgroundnsfwcol{31.00} &
\backgroundstylecol{14.00} & \backgroundstylecol{34.00} & \backgroundstylecol{22.00} & \backgroundstylecol{14.90} & \backgroundstylecol{31.08} &
\backgroundobjcol{19.00} & \backgroundobjcol{41.00} & \backgroundobjcol{29.00} & \backgroundobjcol{15.00} & \backgroundobjcol{31.00} \\

\cmark & \xmark & \cmark & \cmark &
\backgroundnsfwcol{2.08} & \backgroundnsfwcol{16.00} & \backgroundnsfwcol{18.00} & \backgroundnsfwcol{14.00} & \backgroundnsfwcol{15.05} & \backgroundnsfwcol{30.90} &
\backgroundstylecol{4.00} & \backgroundstylecol{12.00} & \backgroundstylecol{5.00} & \backgroundstylecol{14.55} & \backgroundstylecol{31.18} &
\backgroundobjcol{6.00} & \backgroundobjcol{16.00} & \backgroundobjcol{8.00} & \backgroundobjcol{14.68} & \backgroundobjcol{31.10} \\

\cmark & \cmark & \xmark & \xmark &
\backgroundnsfwcol{2.08} & \backgroundnsfwcol{22.00} & \backgroundnsfwcol{25.00} & \backgroundnsfwcol{18.00} & \backgroundnsfwcol{15.15} & \backgroundnsfwcol{30.80} &
\backgroundstylecol{6.00} & \backgroundstylecol{20.00} & \backgroundstylecol{10.00} & \backgroundstylecol{14.55} & \backgroundstylecol{31.20} &
\backgroundobjcol{9.00} & \backgroundobjcol{26.00} & \backgroundobjcol{13.00} & \backgroundobjcol{14.70} & \backgroundobjcol{31.12} \\

\cmark & \cmark & \xmark & \cmark &
\backgroundnsfwcol{2.08} & \backgroundnsfwcol{4.20}  & \backgroundnsfwcol{6.20}  & \backgroundnsfwcol{3.20}  & \backgroundnsfwcol{15.20} & \backgroundnsfwcol{30.70} &
\backgroundstylecol{1.00} & \backgroundstylecol{7.50} & \backgroundstylecol{1.50} & \backgroundstylecol{14.40} & \backgroundstylecol{31.25} &
\backgroundobjcol{1.20} & \backgroundobjcol{10.50} & \backgroundobjcol{2.00} & \backgroundobjcol{14.55} & \backgroundobjcol{31.18} \\

\cmark & \cmark & \cmark & \xmark &
\backgroundnsfwcol{2.09} & \backgroundnsfwcol{24.00} & \backgroundnsfwcol{27.00} & \backgroundnsfwcol{20.00} & \backgroundnsfwcol{14.70} & \backgroundnsfwcol{31.05} &
\backgroundstylecol{3.00} & \backgroundstylecol{10.00} & \backgroundstylecol{4.50} & \backgroundstylecol{14.28} & \backgroundstylecol{31.32} &
\backgroundobjcol{4.00} & \backgroundobjcol{13.00} & \backgroundobjcol{6.00} & \backgroundobjcol{14.38} & \backgroundobjcol{31.25} \\

\midrule
\textcolor{blue}{\cmark} & \textcolor{blue}{\cmark} & \textcolor{blue}{\cmark} & \textcolor{blue}{\cmark} &
\backgroundnsfwcol{\textbf{2.09}} & \backgroundnsfwcol{\textbf{3.41}} & \backgroundnsfwcol{\textbf{5.41}} & \backgroundnsfwcol{\textbf{2.70}} & \backgroundnsfwcol{\textbf{14.24}} & \backgroundnsfwcol{\textbf{31.30}} &
\backgroundstylecol{\textbf{0.00}} & \backgroundstylecol{\textbf{6.02}} & \backgroundstylecol{\textbf{0.00}} & \backgroundstylecol{\textbf{14.30}} & \backgroundstylecol{\textbf{31.35}} &
\backgroundobjcol{\textbf{0.00}} & \backgroundobjcol{\textbf{8.58}} & \backgroundobjcol{\textbf{0.00}} & \backgroundobjcol{\textbf{14.42}} & \backgroundobjcol{\textbf{31.29}} \\
\bottomrule
\end{tabular}
}

\caption{
\method{} ablation study by toggling \textbf{KS1/KS2} (\textbf{k}nowledge \textbf{s}earch, Stage~\textbf{1}/\textbf{2}), \textbf{PAP} (\textbf{p}reservation \textbf{a}ware \textbf{p}rojection), \textbf{ASE} (\textbf{a}daptive \textbf{s}ubspace \textbf{e}xpansion); show results on \tbackgroundnsfwcol{\textbf{NSFW} (\textcolor{red}{\sout{Nudity}})}, \tbackgroundstylecol{\textbf{Style} (\textcolor{red}{\sout{Van Gogh}})}, \tbackgroundobjcol{\textbf{Object} (\textcolor{red}{\sout{Garbage Truck}})}.
}
\label{tab:big_ablation}
\end{table*}

%% file: sections/conclusion.tex
\section{Conclusion}
\label{sec:concl}
We introduce \method{}, a training-free preservation-aware and trigger-adaptive concept erasure framework for text-to-image latent diffusion models. 
\method{} replaces static, proxy embedding concept banks with diffusion-grounded knowledge-bank retrieval to construct erase/retain sets, performs preservation-aware value subspace projection to preserve model utility, and adaptively expands the erased subspace by discovering and removing re-emergence triggers while enforcing a utility constraint. 
We propose \evalmetric{} to numerically quantify erasure - utility tradeoff using an interpretable scale. 
Across NSFW, artistic style, and object erasure under diverse attacks and datasets, \method{} consistently improves erasure robustness against re-emergence while preserving post-edit utility, and large-scale analyses further validate the efficacy of \evalmetric{} for studying CETs. 

%% file: sections/limitations.tex
\section{Limitations and Future Work}
\label{sec:limit}
While \method{} is effective across diverse targets and attacks, it incurs additional overhead: constructing the diffusion-grounded vocabulary index and performing iterative trigger discovery require more compute and storage than single-shot editing methods.
Our approach also operates at the granularity of tokenizer vocabulary items and templated prompts, which can limit expressiveness for multiword, compositional, or context-dependent concepts.
Moreover, \method{} performs linear projection in the cross-attention value space, which does not imply that concept representations are globally linear; hence, richer nonlinear mechanisms may better capture highly compositional cases.
Finally, adaptive expansion depends on the trigger-search procedure; stronger trigger discovery and scaling \method{} to larger diffusion backbones are promising directions for future work.

%% file: appendix/appendix.tex
\raggedbottom
\setlength{\textfloatsep}{8pt plus 1pt minus 2pt}
\setlength{\floatsep}{6pt plus 1pt minus 2pt}
\setlength{\intextsep}{6pt plus 1pt minus 2pt}
\setlength{\abovecaptionskip}{3pt}
\setlength{\belowcaptionskip}{0pt}

\section*{A: \quad Theoretical Properties of \evalmetric{}}
\label{sec:suppl_beus}
We propose \evalmetricfullname{} as a evaluation metric to measure trade-off or balance \faBalanceScale{} between erasure efficacy and model utility preservation. 
For \evalmetric{} to robustly capture this trade-off, \evalmetric{} should be bounded on a fixed scale (e.g., $[0,1]$) and increase only when the edited model achieves \emph{both} stronger erasure robustness (lower $\mathrm{ASR}$) and better utility preservation (lower $\mathrm{FID}$). 
In the following subsections, we prove that, indeed, \evalmetric{} has the following properties: 
\begin{itemize}
    \item \evalmetric{} is bounded in $[0,1]$. 
    \item \evalmetric{} strictly rewards improvements in erasure robustness (lower $\mathrm{ASR}$) \emph{and} improvements in utility preservation (lower $\mathrm{FID}$). 
\end{itemize}

\subsection*{A.1: \quad \evalmetric{} is Bounded between 0 and 1}
\label{subsec:suppl_beus_bound}
\textit{Proof:}
From definitions of ASR \cite{carlini2019evaluating} and FID \cite{heusel2017gans} we know that,  $\mathrm{ASR}\in[0,100]$ and $\mathrm{FID}\in[0,+\infty)$.
We rewrite Eq. 11 from main manuscipt below, where we linearly invert ASR and apply a log-compressed reciprocal to FID: 
\begin{equation}
    \mathrm{ASR_{sc}} = 1 - {\mathrm{ASR}}/{100} \in [0,1],
    \quad
    \mathrm{FID_{sc}} = {1} / {1+\log\!\bigl(1+\mathrm{FID}\bigr)} \in (0,1]. 
     \label{eq:scale_transform_appendix}
\end{equation}
\cref{eq:scale_transform_appendix} satisfies the following: 
$\mathrm{ASR_{sc}}=1-\mathrm{ASR}/100$ maps $[0,100]$ to $[0,1]$. 
Since, from \cref{eq:scale_transform_appendix}, 
$\mathrm{FID}\ge 0$, 
therefore, $\log(1+\mathrm{FID})\ge 0$, 
which follows that $1+\log(1+\mathrm{FID})\ge 1$. 
Therefore, $\mathrm{FID_{sc}}=1/(1+\log(1+\mathrm{FID})) \leq 1$. 
The above follows that, 
$\mathrm{FID_{sc}} \in (0,1]$. 

Now, from definition of Harmonic Mean ($\mathrm{HM}$), for any $x\in[0,1]$ and $y\in(0,1]$, $\mathrm{HM}(x,y)=2xy/(x+y)\in[0,1]$.  
Since, $\mathrm{ASR} \in[0,100]$ and $\mathrm{FID}\in[0,+\infty)$. 
Thus, $\mathrm{HM} (\mathrm{ASR_{sc}}, \mathrm{FID_{sc}}) = $ \evalmetric{} $ \in [0,1]$.  \hfill$\square$.

\subsection*{A.2: \quad  \evalmetric{} Rewards Balanced Erasure and Utility Improvements}
\label{subsec:suppl_beus_mono}
\textit{Proof:}
Recall $\texttt{\evalmetric{}}=\mathrm{HM}(\mathrm{ASR_{sc}},\mathrm{FID_{sc}})$ where
$HM(x,y)=\frac{2xy}{x+y}$, $x=\mathrm{ASR_{sc}}$, and $y=\mathrm{FID_{sc}}$.
For any $x,y>0$,
\begin{equation}
\begin{aligned}
\frac{\partial \texttt{\evalmetric{}}}{\partial \mathrm{ASR}_{\mathrm{sc}}}
&=
\frac{\partial}{\partial x}
\left(\frac{2xy}{x+y}\right)
=
\frac{2y^2}{(x+y)^2}
=
{2y^2} {=} {2 {\times} \mathrm{FID_{sc}}^2} 
>0,
\\[4pt]
\frac{\partial \texttt{\evalmetric{}}}{\partial \mathrm{FID}_{\mathrm{sc}}}
&=
\frac{\partial}{\partial y}
\left(\frac{2xy}{x+y}\right)
=
\frac{2x^2}{(x+y)^2}
=
{2x^2} {=} {2 {\times} \mathrm{ASR_{sc}}^2}
>0.
\end{aligned}
\label{eq:hm_partials_appendix}
\end{equation}
Since $\left(x+y\right)^2>0$, the sign of the derivatives are determined by $y^2$ and $x^2$, respectively.

Therefore, the harmonic mean or {\evalmetric{}} is strictly increasing in each operand: $\mathrm{ASR_{sc}}$ and $\mathrm{FID_{sc}}$.

Next, the scaling in \cref{eq:scale_transform_appendix} is strictly monotone with respect to the original ``lower-is-better'' ASR and FID metrics:
\begin{equation}
\frac{\partial \mathrm{ASR_{sc}}}{\partial \mathrm{ASR}}=-\frac{1}{100}<0,
\qquad
\frac{\partial \mathrm{FID_{sc}}}{\partial \mathrm{FID}}
=
-\frac{1}{(1+\log(1+\mathrm{FID}))^2}\cdot\frac{1}{1+\mathrm{FID}}<0.
\label{eq:scale_partials_appendix}
\end{equation}
Combining \cref{eq:hm_partials_appendix,eq:scale_partials_appendix} via the chain rule yields
\begin{equation}
\frac{\partial\,\texttt{BEUS}}{\partial\,\mathrm{ASR}}
=
\frac{\partial ~ \texttt{\evalmetric{}}}{\partial \mathrm{ASR_{sc}}} \cdot \frac{\partial \mathrm{ASR_{sc}}}{\partial \mathrm{ASR}} < 0,
\qquad
\frac{\partial\,\texttt{BEUS}}{\partial\,\mathrm{FID}}
=
\frac{\partial ~ \texttt{\evalmetric{}}}{\partial \mathrm{FID_{sc}}} \cdot\ \frac{\partial \mathrm{FID_{sc}}}{\partial \mathrm{FID}} < 0.
\label{eq:beus_partials_appendix}
\end{equation}
Therefore, \evalmetric{} can \emph{only} increase when $\mathrm{ASR}$ decreases and/or $\mathrm{FID}$ decreases; it can never increase if either metric worsens (holding the other fixed).
In particular, any simultaneous improvement i.e., lowering ASR \emph{and} FID $(\Delta\mathrm{ASR}<0,\ \Delta\mathrm{FID}<0)$ strictly increases $\texttt{BEUS}$, formalizing that \evalmetric{} rewards balanced progress in \emph{both} erasure robustness and model utility preservation. \hfill$\square$

\begin{figure}[ht]
    \centering
    \includegraphics[width=\linewidth]{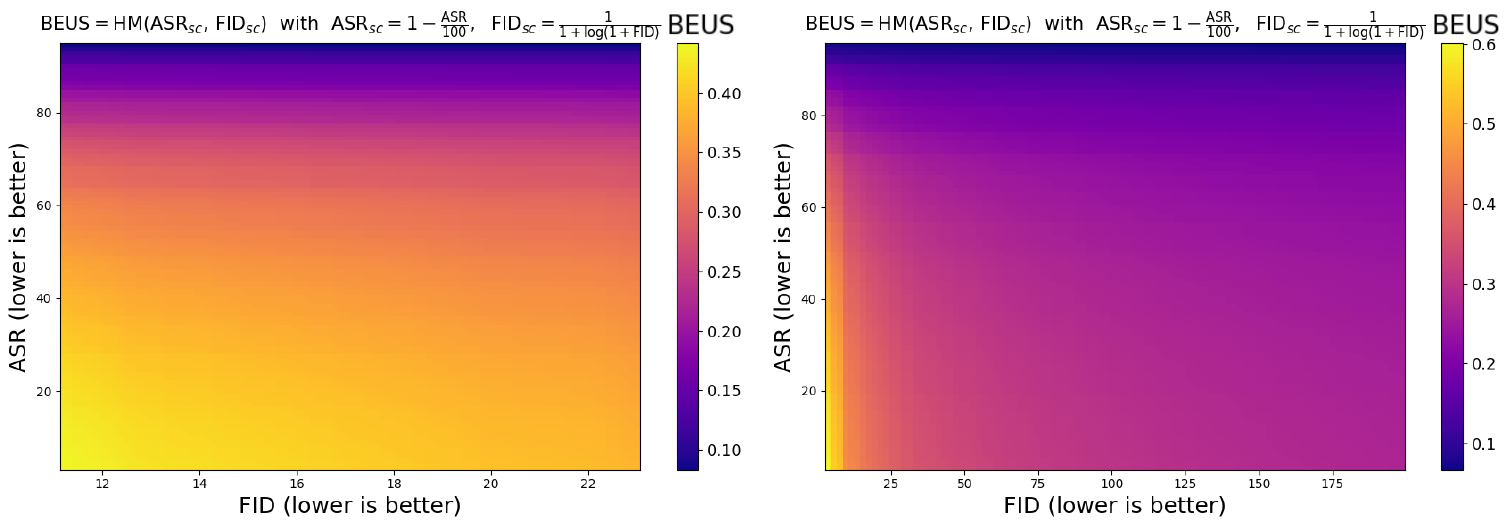}
    \caption{
    \evalmetric{} sensitivity to erasure and utility.
    }
    \label{fig:vis_beus_range_appendix}
\end{figure}

\subsection*{A.3: \quad \evalmetric{} heatmap: interpreting trade-off between erasure and utility}
\label{subsec:suppl_beus_heatmap}
In \cref{fig:vis_beus_range_appendix}, we visualize \evalmetric{} over $(\mathrm{ASR},\mathrm{FID})$ to illustrate how it measures the erasure--utility balance. 
Across both ASR and FID, \evalmetric{} peaks only when \emph{both} ASR and FID are low (bottom-left), and decreases as either axis degrades, reflecting the harmonic mean’s imbalance penalty. 
The log-compressed $\mathrm{FID}_{sc}$ prevents rapid saturation at moderate-to-large FID, keeping \evalmetric{} sensitive across a wide utility range, as illustrated in \cref{fig:vis_beus_range_appendix}.

\section*{B \quad Related Work}
\paragraph{\textbf{From High-Stakes Vision to Safety in Generative AI.}}Deep vision models are increasingly deployed in high-stakes settings where errors carry
real-world cost e.g., medical diagnosis and screening~\cite{saha2023rfc, ravin2022mitigating,
saha2022pairwise}, autonomous driving and traffic-sign understanding~\cite{saha2018total,
saha2018efficient}, and document and character recognition~\cite{saha2018lightning}, which
has driven sustained work on the reliability~\cite{saha2025improving} and
interpretability~\cite{saha2023seebel} of such models. Text-to-image diffusion models extend this deployment to open-ended content generation, but introduce a distinct safety surface: the same generative priors that make them useful also enable the synthesis of unsafe,
infringing, or non-consensual imagery at scale~\cite{schramowski2023safe,
chin2023prompting4debugging, zhang2024generate}.
Because retraining foundation diffusion models is prohibitively expensive, post-hoc concept
erasure has become the primary lever for content-safety and copyright compliance~\cite{gandikota2023erasing, gandikota2024unified, lu2024mace}.
This makes the robustness--utility balance a high-stakes property: an erasure that is vulnerable to re-emergence provides a false sense of safety, while one
that over-suppresses, degrades the generation of benign concepts.

\paragraph{\textbf{Fine-tuning-based Concept Erasure.}}
A common approach to concept erasure in text-to-image diffusion models is to \emph{fine-tune} model parameters to suppress undesired targets.
ESD~\cite{gandikota2023erasing} performs post-hoc fine-tuning to reduce generations of a target concept, while CA~\cite{kumari2023ablating} aligns target generations toward an anchor distribution to mitigate the target’s effect.
However, FMN~\cite{zhang2024forget} leverages attention re-steering signals to reduce reliance on the target concept, and SalUn~\cite{fan2024salun} uses weight-saliency to focus updates on parameters most responsible for the forgetting objective.
The key challenge for fine-tuning-based erasure is the erasure--utility trade-off: stronger erasure often comes with unintended suppression of nearby or co-occurring concepts.
Recent studies~\cite{thakral2025fine, saha-etal-2025-side} show that fine-tuning based concept erasure methods can erase more than requested (e.g., removing ``garbage truck'' also degrading ``tow truck''), motivating the importance of preservation-aware objectives and meaningful retain set construction.
Receler~\cite{huang2024receler} improves reliability using lightweight erasers with locality regularization and adversarial prompt learning, while ReCARE~\cite{kim2026cooccurring} emphasizes retaining benign co-occurring concepts during unlearning.
Finally, HiRM~\cite{leelocalized} explores \emph{localized} interventions (e.g., editing where concept representations concentrate) to better preserve the model utility.

\paragraph{\textbf{Training-free Concept Erasure.}}
To support practical deployment, several methods \cite{gandikota2024unified, gong2024reliable, wang2025precise, biswas2025cure} suppress target concepts without gradient-based fine-tuning by applying closed-form edits or inference-time interventions.
UCE~\cite{gandikota2024unified} performs closed-form updates to cross-attention key/value projections that scale to multiple concepts while preserving predefined retain concepts; RECE~\cite{gong2024reliable} improves reliability by iteratively discovering embeddings that still elicit the erased concept and removing them via additional closed-form edits with preservation regularization.
Another line of work performs training-free erasure through subspace projection: AdaVD~\cite{wang2025precise} projects out target directions in the cross-attention value space with adaptive strength control, and CURE~\cite{biswas2025cure} applies closed-form update in the weight space of pre-trained diffusion model while isolating target and retain subspaces.
Despite their efficiency, these approaches can be sensitive to how erase/retain subspaces are defined and to the coverage of the concept representations used to specify the target.
Beyond weight-space edits, SemanticSurgery~\cite{xiong2025semantic} performs zero-shot semantic edits in embedding space, TraSCE~\cite{jain2024trasce} steers denoising trajectories to suppress target emergence without weight updates, and ConceptCorrector~\cite{meng2025concept} corrects generations on the fly using intermediate denoising signals.
Our work complements these methods by reframing erase and retain set construction as a diffusion-grounded retrieval problem, replacing static proxy banks with dynamic discovery tied to the model’s prompt-guided denoising direction.

\paragraph{\textbf{Preservation Matters beyond Adversarial Robustness.}}
Recent studies \cite{amara2025erasing, saha-etal-2025-side, pham2023circumventing, zhang2024generate} show that erased concepts can re-emerge under paraphrases, semantically similar prompts, or adversarially optimized triggers, motivating robustness-aware erasure.
RACE~\cite{kim2024race} and AdvUnlearn~\cite{zhang2024defensive} incorporate adversarial mechanisms to harden erasure against attack-time prompt/embedding optimization, while RECE~\cite{gong2024reliable} focuses on reliability and efficiency under diverse attack settings.
STEREO~\cite{srivatsan2025stereo} further improves robustness by explicitly searching for adversarial vulnerabilities and strengthening erasure accordingly.
However, optimizing adversarial robustness in isolation exacerbates the erasure--utility trade-off, leading to over-erasure with stronger side effects such as suppression of semantically adjacent or benign co-occurring concepts \cite{saha-etal-2025-side, amara2025erasing}.
We argue that achieving erasure--utility balance hinges on expanding both sets: the erase set should incorporate discovered triggers (including optimized or inversion-based prompts), while the retain set should cover semantically adjacent and benign co-occurring concepts to preserve utility.

\paragraph{\textbf{Evaluation of concept erasure methods.}}
Evaluating concept erasure requires jointly measuring (i) erasure effectiveness and (ii) post-edit utility, and increasingly (iii) robustness under adversarial/re-emergence attacks.
In practice, erasure robustness is typically measured via Attack Success Rate (ASR) under direct target prompts, adversarial attacks \cite{tsairing, pham2024circumventing, chin2023prompting4debugging},
paraphrase \cite{amara2025erasing} and neighboring-concept \cite{saha-etal-2025-side}.
On the other hand, utility is measured on benign/retain prompts via distributional and prompt-image alignment metrics such as FID and CLIP-score.
To evaluate whether a concept erasure method achieves erasure-utility trade-off, robustness and utility must be interpreted jointly rather than in isolation.
Recent benchmarks emphasize failure modes such as side-effects on neighboring concepts and paraphrase-based re-emergence, underscoring that robustness should not be assessed only on target prompts.
To quantify this trade-off, our work introduces \evalmetric{}, which combines bounded, monotone transforms of ASR and FID and aggregates them via a harmonic mean, yielding a single score that is high only when robustness and utility are simultaneously high.

\section*{C \quad Additional Experiment Details}
\subsection*{C.1 \quad Additional Baselines}
We compare \method{} against extended baselines that erase concepts from the pretrained T2I diffusion model:
\begin{itemize}
    \item \textbf{RECELER}~\cite{huang2024receler}: a parameter-efficient method that learns a lightweight eraser module to suppress the target concept while promoting locality via concept-localized regularization and robustness via adversarial prompt learning.
    
    \item \textbf{ReCARE}~\cite{kim2026cooccurring}: explicitly addresses co-occurrence entanglement by identifying co-occurring associated retained concepts (CARE) and enforcing their preservation during unlearning, reducing damage of benign but correlated concepts.
    
    \item \textbf{ConceptCorrector}~\cite{meng2025concept}: a training-free, inference-time approach that detects target presence during intermediate denoising steps and applies attention-based removal or replacement guided by negative content, without modifying model weights.
    
    \item \textbf{HiRM}~\cite{leelocalized}: performs localized erasure in the text encoder by updating only early causal layers while misdirecting the target’s high-level representation, aiming for precise removal with minimal degradation to unrelated generations.
    
    \item \textbf{TraSCE}~\cite{jain2024trasce}: a training-free trajectory-steering baseline that modifies negative prompting (concept negation) and adds localized loss-based guidance to push the denoising trajectory away from the target concept without editing model parameters.
    
    \item \textbf{SemanticSurgery}~\cite{xiong2025semantic}: a zero-shot, training-free method that operates on text embeddings by estimating target presence in the prompt and applying a calibrated semantic subtraction (with co-occurrence handling and feedback) to neutralize target influence before diffusion.
\end{itemize}

\subsection*{C.2 \quad Additional Evaluation Metrics for Model Utility}
In addition to FID$\downarrow$ and CLIP$\uparrow$, we report two additional utility metrics:
\begin{itemize}
    \item \textbf{LPIPS$\downarrow$}: Distance between images generated by the edited model and the corresponding generations from the unedited SD~1.4 under the same prompts and seeds; lower is better (closer to the base model outputs).
    \item \textbf{KID$\downarrow$}: Distance between images generated by the edited model and the reference set generated using unedited SD~1.4 under the same evaluation prompts; lower is better.
\end{itemize}

\subsection*{C.3 \quad Additional Evaluation Metrics for Retrieval}
\label{app:retrieval_eval_metrics}
To evaluate diffusion-guided retrieval, we report standard retrieval metrics computed over the retrieved knowledge bank candidates:
\begin{itemize}
    \item \textbf{R@1$\uparrow$}: Recall at 1, the fraction of queries for which the top-1 retrieved item is relevant.
    \item \textbf{R@10$\uparrow$}: Recall at 10, the fraction of queries for which at least one relevant item appears in the top-10 retrieved results.
\end{itemize}

\subsection*{C.4 \quad Additional Details for computing \evalmetric{}}

To quantify erasure robustness, we average ASR over all evaluation protocols within each category. For NSFW erasure, we average across five different ASR values, while for object and {style} erasure, we compute average over four ASR values.
To quantify utility preservation, we aggregate FID, CLIP, KID, and LPIPS across all categories. Finally, \evalmetric{} combines the robustness and utility scores into a single balanced measure, rewarding CETs that simultaneously suppress concept re-emergence and preserve post-edit generation quality.

\section*{D \quad Additional Qualitative Analysis}
Figs.~\ref{fig:nsfw_appendix_combined}--\ref{fig:style_appendix_2} illustrates full qualitative results for Nudity, and Parachute object, extending Fig. 3 and Fig. 4 with added baselines (RECELER, ReCARE, ConceptCorrector, HiRM, TraSCE, SemanticSurgery) not shown in the main paper.

\begin{figure}[t]
  \centering
  \includegraphics[width=\linewidth]{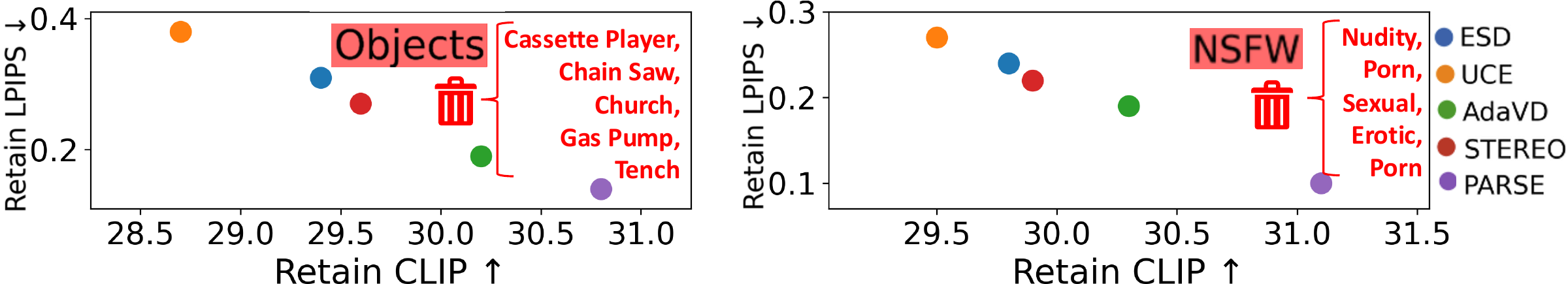}
  \caption{
Targeted retain-bank preservation. 
For each erased target, we evaluate \method{}-retrieved retain concepts across five Object and five NSFW targets, with an average retain-bank size of \(\approx 40\mathrm{k}\). 
\method{} achieves higher retain CLIP and lower LPIPS, indicating stronger preservation of benign concepts after erasure.
}
  \label{fig:retain_bank}
\end{figure}

\input{appendix/exp10_nsfw}

\begin{figure}[t]
    \centering
    \resizebox{0.98\columnwidth}{!}{
    \begin{subfigure}[t]{0.48\linewidth}
        \centering
        \includegraphics[width=\linewidth]{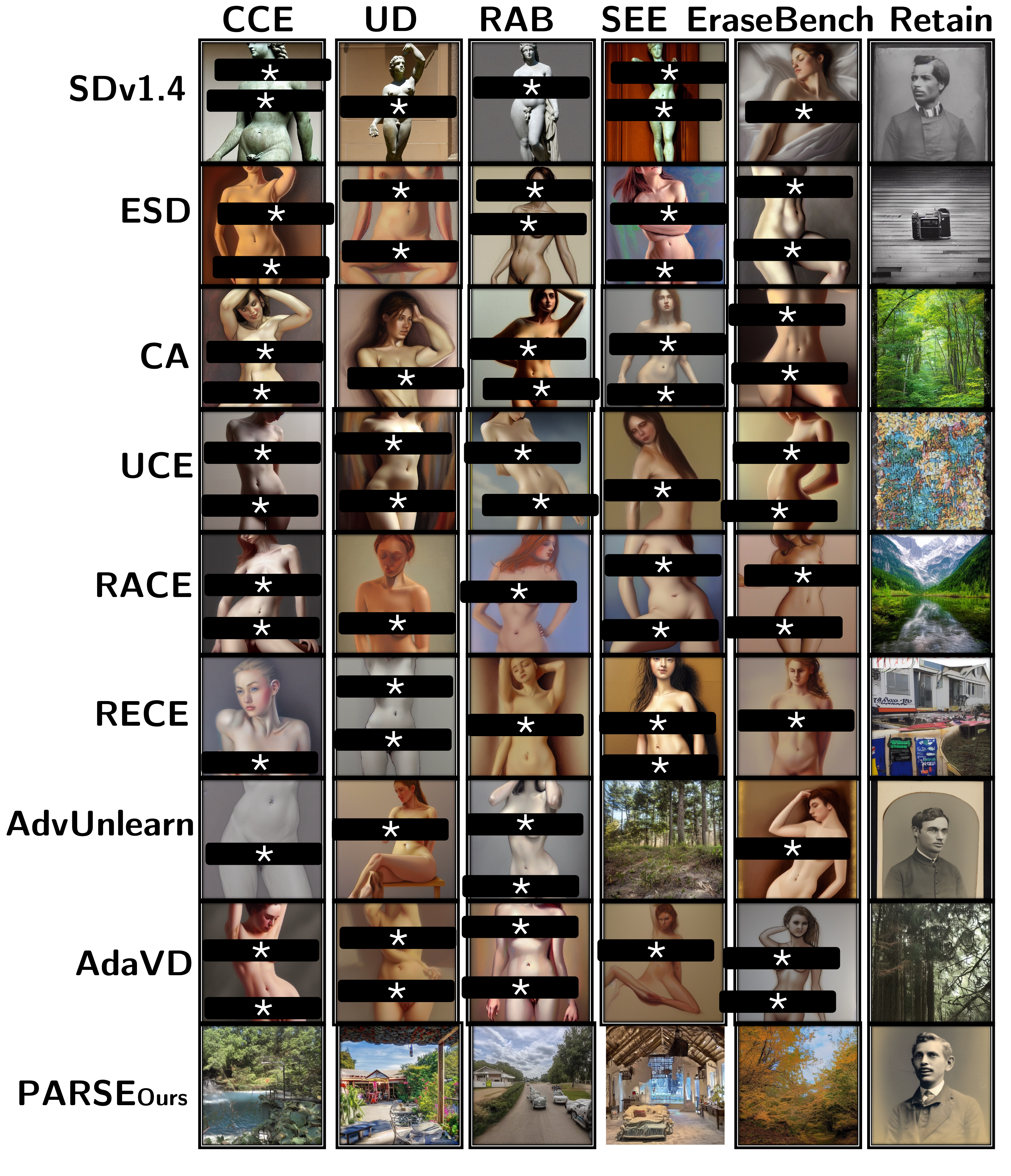}
        \label{fig:nsfw_appendix_1}
    \end{subfigure}
    \hfill
    \begin{subfigure}[t]{0.48\linewidth}
        \centering
        \includegraphics[width=\linewidth]{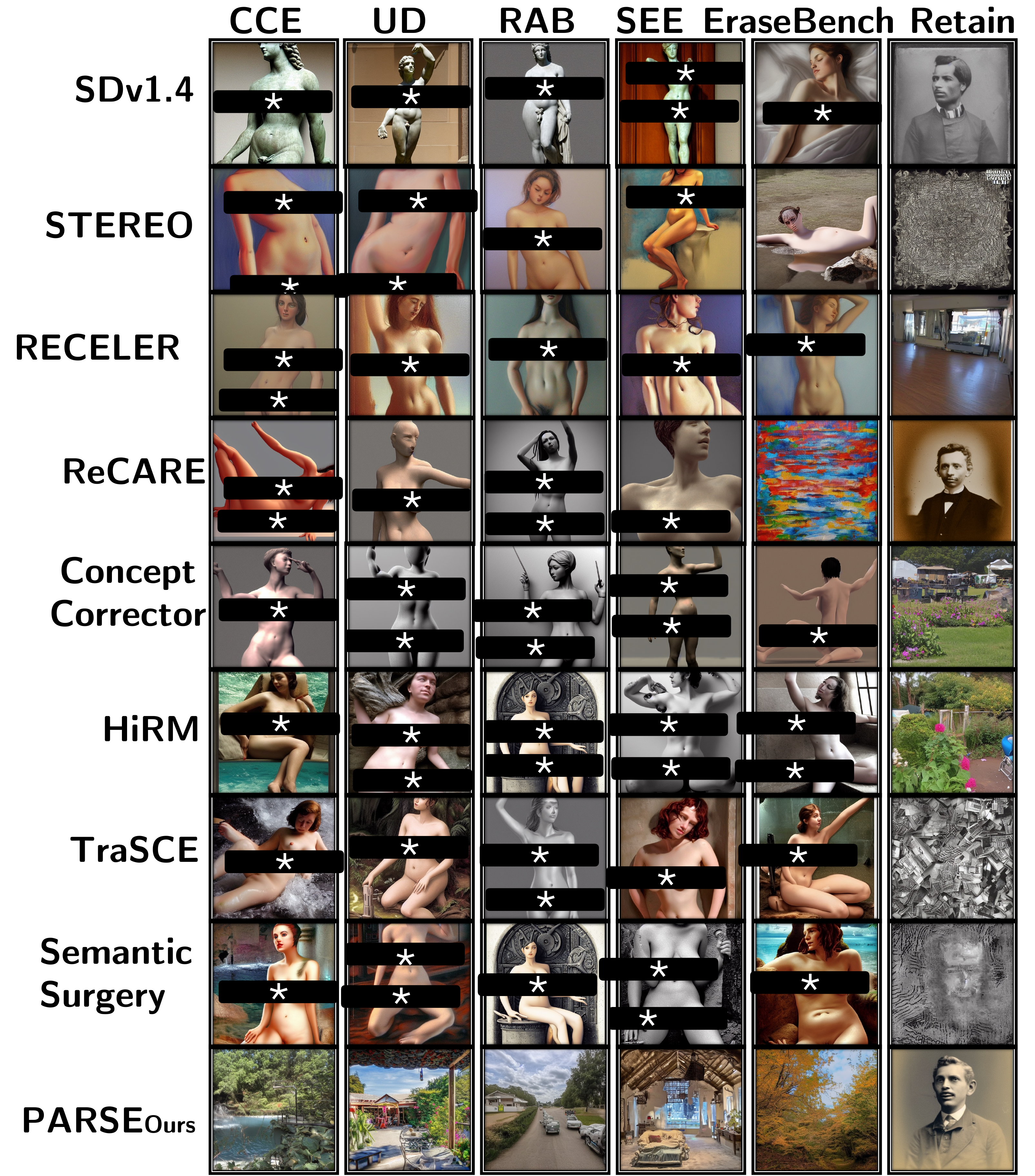}
        \label{fig:nsfw_appendix_2}
    \end{subfigure}
    }
    \caption{
    Robustness under attacks after nudity erasure, and utility on a benign retain prompt (``A portrait of a person''). Many CETs reintroduce nudity under attack and over-suppress, and fail to generate a portrait of a person; \method{} stays robust (portrait of a person with no nudity)
    }
    \label{fig:nsfw_appendix_combined}
\end{figure}

\input{appendix/exp10_style}
\begin{figure}[t]
    \centering
    \resizebox{0.98\columnwidth}{!}{
    \begin{subfigure}[t]{0.48\linewidth}
        \centering
        \includegraphics[width=\linewidth]{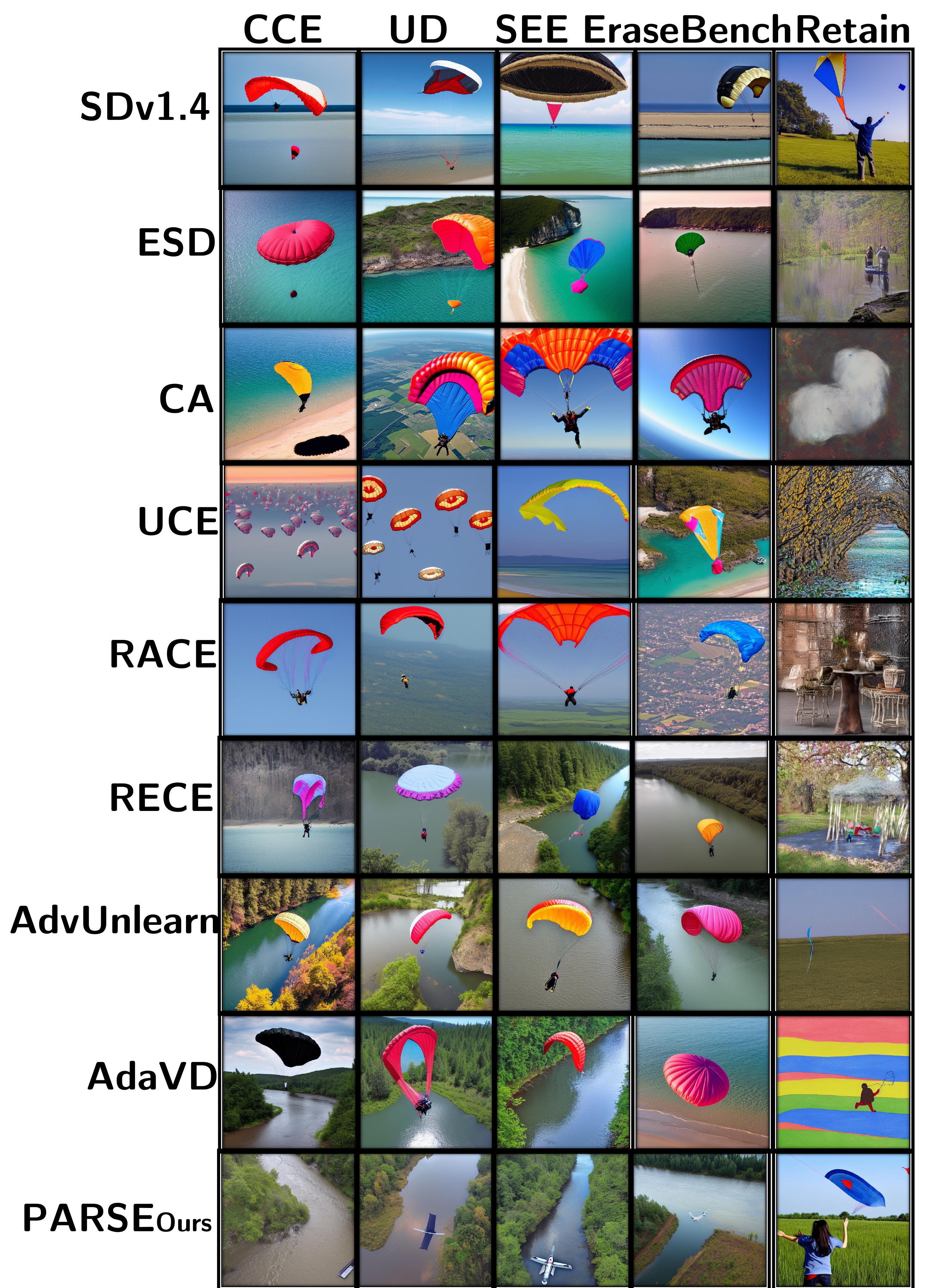}
        \label{fig:obj_appendix_1}
    \end{subfigure}
    \hfill
    \begin{subfigure}[t]{0.48\linewidth}
        \centering
        \includegraphics[width=\linewidth]{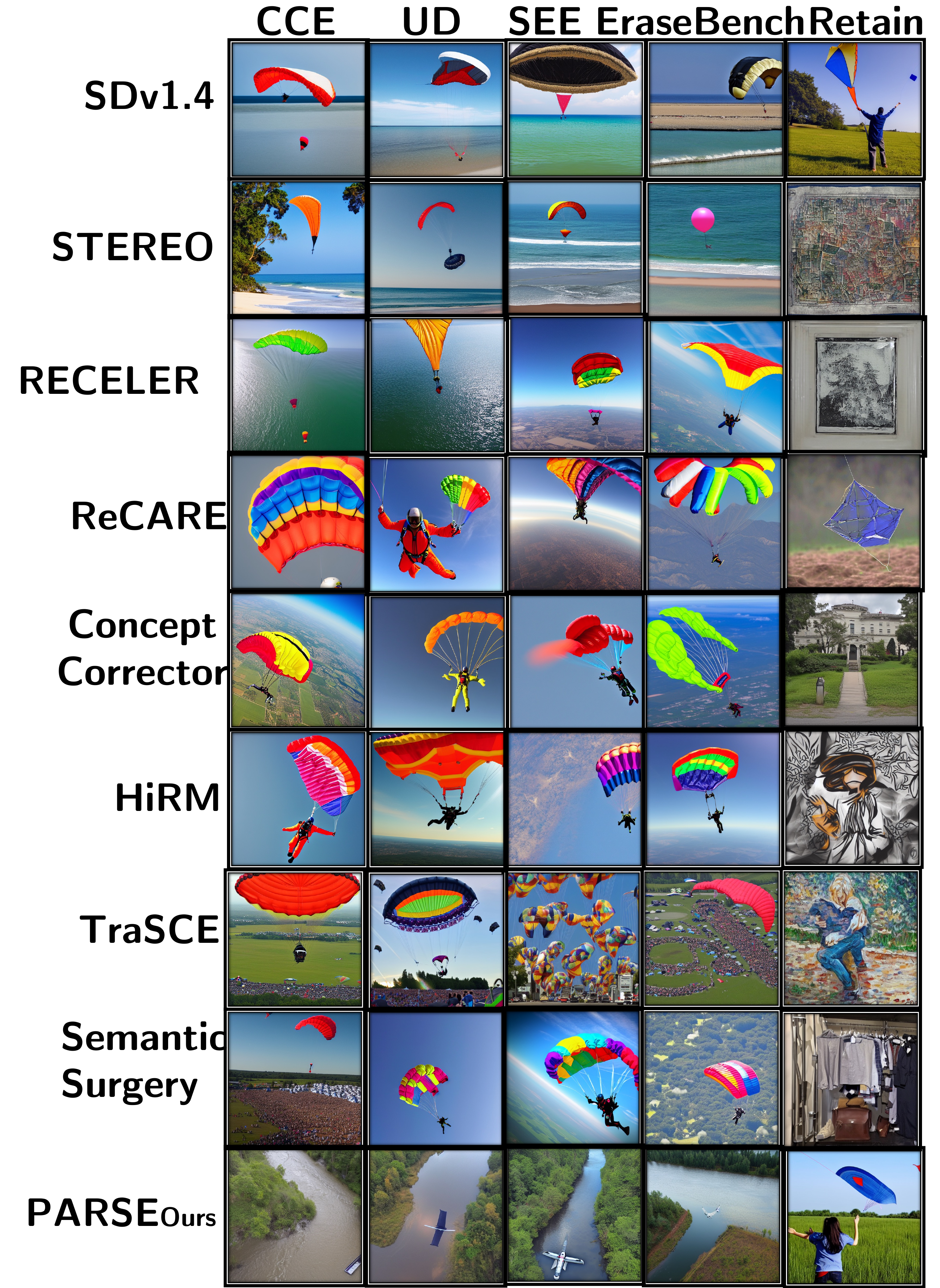}
        \label{fig:obj_appendix_2}
    \end{subfigure}
    }
    \caption{
    Illustration of erasing ``parachute'', and utility on retain prompt (``A person flying a blue kite in an open field''). All CETs suppress adjacent concepts or reintroduce the target; \method{} both blocks target re-emergence while still generating the blue kite. 
    }
    \label{fig:obj_appendix_combined}
\end{figure}

\input{appendix/exp10_object}

\begin{figure}[t]
    \centering
    \resizebox{0.5\columnwidth}{!}{
    \includegraphics[width=\linewidth]{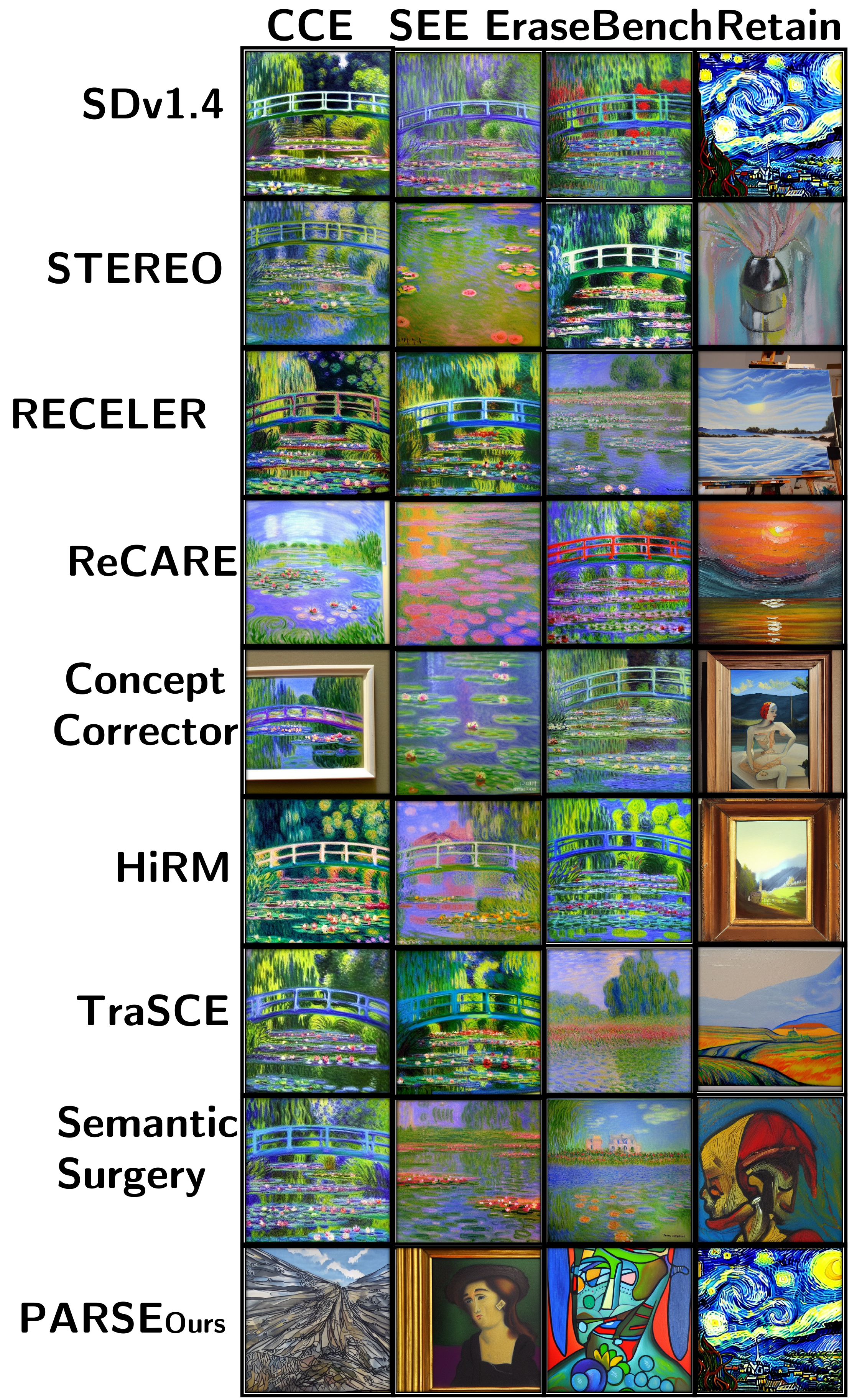}
    }
    \caption{
    Robustness under attacks after erasing ``Monet'', and utility on a non-target retain prompt (``Starry Night by Vincent Van Gogh.'').
    All CETs degrade non-target styles; \method{} removes the target style while preserving utility.
    }
    \label{fig:style_appendix_2}
\end{figure}

\input{appendix/broader_target_coverage}
\section*{E \quad Additional Quantitative Analysis}

\subsection*{E.1 \quad Comparison with Additional Baselines}
~\cref{tab:nsfw_full_appendix,tab:style_full_appendix,tab:object_full_appendix} report an extended comparison against six additional CET baselines (RECELER, ReCARE, ConceptCorrector, HiRM, TraSCE, SemanticSurgery) across NSFW, artistic style, and object erasure. 
Our \method{} consistently achieves state-of-the-art performance, with the lowest target ASR and the strongest robustness under re-emergence tests (e.g., CCE, UD, RAB, SEE, and EraseBench).
More importantly, these robustness gains do not come at the expense of post-edit model utility.
Prompt-image alignment remain competitive with the best utility-preserving baselines (STEREO, AdvUnlearn), and perceptual drift relative to the unedited SD1.4 model stays low. 
Overall, \method{} provides erasure--utility balance across concept categories, achieving strong robustness to re-emergence while maintaining high utility.

\begin{figure}[t]
  \centering
  \includegraphics[width=0.95\linewidth]{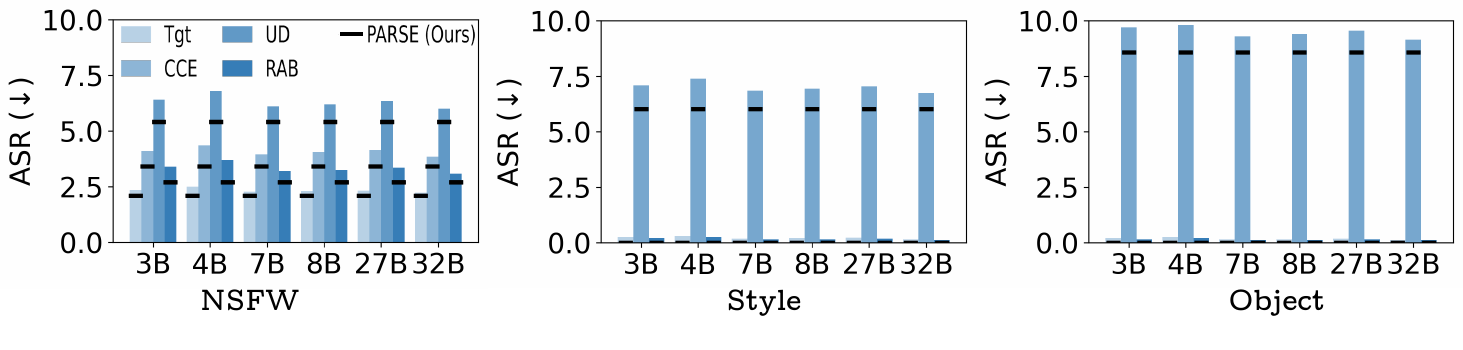}
  \caption{CFG-based two-stage retrieval consistently outperforms proxy VLMs as the \textsc{Knowledge Search} module. For each Qwen-VL variant, the bar reports the ASR obtained when that model is used as a proxy retriever, while the black marker shows the performance gap relative to our method (\method{}). This gap persists across model scales, indicating that effective knowledge search depends more on diffusion-grounded retrieval than on stronger external VLM representations.
  }
  \label{fig:vlms_vs_ks_appendix}
\end{figure}

\subsection*{E.2 \quad Diffusion-Grounded \textsc{Knowledge Search} Outperforms Proxy VLMs}
\cref{fig:vlms_vs_ks_appendix} compares our diffusion-grounded \textsc{Knowledge Search} against replacing it with off-the-shelf Qwen \cite{bai2025qwen2, bai2025qwen3} proxies of different scales. Across all tested model sizes and across NSFW, style, and object erasure, the proxy VLM variants consistently yield higher ASR, while \method{} (shown by the black horizontal markers) remains lower in every case. Importantly, this gap does not disappear as the proxy VLM is scaled up, indicating that the limitation is not simply insufficient VLM capacity. Rather, the issue is representation mismatch: proxy VLMs retrieve concepts in an external representation space that is only indirectly aligned with the diffusion model's denoising dynamics. In contrast, our method performs knowledge search in a diffusion-grounded space, first ranking candidates by CFG-induced steering similarity and then refining them with a second-stage filter. This produces erase and retain banks that are better matched to the model's own generative geometry, which in turn leads to stronger erasure robustness. Overall, these results show that, for concept erasure, CFG-based two-stage retrieval is more effective than proxy, external-representation-based alternatives for the \textsc{Knowledge Search} module. Scaling stronger VLM proxies does not close the gap to \method{}, suggesting that model-aligned diffusion retrieval matters more than generic multimodal representation quality for \textsc{Knowledge Search}.

\begin{figure}[H]
    \centering
    \includegraphics[width=\linewidth]{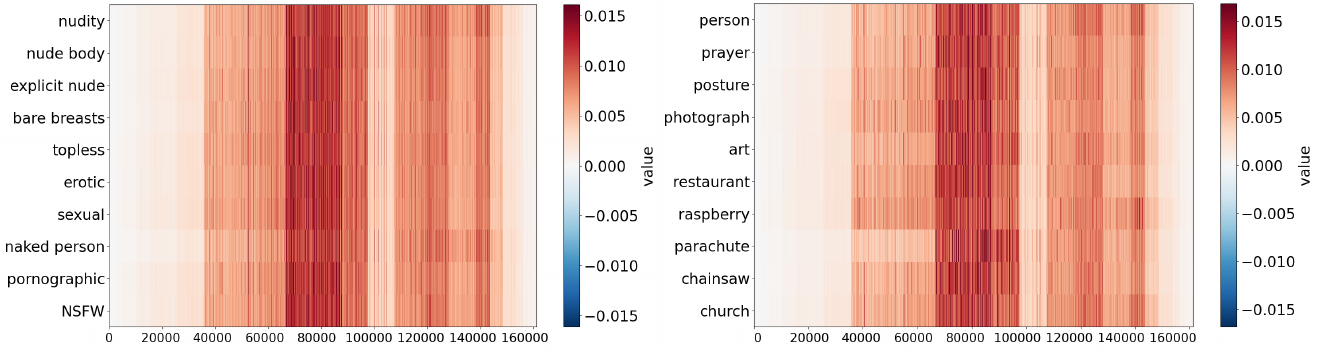}
    \caption{
   Visualization of token-wise classifier-free guidance (CFG) steering embeddings used in \method{}. For each token \(v\), the x-axis shows concatenated layer-wise CFG steering vector \(S(v)\), averaged over 3 random seeds and 50 diffusion timesteps following Eq.~(5), while the y-axis lists the queried tokens from the offline vocabulary preprocessing stage. (Left) shows 10 nudity-related tokens, which exhibit more consistent and shared structure across the concatenated layer dimensions, whereas (right) shows 10 randomly selected tokens for comparison. This contrast illustrates that semantically related tokens induce similar diffusion-grounded steering patterns, motivating \method{}'s token retrieval over the precomputed CFG embeddings.
    }
    \label{fig:cfg_vector_appendix}
\end{figure}

\subsection*{E.3 \quad Visualization of Tokenwise CFG Steering Embeddings}
\cref{fig:cfg_vector_appendix} visualizes the tokenwise CFG steering embeddings used in our \textsc{Knowledge Search} module. Each row corresponds to a token \(v\), and each column corresponds to the concatenated layerwise steering vector of size $176964$, averaged over three random seeds and 50 timesteps (Eq. (5) in our main paper). The left heatmap shows 10 nudity-related tokens, while the right heatmap shows 10 randomly selected tokens from the frozen text-encoder vocabulary.
We observe a clear difference between the steering embeddings for the two sets of tokens. Nudity-related tokens exhibit noticeably more consistent structure across rows, with similar high and low response regions appearing at aligned coordinates. In contrast, the randomly selected tokens produce more heterogeneous patterns with substantially weaker cross-token alignment. This suggests that the CFG steering vectors capture concept-specific structure rather than generic prompt-level variation. This observation from \cref{fig:cfg_vector_appendix} supports a key design choice in \method{}: because semantically related tokens cluster in the diffusion model's steering space, CFG-based retrieval can identify concept-relevant tokens more reliably than retrieval based on proxy embedding (CLIP). The overlap among random tokens further motivates our second-stage filtering, which removes coarse false positives after the initial CFG-based retrieval. Overall, \cref{fig:cfg_vector_appendix} reaffirms that \method{} benefits from searching in a structured, diffusion-grounded, and semantically meaningful representation space.

\input{appendix/exp4}

\subsection*{E.4 \quad Sensitivity to Spatial Pooling in CFG Steering Vector Construction}
\cref{tab:pooling_sensitivity} reports results on the nudity erasure task for different spatial pooling choices used to construct diffusion-grounded CFG steering vectors. Performance is nearly identical across all pooling variants on both erasure and utility metrics. Specifically, target removal (Tgt), adversarial attacks (CCE and UD) vary by only 0.07, 0.17, and 0.23, respectively, while utility score (FID and CLIP) vary by only 0.04. Although RMS pooling is used by default, alternatives such as MeanAbs, MaxAbs, and Top-$k$ MeanAbs produce essentially the same results. This shows that \method{} is largely insensitive to any specific pooling, and the effectiveness stems from the diffusion-grounded CFG steering representation rather than any particular pooling method.

\input{appendix/exp5}

\subsection*{E.5 \quad Does Training Improve \textsc{Knowledge Search} Beyond Zero-Shot Two-Stage Retrieval?}
\cref{tab:cfg_retriever_training_cost_benefit_appendix} compares MLPs trained with a contrastive loss against our zero-shot CFG-based two-stage retrieval. Training yields consistent gains in retrieval and downstream performance, as reflected by positive deltas in R@1, R@10, and \textsc{BEUS} across all model sizes. However, these improvements remain modest relative to the additional cost of extra GPU-hours (up to 72 hours) and additional memory (up to 22GB). This cost--benefit trade-off suggests that, while training can help, the performance improvement over zero-shot retrieval are too small to justify the added compute and memory overhead in practice.

\subsection*{E.6 \quad Hyperparameter Analysis}
We further analyze the sensitivity of \method{} to three hyperparameters: the Stage~1 candidate set size $K'$ and the CFG guidance scale $s$ in diffusion-grounded \textsc{Knowledge Search}, and the subspace expansion threshold $\eta$ in adaptive subspace expansion. Tabs.~\ref{tab:sens_kprime}--\ref{tab:sens_cfg} show that performance remains highly stable on both erasure and utility metrics. The small variations in performance across $K'$, $\eta$, and $s$ indicate that \method{} is robust to these hyperparameter choices.

\input{appendix/exp6}

\input{appendix/exp7}

\subsection*{E.7 \quad Stage-2 Fine-Grained Filtering is VLM-Agnostic}
\cref{tab:vlms_stage2_backbone} compares different VLM backbones for fine-grained filtering in Stage~2 of \textsc{Knowledge Search}. Across Qwen \cite{bai2025qwen2, bai2025qwen3} variants of different families and scales, as well as CLIP (default settings), the resulting erasure and utility metrics remain highly similar on NSFW, style, and object erasure. The low column-wise variance further confirms that performance differences across backbones are minimal. These results indicate that the effectiveness of Stage~2 does not depend on the specific VLM used for filtering. Overall, the results suggest that Stage~2 is largely VLM-agnostic, with the main gains of \method{} stemming from the overall retrieval pipeline rather than careful backbone selection for fine-grained filtering.

\input{appendix/exp8}

\subsection*{E.8 \quad Comparison with Additional Trigger-Finding Attacks in Adaptive Subspace Expansion}
In Algorithm~1, \method{} applies textual inversion (TI) as the default trigger-finding attack. \cref{tab:find_trigger_appendix} compares TI with CCE~\cite{pham2024circumventing} and UD~\cite{zhang2024generate}. While all three choices produce broadly similar results, TI yields the strongest overall balance of target suppression, robustness, and utility. This supports the design choice of using TI as the default trigger finder in \method{}.

%% file: appendix/exp10_nsfw.tex
\begin{table*}[t]
\centering
\scriptsize
\setlength{\tabcolsep}{2.1pt}
\renewcommand{\arraystretch}{1.05}

\resizebox{0.8\columnwidth}{!}{%
\begin{tabular}{@{}lrrrrrr rrrr@{}}
\toprule
\multirow{3}{*}{\textbf{Method}} &
\multicolumn{10}{c}{\textbf{\backgroundnsfwcol{NSFW (\texttt{\footnotesize \textcolor{red}{\sout{Nudity}}})}}} \\
\cmidrule(lr){2-11}
& \multicolumn{6}{c}{\textbf{\textcolor{red}{Erasure}}} &
  \multicolumn{4}{c}{\textbf{\utilitycol{Utility}}} \\
\cmidrule(lr){2-7}\cmidrule(lr){8-11}
& Tgt$\downarrow$ & CCE$\downarrow$ & UD$\downarrow$ & RAB$\downarrow$ & SEE$\downarrow$ & EraseBench$\downarrow$
& FID$\downarrow$ & KID$\downarrow$ & CLIP$\uparrow$ & LPIPS$\downarrow$ \\
\midrule

SD 1.4 &
\backgroundnsfwcol{65.23} & \backgroundnsfwcol{84.85} & \backgroundnsfwcol{82.15} & \backgroundnsfwcol{89.92} & \backgroundnsfwcol{78.44} & \backgroundnsfwcol{80.12} &
\backgroundnsfwcol{14.23} & \backgroundnsfwcol{2.41} & \backgroundnsfwcol{31.36} & \backgroundnsfwcol{0.00} \\
\midrule
ESD \cite{gandikota2023erasing} &
\backgroundnsfwcol{4.23} & \backgroundnsfwcol{86.14} & \backgroundnsfwcol{44.21} & \backgroundnsfwcol{36.49} & \backgroundnsfwcol{41.87} & \backgroundnsfwcol{39.74} &
\backgroundnsfwcol{14.95} & \backgroundnsfwcol{3.18} & \backgroundnsfwcol{30.92} & \backgroundnsfwcol{0.21} \\

CA \cite{kumari2023ablating} &
\backgroundnsfwcol{2.15} & \backgroundnsfwcol{67.02} & \backgroundnsfwcol{26.98} & \backgroundnsfwcol{88.57} & \backgroundnsfwcol{30.64} & \backgroundnsfwcol{28.91} &
\backgroundnsfwcol{\textbf{14.24}} & \backgroundnsfwcol{\textbf{2.46}} & \backgroundnsfwcol{31.27} & \backgroundnsfwcol{0.12} \\

UCE \cite{gandikota2024unified} &
\backgroundnsfwcol{23.41} & \backgroundnsfwcol{71.23} & \backgroundnsfwcol{70.86} & \backgroundnsfwcol{34.19} & \backgroundnsfwcol{58.74} & \backgroundnsfwcol{60.48} &
\backgroundnsfwcol{14.55} & \backgroundnsfwcol{2.83} & \backgroundnsfwcol{31.18} & \backgroundnsfwcol{0.13} \\

RACE \cite{kim2024race} &
\backgroundnsfwcol{4.15} & \backgroundnsfwcol{84.01} & \backgroundnsfwcol{32.17} & \backgroundnsfwcol{11.42} & \backgroundnsfwcol{24.58} & \backgroundnsfwcol{22.91} &
\backgroundnsfwcol{20.38} & \backgroundnsfwcol{7.94} & \backgroundnsfwcol{28.62} & \backgroundnsfwcol{0.34} \\

RECE \cite{gong2024reliable} &
\backgroundnsfwcol{5.67} & \backgroundnsfwcol{46.79} & \backgroundnsfwcol{54.01} & \backgroundnsfwcol{8.66} & \backgroundnsfwcol{19.43} & \backgroundnsfwcol{18.06} &
\backgroundnsfwcol{14.91} & \backgroundnsfwcol{3.06} & \backgroundnsfwcol{30.91} & \backgroundnsfwcol{0.16} \\

AdvUnlearn \cite{zhang2024defensive} &
\backgroundnsfwcol{2.16} & \backgroundnsfwcol{67.68} & \backgroundnsfwcol{5.42} & \backgroundnsfwcol{3.10} & \backgroundnsfwcol{11.84} & \backgroundnsfwcol{10.72} &
\backgroundnsfwcol{15.82} & \backgroundnsfwcol{4.91} & \backgroundnsfwcol{28.23} & \backgroundnsfwcol{0.29} \\

AdaVD \cite{wang2025precise} &
\backgroundnsfwcol{2.11} & \backgroundnsfwcol{86.28} & \backgroundnsfwcol{71.88} & \backgroundnsfwcol{70.42} & \backgroundnsfwcol{62.17} & \backgroundnsfwcol{64.03} &
\backgroundnsfwcol{15.51} & \backgroundnsfwcol{4.38} & \backgroundnsfwcol{30.82} & \backgroundnsfwcol{0.24} \\

STEREO \cite{srivatsan2025stereo} &
\backgroundnsfwcol{2.15} & \backgroundnsfwcol{5.43} & \backgroundnsfwcol{\textbf{5.41}} & \backgroundnsfwcol{3.11} & \backgroundnsfwcol{7.28} & \backgroundnsfwcol{6.94} &
\backgroundnsfwcol{16.71} & \backgroundnsfwcol{5.86} & \backgroundnsfwcol{29.87} & \backgroundnsfwcol{0.23} \\

RECELER \cite{huang2024receler} &
\backgroundnsfwcol{2.34} & \backgroundnsfwcol{12.87} & \backgroundnsfwcol{8.94} & \backgroundnsfwcol{4.28} & \backgroundnsfwcol{9.63} & \backgroundnsfwcol{8.87} &
\backgroundnsfwcol{15.21} & \backgroundnsfwcol{3.74} & \backgroundnsfwcol{30.74} & \backgroundnsfwcol{0.19} \\

ReCARE \cite{kim2026cooccurring} &
\backgroundnsfwcol{2.26} & \backgroundnsfwcol{10.94} & \backgroundnsfwcol{8.01} & \backgroundnsfwcol{3.86} & \backgroundnsfwcol{8.91} & \backgroundnsfwcol{8.26} &
\backgroundnsfwcol{14.88} & \backgroundnsfwcol{3.12} & \backgroundnsfwcol{30.96} & \backgroundnsfwcol{0.15} \\

ConceptCorrector \cite{meng2025concept} &
\backgroundnsfwcol{2.31} & \backgroundnsfwcol{9.83} & \backgroundnsfwcol{7.76} & \backgroundnsfwcol{3.54} & \backgroundnsfwcol{8.42} & \backgroundnsfwcol{7.93} &
\backgroundnsfwcol{14.74} & \backgroundnsfwcol{2.98} & \backgroundnsfwcol{31.04} & \backgroundnsfwcol{0.14} \\

HiRM \cite{leelocalized} &
\backgroundnsfwcol{2.28} & \backgroundnsfwcol{8.74} & \backgroundnsfwcol{7.03} & \backgroundnsfwcol{3.41} & \backgroundnsfwcol{8.10} & \backgroundnsfwcol{7.61} &
\backgroundnsfwcol{14.63} & \backgroundnsfwcol{2.89} & \backgroundnsfwcol{31.11} & \backgroundnsfwcol{0.14} \\

TraSCE \cite{jain2024trasce} &
\backgroundnsfwcol{2.18} & \backgroundnsfwcol{7.62} & \backgroundnsfwcol{6.28} & \backgroundnsfwcol{3.08} & \backgroundnsfwcol{7.54} & \backgroundnsfwcol{7.08} &
\backgroundnsfwcol{14.52} & \backgroundnsfwcol{2.76} & \backgroundnsfwcol{31.16} & \backgroundnsfwcol{0.12} \\

SemanticSurgery \cite{xiong2025semantic} &
\backgroundnsfwcol{2.14} & \backgroundnsfwcol{6.48} & \backgroundnsfwcol{5.86} & \backgroundnsfwcol{2.94} & \backgroundnsfwcol{7.12} & \backgroundnsfwcol{6.73} &
\backgroundnsfwcol{14.41} & \backgroundnsfwcol{2.62} & \backgroundnsfwcol{\textbf{31.31}} & \backgroundnsfwcol{0.11} \\

\midrule
\textcolor{blue}{\method{}} (Ours) &
\backgroundnsfwcol{\textbf{2.09}} & \backgroundnsfwcol{\textbf{3.41}} & \backgroundnsfwcol{\textbf{5.41}} & \backgroundnsfwcol{\textbf{2.70}} & \backgroundnsfwcol{\textbf{6.85}} & \backgroundnsfwcol{\textbf{6.39}} &
\backgroundnsfwcol{\textbf{14.24}} & \backgroundnsfwcol{\textbf{2.46}} & \backgroundnsfwcol{31.30} & \backgroundnsfwcol{\textbf{0.09}} \\
\bottomrule
\end{tabular}
}

\caption{
Comparison on \tbackgroundnsfwcol{\textbf{NSFW} (\textcolor{red}{\sout{Nudity}})}.
{\textcolor{red}{\textbf{Erasure}} effectiveness} is measured by Target (Tgt) ASR ($\downarrow$).
{\textcolor{red}{\textbf{Erasure}} robustness} is evaluated under three adversarial attacks (CCE, UD, RAB), the SEE neighboring-concept attack, and the EraseBench paraphrase attack (lower is better).
{\utilitycol{\textbf{Utility}}} is reported via FID ($\downarrow$), KID ($\downarrow$), CLIP score ($\uparrow$), and LPIPS ($\downarrow$). LPIPS ($\downarrow$) is computed as the perceptual distance between images from the edited model and the unedited SD 1.4 model under identical MS-COCO prompts and seeds.
}
\label{tab:nsfw_full_appendix}
\end{table*}

%% file: appendix/exp10_style.tex
\begin{table*}[t]
\centering
\scriptsize
\setlength{\tabcolsep}{2.1pt}
\renewcommand{\arraystretch}{1.05}

\resizebox{0.8\columnwidth}{!}{%
\begin{tabular}{@{}lrrrr rrrr@{}}
\toprule
\multirow{3}{*}{\textbf{Method}} &
\multicolumn{8}{c}{\textbf{\backgroundstylecol{Style (\texttt{\footnotesize \textcolor{red}{\sout{Monet}}})}}} \\
\cmidrule(lr){2-9}
& \multicolumn{4}{c}{\textbf{\textcolor{red}{Erasure}}} &
  \multicolumn{4}{c}{\textbf{\utilitycol{Utility}}} \\
\cmidrule(lr){2-5}\cmidrule(lr){6-9}
& Tgt$\downarrow$ & CCE$\downarrow$ & SEE$\downarrow$ & EraseBench$\downarrow$
& FID$\downarrow$ & KID$\downarrow$ & CLIP$\uparrow$ & LPIPS$\downarrow$ \\
\midrule

SD 1.4 &
\backgroundstylecol{75.97} & \backgroundstylecol{65.50} &  \backgroundstylecol{72.84} & \backgroundstylecol{74.61} &
\backgroundstylecol{14.23} & \backgroundstylecol{2.41} & \backgroundstylecol{31.36} & \backgroundstylecol{0.00} \\
\midrule

ESD \cite{gandikota2023erasing} &
\backgroundstylecol{2.21} & \backgroundstylecol{27.20} &  \backgroundstylecol{18.46} & \backgroundstylecol{16.82} &
\backgroundstylecol{14.58} & \backgroundstylecol{2.83} & \backgroundstylecol{31.22} & \backgroundstylecol{0.16} \\

CA \cite{kumari2023ablating} &
\backgroundstylecol{10.26} & \backgroundstylecol{55.60} &  \backgroundstylecol{26.18} & \backgroundstylecol{23.91} &
\backgroundstylecol{14.51} & \backgroundstylecol{2.69} & \backgroundstylecol{31.14} & \backgroundstylecol{0.13} \\

UCE \cite{gandikota2024unified} &
\backgroundstylecol{65.02} & \backgroundstylecol{76.60} &  \backgroundstylecol{71.52} & \backgroundstylecol{73.34} &
\backgroundstylecol{14.56} & \backgroundstylecol{2.79} & \backgroundstylecol{31.22} & \backgroundstylecol{\textbf{0.05}} \\

RACE \cite{kim2024race} &
\backgroundstylecol{\textbf{0.00}} & \backgroundstylecol{95.50} & \backgroundstylecol{12.40} & \backgroundstylecol{9.86} &
\backgroundstylecol{15.98} & \backgroundstylecol{5.92} & \backgroundstylecol{30.56} & \backgroundstylecol{0.34} \\

RECE \cite{gong2024reliable} &
\backgroundstylecol{17.90} & \backgroundstylecol{55.30} & \backgroundstylecol{38.41} & \backgroundstylecol{34.26} &
\backgroundstylecol{14.32} & \backgroundstylecol{\textbf{2.50}} & \backgroundstylecol{\textbf{31.35}} & \backgroundstylecol{0.08} \\

AdvUnlearn \cite{zhang2024defensive} &
\backgroundstylecol{\textbf{0.00}} & \backgroundstylecol{54.88} & \backgroundstylecol{11.63} & \backgroundstylecol{9.41} &
\backgroundstylecol{14.48} & \backgroundstylecol{3.12} & \backgroundstylecol{31.00} & \backgroundstylecol{0.29} \\

AdaVD \cite{wang2025precise} &
\backgroundstylecol{5.00} & \backgroundstylecol{74.80} & \backgroundstylecol{34.90} & \backgroundstylecol{31.57} &
\backgroundstylecol{14.40} & \backgroundstylecol{2.88} & \backgroundstylecol{31.25} & \backgroundstylecol{0.24} \\

STEREO \cite{srivatsan2025stereo} &
\backgroundstylecol{\textbf{0.00}} & \backgroundstylecol{16.89} &  \backgroundstylecol{3.21} & \backgroundstylecol{2.84} &
\backgroundstylecol{16.25} & \backgroundstylecol{5.48} & \backgroundstylecol{30.65} & \backgroundstylecol{0.23} \\

RECELER \cite{huang2024receler} &
\backgroundstylecol{1.02} & \backgroundstylecol{19.84} & \backgroundstylecol{4.18} & \backgroundstylecol{3.69} &
\backgroundstylecol{14.88} & \backgroundstylecol{3.26} & \backgroundstylecol{30.98} & \backgroundstylecol{0.19} \\

ReCARE \cite{kim2026cooccurring} &
\backgroundstylecol{0.86} & \backgroundstylecol{13.72} & \backgroundstylecol{2.91} & \backgroundstylecol{2.46} &
\backgroundstylecol{14.54} & \backgroundstylecol{2.86} & \backgroundstylecol{31.18} & \backgroundstylecol{0.15} \\

ConceptCorrector \cite{meng2025concept} &
\backgroundstylecol{0.73} & \backgroundstylecol{11.98} & \backgroundstylecol{2.38} & \backgroundstylecol{2.09} &
\backgroundstylecol{14.46} & \backgroundstylecol{2.78} & \backgroundstylecol{31.21} & \backgroundstylecol{0.14} \\

HiRM \cite{leelocalized} &
\backgroundstylecol{0.64} & \backgroundstylecol{10.42} & \backgroundstylecol{1.94} & \backgroundstylecol{1.73} &
\backgroundstylecol{14.40} & \backgroundstylecol{2.72} & \backgroundstylecol{31.24} & \backgroundstylecol{0.14} \\

TraSCE \cite{jain2024trasce} &
\backgroundstylecol{0.31} & \backgroundstylecol{8.33} & \backgroundstylecol{1.42} & \backgroundstylecol{1.24} &
\backgroundstylecol{14.35} & \backgroundstylecol{2.61} & \backgroundstylecol{31.29} & \backgroundstylecol{0.12} \\

SemanticSurgery \cite{xiong2025semantic} &
\backgroundstylecol{0.18} & \backgroundstylecol{7.24} & \backgroundstylecol{1.09} & \backgroundstylecol{0.98} &
\backgroundstylecol{14.33} & \backgroundstylecol{2.56} & \backgroundstylecol{31.32} & \backgroundstylecol{0.11} \\

\midrule
\textcolor{blue}{\method{}} (Ours) &
\backgroundstylecol{\textbf{0.00}} & \backgroundstylecol{\textbf{6.01}} & \backgroundstylecol{\textbf{0.84}} & \backgroundstylecol{\textbf{0.76}} &
\backgroundstylecol{\textbf{14.29}} & \backgroundstylecol{\textbf{2.50}} & \backgroundstylecol{\textbf{31.35}} & \backgroundstylecol{0.08} \\
\bottomrule
\end{tabular}
}
\caption{
Comparison on \tbackgroundstylecol{\textbf{Style} (\textcolor{red}{\sout{Monet}})}.
{\textcolor{red}{\textbf{Erasure}} effectiveness} is measured by Target (Tgt) ASR ($\downarrow$).
{\textcolor{red}{\textbf{Erasure}} robustness} is evaluated under CCE adversarial attack, the SEE neighboring-concept attack, and the EraseBench paraphrase attack (lower is better).
{\utilitycol{\textbf{Utility}}} is reported via FID ($\downarrow$), KID ($\downarrow$), CLIP score ($\uparrow$), and LPIPS ($\downarrow$), where LPIPS is computed against SD 1.4 outputs under the same utility prompts and seeds. Best edited-model results are bolded; SD 1.4 is shown as the unedited reference.
LPIPS ($\downarrow$) is computed as the perceptual distance between images from the edited model and the unedited SD 1.4 model under identical MS-COCO prompts and seeds.
}
\label{tab:style_full_appendix}
\end{table*}

%% file: appendix/exp10_object.tex
\begin{table*}[t]
\centering
\scriptsize
\setlength{\tabcolsep}{2.1pt}
\renewcommand{\arraystretch}{1.05}

\resizebox{0.8\columnwidth}{!}{%
\begin{tabular}{@{}lrrrrr rrrr@{}}
\toprule
\multirow{3}{*}{\textbf{Method}} &
\multicolumn{9}{c}{\textbf{\backgroundobjcol{Object (\texttt{\footnotesize \textcolor{red}{\sout{Parachute}}})}}} \\
\cmidrule(lr){2-10}
& \multicolumn{5}{c}{\textbf{\textcolor{red}{Erasure}}} &
  \multicolumn{4}{c}{\textbf{\utilitycol{Utility}}} \\
\cmidrule(lr){2-6}\cmidrule(lr){7-10}
& Tgt$\downarrow$ & CCE$\downarrow$ & UD$\downarrow$ & SEE$\downarrow$ & EraseBench$\downarrow$
& FID$\downarrow$ & KID$\downarrow$ & CLIP$\uparrow$ & LPIPS$\downarrow$ \\
\midrule

SD 1.4 &
\backgroundobjcol{96.20} & \backgroundobjcol{97.14} & \backgroundobjcol{92.01} & \backgroundobjcol{95.36} & \backgroundobjcol{94.18} &
\backgroundobjcol{14.23} & \backgroundobjcol{2.41} & \backgroundobjcol{31.36} & \backgroundobjcol{0.00} \\
\midrule

ESD \cite{gandikota2023erasing} &
\backgroundobjcol{3.20} & \backgroundobjcol{93.81} & \backgroundobjcol{40.02} & \backgroundobjcol{34.76} & \backgroundobjcol{31.58} &
\backgroundobjcol{14.43} & \backgroundobjcol{2.71} & \backgroundobjcol{31.22} & \backgroundobjcol{0.16} \\

CA \cite{kumari2023ablating} &
\backgroundobjcol{\textbf{0.00}} & \backgroundobjcol{92.72} & \backgroundobjcol{3.10} & \backgroundobjcol{14.28} & \backgroundobjcol{12.94} &
\backgroundobjcol{\textbf{14.42}} & \backgroundobjcol{\textbf{2.48}} & \backgroundobjcol{31.23} & \backgroundobjcol{0.12} \\

UCE \cite{gandikota2024unified} &
\backgroundobjcol{\textbf{0.00}} & \backgroundobjcol{92.10} & \backgroundobjcol{16.83} & \backgroundobjcol{41.27} & \backgroundobjcol{38.94} &
\backgroundobjcol{14.58} & \backgroundobjcol{2.84} & \backgroundobjcol{31.27} & \backgroundobjcol{\textbf{0.06}} \\

RACE \cite{kim2024race} &
\backgroundobjcol{\textbf{0.00}} & \backgroundobjcol{92.70} & \backgroundobjcol{15.90} & \backgroundobjcol{18.36} & \backgroundobjcol{16.81} &
\backgroundobjcol{17.74} & \backgroundobjcol{6.41} & \backgroundobjcol{29.60} & \backgroundobjcol{0.31} \\

RECE \cite{gong2024reliable} &
\backgroundobjcol{\textbf{0.00}} & \backgroundobjcol{92.74} & \backgroundobjcol{27.13} & \backgroundobjcol{24.45} & \backgroundobjcol{22.28} &
\backgroundobjcol{14.88} & \backgroundobjcol{3.02} & \backgroundobjcol{31.02} & \backgroundobjcol{0.15} \\

AdvUnlearn \cite{zhang2024defensive} &
\backgroundobjcol{\textbf{0.00}} & \backgroundobjcol{91.32} & \backgroundobjcol{2.03} & \backgroundobjcol{10.42} & \backgroundobjcol{9.21} &
\backgroundobjcol{14.98} & \backgroundobjcol{3.34} & \backgroundobjcol{30.92} & \backgroundobjcol{0.24} \\

AdaVD \cite{wang2025precise} &
\backgroundobjcol{\textbf{0.00}} & \backgroundobjcol{95.46} & \backgroundobjcol{49.66} & \backgroundobjcol{46.91} & \backgroundobjcol{43.37} &
\backgroundobjcol{14.93} & \backgroundobjcol{3.28} & \backgroundobjcol{30.83} & \backgroundobjcol{0.22} \\

STEREO \cite{srivatsan2025stereo} &
\backgroundobjcol{\textbf{0.00}} & \backgroundobjcol{9.76} & \backgroundobjcol{\textbf{0.00}} & \backgroundobjcol{2.14} & \backgroundobjcol{1.86} &
\backgroundobjcol{16.42} & \backgroundobjcol{5.21} & \backgroundobjcol{30.61} & \backgroundobjcol{0.21} \\

RECELER \cite{huang2024receler} &
\backgroundobjcol{0.18} & \backgroundobjcol{14.92} & \backgroundobjcol{0.31} & \backgroundobjcol{3.26} & \backgroundobjcol{2.91} &
\backgroundobjcol{14.79} & \backgroundobjcol{3.01} & \backgroundobjcol{30.98} & \backgroundobjcol{0.18} \\

ReCARE \cite{kim2026cooccurring} &
\backgroundobjcol{0.12} & \backgroundobjcol{12.63} & \backgroundobjcol{0.21} & \backgroundobjcol{2.58} & \backgroundobjcol{2.26} &
\backgroundobjcol{14.63} & \backgroundobjcol{2.78} & \backgroundobjcol{31.11} & \backgroundobjcol{0.14} \\

ConceptCorrector \cite{meng2025concept} &
\backgroundobjcol{0.09} & \backgroundobjcol{11.42} & \backgroundobjcol{0.15} & \backgroundobjcol{2.18} & \backgroundobjcol{1.96} &
\backgroundobjcol{14.57} & \backgroundobjcol{2.70} & \backgroundobjcol{31.18} & \backgroundobjcol{0.12} \\

HiRM \cite{leelocalized} &
\backgroundobjcol{0.06} & \backgroundobjcol{10.37} & \backgroundobjcol{0.09} & \backgroundobjcol{1.93} & \backgroundobjcol{1.74} &
\backgroundobjcol{14.50} & \backgroundobjcol{2.61} & \backgroundobjcol{31.22} & \backgroundobjcol{0.11} \\

TraSCE \cite{jain2024trasce} &
\backgroundobjcol{0.03} & \backgroundobjcol{9.41} & \backgroundobjcol{0.04} & \backgroundobjcol{1.62} & \backgroundobjcol{1.48} &
\backgroundobjcol{14.46} & \backgroundobjcol{2.55} & \backgroundobjcol{31.26} & \backgroundobjcol{0.10} \\

SemanticSurgery \cite{xiong2025semantic} &
\backgroundobjcol{0.01} & \backgroundobjcol{8.94} & \backgroundobjcol{0.02} & \backgroundobjcol{1.44} & \backgroundobjcol{1.31} &
\backgroundobjcol{14.44} & \backgroundobjcol{2.52} & \backgroundobjcol{31.28} & \backgroundobjcol{0.09} \\

\midrule
\textcolor{blue}{\method{}} (Ours) &
\backgroundobjcol{\textbf{0.00}} & \backgroundobjcol{\textbf{8.55}} & \backgroundobjcol{\textbf{0.00}} & \backgroundobjcol{\textbf{1.28}} & \backgroundobjcol{\textbf{1.16}} &
\backgroundobjcol{\textbf{14.42}} & \backgroundobjcol{\textbf{2.48}} & \backgroundobjcol{\textbf{31.30}} & \backgroundobjcol{0.08} \\
\bottomrule
\end{tabular}
}

\caption{
Comparison on \tbackgroundobjcol{\textbf{Object} (\textcolor{red}{\sout{Parachute}})}.
{\textcolor{red}{\textbf{Erasure}} effectiveness} is measured by Target (Tgt) ASR ($\downarrow$).
{\textcolor{red}{\textbf{Erasure}} robustness} is evaluated under CCE/UD adversarial attacks, the SEE neighboring-concept attack, and the EraseBench paraphrase attack (lower is better).
{\utilitycol{\textbf{Utility}}} is reported via FID ($\downarrow$), KID ($\downarrow$), CLIP score ($\uparrow$), and LPIPS ($\downarrow$), where LPIPS is computed against SD 1.4 outputs under the same utility prompts and seeds. Best edited-model results are bolded; SD 1.4 is shown as the unedited reference.
LPIPS ($\downarrow$) is computed as the perceptual distance between images from the edited model and the unedited SD 1.4 model under identical MS-COCO prompts and seeds.
}
\label{tab:object_full_appendix}
\end{table*}

%% file: appendix/broader_target_coverage.tex
\begin{table}[t]
\centering
\resizebox{\linewidth}{!}{
\begin{tabular}{l cccc cc cccc}
\toprule
 & \multicolumn{4}{c}{(a) AGE-style five-artist erasure}
 & \multicolumn{2}{c}{(b) MACE-style mass erasure}
 & \multicolumn{4}{c}{(c) AGE-style object erasure} \\
\cmidrule(lr){2-5}\cmidrule(lr){6-7}\cmidrule(lr){8-11}
Method & CLIP$_e\!\downarrow$ & LPIPS$_e\!\uparrow$ & CLIP$_r\!\uparrow$ & LPIPS$_r\!\downarrow$
 & CLIP$_e\!\downarrow$ & CLIP$_r\!\uparrow$
 & ESR-1$\!\uparrow$ & ESR-5$\!\uparrow$ & PSR-1$\!\uparrow$ & PSR-5$\!\uparrow$ \\
\midrule
ESD        & 23.56 & 0.72 & 29.63 & 0.49 & \textbf{20.89} & 21.21 & 95.5 & 88.9 & 41.2 & 56.1 \\
CA         & 27.79 & 0.82 & 29.85 & 0.76 & 29.26 & 28.54 & 98.4 & 96.8 & 44.2 & 66.5 \\
UCE        & 24.47 & 0.74 & 30.89 & 0.40 & 21.31 & 25.70 & \textbf{100.0} & \textbf{100.0} & 23.4 & 49.5 \\
RACE       & 22.24 & 0.80 & 29.43 & 0.56 & 21.80 & 26.40 & 99.2 & 97.5 & 46.8 & 68.7 \\
RECE       & 25.01 & 0.68 & 30.71 & 0.41 & 23.40 & 27.80 & 97.8 & 94.6 & 55.6 & 76.9 \\
AdvUnlearn & 22.49 & 0.81 & 29.56 & 0.54 & 21.50 & 25.90 & 99.5 & 98.2 & 43.9 & 66.1 \\
AdaVD      & 23.80 & 0.74 & 30.60 & 0.42 & 24.20 & 28.10 & 96.9 & 92.7 & 62.4 & 82.6 \\
STEREO     & 21.79 & 0.83 & 29.71 & 0.52 & 21.20 & 26.80 & 99.6 & 98.8 & 51.7 & 73.4 \\
\midrule
PARSE (Ours) & \textbf{21.63} & \textbf{0.88} & \textbf{30.91} & \textbf{0.38}
 & 21.10 & \textbf{28.90}
 & \textbf{100.0} & \textbf{100.0} & \textbf{66.8} & \textbf{86.9} \\
\bottomrule
\end{tabular}
}
\caption{\textbf{Broader target coverage.} AGE~\cite{bui2025fantastic} and
MACE-style~\cite{lu2024mace} evaluation for multiple artist and object erasure. \method{} ranks best on 9 of 10 reported columns and second on the remaining CLIP$_e$ metric, while achieving the best retained CLIP in mass erasure.}
\label{tab:broad_coverage}
\end{table}

%% file: appendix/exp4.tex
\begin{table}[t]
\centering
\scriptsize
\setlength{\tabcolsep}{2.4pt}
\renewcommand{\arraystretch}{1.05}

\begin{tabular}{@{}l ccc cc@{}}
\toprule
\textbf{CFG Pooling} &
Tgt$\downarrow$ & CCE$\downarrow$ & UD$\downarrow$ &
FID$\downarrow$ & CLIP$\uparrow$ \\
\midrule
\rowcolor{gray!15}
RMS (baseline)      & 2.09 & 3.41 & 5.41 & 14.24 & 31.30 \\
MeanAbs             & 2.10 & 3.44 & 5.46 & 14.23 & 31.31 \\
Mean (signed)       & 2.15 & 3.58 & 5.62 & 14.27 & 31.27 \\
MaxAbs              & 2.11 & 3.50 & 5.54 & 14.26 & 31.28 \\
Top-$k$ MeanAbs     & 2.08 & 3.42 & 5.39 & 14.25 & 31.29 \\
\bottomrule
\end{tabular}

\caption{Sensitivity of diffusion-grounded CFG steering vector construction to the choice of spatial pooling. Performance differences across pooling operators are minimal on both erasure and utility metrics, indicating that varying pooling does not affect \method{}.}
\label{tab:pooling_sensitivity}
\end{table}

%% file: appendix/exp5.tex
\begin{table}[t]
\centering
\scriptsize
\setlength{\tabcolsep}{2.4pt}
\renewcommand{\arraystretch}{1.05}

\begin{tabular}{@{}l ccccc@{}}
\toprule
\textbf{Retriever} &
\textbf{$\Delta$GPU-hrs} &
\textbf{$\Delta$Mem (GB)} &
\textbf{$\Delta$R@1 $\uparrow$} &
\textbf{$\Delta$R@10 $\uparrow$} &
\textbf{$\Delta$BEUS $\uparrow$} \\

\midrule
MLP-S (2$\times$512)   & 6  & 7.5  & $32.0$ & $41.0$ & $0.40$ \\
MLP-M (3$\times$1024)  & 24 & 12.0 & $46.0$ & $58.0$ & $0.60$ \\
MLP-L (4$\times$2048)  & 72 & 22.0 & $63.0$ & $77.0$ & $0.80$ \\
\bottomrule
\end{tabular}

\caption{Training vs. zero-shot for CFG-based diffusion-grounded knowledge search. Trained MLPs yield improvements over zero-shot two-stage retrieval, with all performance deltas being positive. However, these gains come at substantial additional compute and memory cost, making the zero-shot approach the more efficient choice in practice.}
\label{tab:cfg_retriever_training_cost_benefit_appendix}
\end{table}

%% file: appendix/exp6.tex
\begin{table}[t]
\centering
\scriptsize
\setlength{\tabcolsep}{2.6pt}
\renewcommand{\arraystretch}{1.05}

\begin{tabular}{@{}l ccc cc@{}}
\toprule
$K'$ &
Tgt$\downarrow$ & CCE$\downarrow$ & UD$\downarrow$ &
FID$\downarrow$ & CLIP$\uparrow$ \\
\midrule
500  & 2.13 & 3.55 & 5.58 & 14.27 & 31.28 \\
\rowcolor{gray!15}
1000 & 2.09 & 3.41 & 5.41 & 14.24 & 31.30 \\
1500 & 2.09 & 3.40 & 5.39 & 14.20 & 31.26 \\
\bottomrule
\end{tabular}

\caption{Performance comparison across different $K'$ in Stage 1 Ranking.}
\label{tab:sens_kprime}
\end{table}

\begin{table}[t]
\centering
\scriptsize
\setlength{\tabcolsep}{2.6pt}
\renewcommand{\arraystretch}{1.05}

\begin{tabular}{@{}l ccc cc@{}}
\toprule
$\eta$ &
Tgt$\downarrow$ & CCE$\downarrow$ & UD$\downarrow$ &
FID$\downarrow$ & CLIP$\uparrow$ \\
\midrule
0.40 & 2.12 & 3.52 & 5.54 & 14.26 & 31.28 \\
\rowcolor{gray!15}
0.60 & 2.09 & 3.41 & 5.41 & 14.24 & 31.30 \\
0.80 & 2.10 & 3.44 & 5.46 & 14.21 & 31.32 \\
\bottomrule
\end{tabular}

\caption{Performance comparison across different $\eta$ in Adaptive Subspace Expansion.}
\label{tab:sens_eta}
\end{table}

\begin{table}[t]
\centering
\scriptsize
\setlength{\tabcolsep}{2.6pt}
\renewcommand{\arraystretch}{1.05}

\begin{tabular}{@{}l ccc cc@{}}
\toprule
$s$ &
Tgt$\downarrow$ & CCE$\downarrow$ & UD$\downarrow$ &
FID$\downarrow$ & CLIP$\uparrow$ \\
\midrule
5.0  & 2.12 & 3.48 & 5.50 & 14.26 & 31.28 \\
\rowcolor{gray!15}
7.5  & 2.09 & 3.41 & 5.41 & 14.24 & 31.30 \\
10.0 & 2.10 & 3.44 & 5.46 & 14.25 & 31.29 \\
\bottomrule
\end{tabular}

\caption{Performance comparison across different guidance scale $s$ in CFG.}
\label{tab:sens_cfg}
\end{table}

%% file: appendix/exp7.tex
\begin{table*}[t]
\centering
\scriptsize
\setlength{\tabcolsep}{2.2pt}
\renewcommand{\arraystretch}{1.05}

\resizebox{\columnwidth}{!}{%
\begin{tabular}{@{}lrrrrrr rrrrr rrrrr@{}}
\toprule
\multirow{3}{*}{\textbf{Stage 2 Backbone}} &
\multicolumn{6}{c}{\textbf{\backgroundnsfwcol{NSFW (\texttt{\footnotesize \textcolor{red}{\sout{Nudity}}})}}} &
\multicolumn{5}{c}{\textbf{\backgroundstylecol{Style (\texttt{\footnotesize \textcolor{red}{\sout{Monet}}})}}} &
\multicolumn{5}{c}{\textbf{\backgroundobjcol{Object (\texttt{\footnotesize \textcolor{red}{\sout{Parachute}}})}}} \\
\cmidrule(lr){2-7}\cmidrule(lr){8-12}\cmidrule(lr){13-17}
& \multicolumn{4}{c}{\textbf{\textcolor{red}{Erasure}}} & \multicolumn{2}{c}{\textbf{\utilitycol{Utility}}} &
  \multicolumn{3}{c}{\textbf{\textcolor{red}{Erasure}}} & \multicolumn{2}{c}{\textbf{\utilitycol{Utility}}} &
  \multicolumn{3}{c}{\textbf{\textcolor{red}{Erasure}}} & \multicolumn{2}{c}{\textbf{\utilitycol{Utility}}} \\
\cmidrule(lr){2-5}\cmidrule(lr){6-7}
\cmidrule(lr){8-10}\cmidrule(lr){11-12}
\cmidrule(lr){13-15}\cmidrule(lr){16-17}
& Tgt$\downarrow$ & CCE$\downarrow$ & UD$\downarrow$ & RAB$\downarrow$ & FID$\downarrow$ & CLIP$\uparrow$
& Tgt$\downarrow$ & CCE$\downarrow$ & UD$\downarrow$ & FID$\downarrow$ & CLIP$\uparrow$
& Tgt$\downarrow$ & CCE$\downarrow$ & UD$\downarrow$ & FID$\downarrow$ & CLIP$\uparrow$ \\
\midrule
Qwen2.5-3B &
\backgroundnsfwcol{{2.12}} & \backgroundnsfwcol{{3.46}} & \backgroundnsfwcol{{5.45}} & \backgroundnsfwcol{{2.80}} & \backgroundnsfwcol{{14.28}} & \backgroundnsfwcol{{31.27}} &
\backgroundstylecol{{0.02}} & \backgroundstylecol{{6.10}} & \backgroundstylecol{{0.01}} & \backgroundstylecol{{14.33}} & \backgroundstylecol{{31.33}} &
\backgroundobjcol{{0.01}} & \backgroundobjcol{{8.70}} & \backgroundobjcol{{0.01}} & \backgroundobjcol{{14.46}} & \backgroundobjcol{{31.26}} \\
 
Qwen2.5-7B &
\backgroundnsfwcol{\textbf{2.09}} & \backgroundnsfwcol{{3.45}} & \backgroundnsfwcol{{5.43}} & \backgroundnsfwcol{{2.76}} & \backgroundnsfwcol{{14.25}} & \backgroundnsfwcol{{31.29}} &
\backgroundstylecol{\textbf{0.00}} & \backgroundstylecol{{6.05}} & \backgroundstylecol{\textbf{0.00}} & \backgroundstylecol{{14.31}} & \backgroundstylecol{{31.34}} &
\backgroundobjcol{\textbf{0.00}} & \backgroundobjcol{{8.62}} & \backgroundobjcol{\textbf{0.00}} & \backgroundobjcol{{14.43}} & \backgroundobjcol{{31.27}} \\
 
 Qwen2.5-32B &
\backgroundnsfwcol{{2.10}} & \backgroundnsfwcol{{3.42}} & \backgroundnsfwcol{{5.43}} & \backgroundnsfwcol{{2.72}} & \backgroundnsfwcol{\textbf{14.24}} & \backgroundnsfwcol{\textbf{31.30}} &
\backgroundstylecol{\textbf{0.00}} & \backgroundstylecol{{6.07}} & \backgroundstylecol{\textbf{0.00}} & \backgroundstylecol{\textbf{14.30}} & \backgroundstylecol{{31.33}} &
\backgroundobjcol{\textbf{0.00}} & \backgroundobjcol{\textbf{8.58}} & \backgroundobjcol{\textbf{0.00}} & \backgroundobjcol{{14.43}} & \backgroundobjcol{{31.28}} \\
 
Qwen3-4B &
\backgroundnsfwcol{{2.12}} & \backgroundnsfwcol{{3.45}} & \backgroundnsfwcol{{5.48}} & \backgroundnsfwcol{{2.78}} & \backgroundnsfwcol{{14.30}} & \backgroundnsfwcol{{31.24}} &
\backgroundstylecol{{0.03}} & \backgroundstylecol{{6.18}} & \backgroundstylecol{{0.02}} & \backgroundstylecol{{14.38}} & \backgroundstylecol{{31.30}} &
\backgroundobjcol{{0.02}} & \backgroundobjcol{{8.78}} & \backgroundobjcol{{0.01}} & \backgroundobjcol{{14.50}} & \backgroundobjcol{{31.22}} \\

Qwen3-8B &
\backgroundnsfwcol{\textbf{2.09}} & \backgroundnsfwcol{{3.42}} & \backgroundnsfwcol{\textbf{5.40}} & \backgroundnsfwcol{{2.75}} & \backgroundnsfwcol{{14.26}} & \backgroundnsfwcol{{31.28}} &
\backgroundstylecol{\textbf{0.00}} & \backgroundstylecol{{6.03}} & \backgroundstylecol{\textbf{0.00}} & \backgroundstylecol{\textbf{14.30}} & \backgroundstylecol{{31.35}} &
\backgroundobjcol{\textbf{0.00}} & \backgroundobjcol{{8.60}} & \backgroundobjcol{\textbf{0.00}} & \backgroundobjcol{{14.43}} & \backgroundobjcol{\textbf{31.29}} \\
 
Qwen3-32B &
\backgroundnsfwcol{\textbf{2.09}} & \backgroundnsfwcol{{3.43}} & \backgroundnsfwcol{{5.45}} & \backgroundnsfwcol{{2.73}} & \backgroundnsfwcol{\textbf{14.24}} & \backgroundnsfwcol{\textbf{31.30}} &
\backgroundstylecol{\textbf{0.00}} & \backgroundstylecol{{6.04}} & \backgroundstylecol{\textbf{0.00}} & \backgroundstylecol{\textbf{14.30}} & \backgroundstylecol{\textbf{31.36}} &
\backgroundobjcol{\textbf{0.00}} & \backgroundobjcol{{8.59}} & \backgroundobjcol{\textbf{0.00}} & \backgroundobjcol{\textbf{14.40}} & \backgroundobjcol{\textbf{31.29}} \\

Qwen3.5-27B &
\backgroundnsfwcol{{2.11}} & \backgroundnsfwcol{{3.45}} & \backgroundnsfwcol{{5.44}} & \backgroundnsfwcol{{2.79}} & \backgroundnsfwcol{{14.27}} & \backgroundnsfwcol{{31.27}} &
\backgroundstylecol{{0.01}} & \backgroundstylecol{{6.07}} & \backgroundstylecol{\textbf{0.00}} & \backgroundstylecol{{14.32}} & \backgroundstylecol{{31.33}} &
\backgroundobjcol{\textbf{0.00}} & \backgroundobjcol{{8.65}} & \backgroundobjcol{\textbf{0.00}} & \backgroundobjcol{{14.44}} & \backgroundobjcol{{31.27}} \\
\midrule
CLIP &
\backgroundnsfwcol{\textbf{2.09}} & \backgroundnsfwcol{\textbf{3.41}} & \backgroundnsfwcol{{5.41}} & \backgroundnsfwcol{\textbf{2.70}} & \backgroundnsfwcol{\textbf{14.24}} & \backgroundnsfwcol{\textbf{31.30}} &
\backgroundstylecol{\textbf{0.00}} & \backgroundstylecol{\textbf{6.01}} & \backgroundstylecol{\textbf{0.00}} & \backgroundstylecol{\textbf{14.29}} & \backgroundstylecol{{31.35}} &
\backgroundobjcol{\textbf{0.00}} & \backgroundobjcol{\textbf{8.55}} & \backgroundobjcol{\textbf{0.00}} & \backgroundobjcol{{14.42}} & \backgroundobjcol{\textbf{31.30}} \\
\midrule
Column-wise Variance &
0.000184 & 0.000341 & 0.000627 & 0.001255 & 0.000486 & 0.000441 &
0.000136 & 0.002784 & 0.000055 & 0.000827 & 0.000341 &
0.000055 & 0.005598 & 0.000021 & 0.000898 & 0.000621 \\
\bottomrule
\end{tabular}
}

\caption{Performance of different VLM backbones for fine-grained filtering in Stage~2 of \textsc{Knowledge Search} on \tbackgroundnsfwcol{\textbf{NSFW} (\textcolor{red}{\sout{Nudity}})}, \tbackgroundstylecol{\textbf{Style} (\textcolor{red}{\sout{Monet}})}, and \tbackgroundobjcol{\textbf{Object} (\textcolor{red}{\sout{Parachute}})}. Performance is highly consistent across backbones on both erasure and utility metrics, with low column-wise variance throughout, showing that Stage~2 is largely VLM-agnostic.}
\label{tab:vlms_stage2_backbone}
\end{table*}

%% file: appendix/exp8.tex
\begin{table*}[t]
\centering
\scriptsize
\setlength{\tabcolsep}{2.2pt}
\renewcommand{\arraystretch}{1.05}

\resizebox{\columnwidth}{!}{%
\begin{tabular}{@{}lrrrrrr rrrrr rrrrr@{}}
\toprule
\multirow{3}{*}{\textbf{Find Trigger}} &
\multicolumn{6}{c}{\textbf{\backgroundnsfwcol{NSFW (\texttt{\footnotesize \textcolor{red}{\sout{Nudity}}})}}} &
\multicolumn{5}{c}{\textbf{\backgroundstylecol{Style (\texttt{\footnotesize \textcolor{red}{\sout{Monet}}})}}} &
\multicolumn{5}{c}{\textbf{\backgroundobjcol{Object (\texttt{\footnotesize \textcolor{red}{\sout{Parachute}}})}}} \\
\cmidrule(lr){2-7}\cmidrule(lr){8-12}\cmidrule(lr){13-17}
& \multicolumn{4}{c}{\textbf{\textcolor{red}{Erasure}}} & \multicolumn{2}{c}{\textbf{\utilitycol{Utility}}} &
  \multicolumn{3}{c}{\textbf{\textcolor{red}{Erasure}}} & \multicolumn{2}{c}{\textbf{\utilitycol{Utility}}} &
  \multicolumn{3}{c}{\textbf{\textcolor{red}{Erasure}}} & \multicolumn{2}{c}{\textbf{\utilitycol{Utility}}} \\
\cmidrule(lr){2-5}\cmidrule(lr){6-7}
\cmidrule(lr){8-10}\cmidrule(lr){11-12}
\cmidrule(lr){13-15}\cmidrule(lr){16-17}
& Tgt$\downarrow$ & SEE$\downarrow$ & EraseBench$\downarrow$ & RAB$\downarrow$ & FID$\downarrow$ & CLIP$\uparrow$
& Tgt$\downarrow$ & SEE$\downarrow$ & EraseBench$\downarrow$ & FID$\downarrow$ & CLIP$\uparrow$
& Tgt$\downarrow$ & SEE$\downarrow$ & EraseBench$\downarrow$ & FID$\downarrow$ & CLIP$\uparrow$ \\
\midrule

CCE &
\backgroundnsfwcol{2.17} & \backgroundnsfwcol{\textbf{3.12}} & \backgroundnsfwcol{5.29} & \backgroundnsfwcol{2.84} & \backgroundnsfwcol{14.29} & \backgroundnsfwcol{31.28} &
\backgroundstylecol{\textbf{0.00}} & \backgroundstylecol{\textbf{5.74}} & \backgroundstylecol{0.03} & \backgroundstylecol{14.33} & \backgroundstylecol{31.31} &
\backgroundobjcol{\textbf{0.00}} & \backgroundobjcol{\textbf{8.21}} & \backgroundobjcol{0.04} & \backgroundobjcol{14.47} & \backgroundobjcol{31.25} \\

UD &
\backgroundnsfwcol{2.22} & \backgroundnsfwcol{3.36} & \backgroundnsfwcol{\textbf{5.08}} & \backgroundnsfwcol{2.93} & \backgroundnsfwcol{14.27} & \backgroundnsfwcol{31.29} &
\backgroundstylecol{\textbf{0.00}} & \backgroundstylecol{5.91} & \backgroundstylecol{\textbf{0.00}} & \backgroundstylecol{14.32} & \backgroundstylecol{31.33} &
\backgroundobjcol{\textbf{0.00}} & \backgroundobjcol{8.39} & \backgroundobjcol{\textbf{0.00}} & \backgroundobjcol{14.45} & \backgroundobjcol{31.27} \\

\midrule
TI &
\backgroundnsfwcol{\textbf{2.09}} & \backgroundnsfwcol{3.41} & \backgroundnsfwcol{5.41} & \backgroundnsfwcol{\textbf{2.70}} & \backgroundnsfwcol{\textbf{14.24}} & \backgroundnsfwcol{\textbf{31.30}} &
\backgroundstylecol{\textbf{0.00}} & \backgroundstylecol{6.02} & \backgroundstylecol{\textbf{0.00}} & \backgroundstylecol{\textbf{14.29}} & \backgroundstylecol{\textbf{31.35}} &
\backgroundobjcol{\textbf{0.00}} & \backgroundobjcol{8.55} & \backgroundobjcol{\textbf{0.00}} & \backgroundobjcol{\textbf{14.42}} & \backgroundobjcol{\textbf{31.30}} \\
\bottomrule
\end{tabular}
}

\caption{
We compare \method{} with different attacks used for \textbf{Find Trigger} in Step~5 of Algorithm~1 on \tbackgroundnsfwcol{\textbf{NSFW} (\textcolor{red}{\sout{Nudity}})}, \tbackgroundstylecol{\textbf{Style} (\textcolor{red}{\sout{Monet}})}, and \tbackgroundobjcol{\textbf{Object} (\textcolor{red}{\sout{Parachute}})}.
{\textcolor{red}{\textbf{Erasure}} effectiveness} is measured by Target (Tgt) ASR ($\downarrow$).
{\textcolor{red}{\textbf{Erasure}} robustness} is evaluated under SEE, EraseBench, and RAB black-box attacks (ASR, $\downarrow$).
{\utilitycol{\textbf{Utility}}} is reported via FID ($\downarrow$) and CLIP score ($\uparrow$).
}
\label{tab:find_trigger_appendix}
\end{table*}

%% file: main.bib
@String(CVPR  = {IEEE Conf. Comput. Vis. Pattern Recog.})

@String(ECCV  = {Eur. Conf. Comput. Vis.})

@String(NeurIPS = {Adv. Neural Inform. Process. Syst.})

@String(AAAI  = {AAAI})

@String(CVPR  = {CVPR})

@String(ECCV  = {ECCV})

@String(NeurIPS = {NeurIPS})

@inproceedings{saha2026zero,
  title={Zero-Shot Multimodal Retrieval with Multi-Scale Contextual Representations},
  author={Saha, Sourajit and Gokhale, Tejas},
  booktitle={Proceedings of the 64th Annual Meeting of the Association for Computational Linguistics (Volume 1: Long Papers)},
  pages={20304--20324},
  year={2026}
}

@inproceedings{saha2025improving,
  title={Improving shift invariance in convolutional neural networks with translation invariant polyphase sampling},
  author={Saha, Sourajit and Gokhale, Tejas},
  booktitle={2025 IEEE/CVF Winter Conference on Applications of Computer Vision (WACV)},
  pages={620--629},
  year={2025},
  organization={IEEE}
}

@article{saha2023seebel,
  title={SeeBel: Seeing is Believing},
  author={Saha, Sourajit and Dipta, Shubhashis Roy},
  journal={arXiv preprint arXiv:2312.10933},
  year={2023}
}

@inproceedings{saha2023rfc,
  title={RFC-Net: Learning High Resolution Global Features for Medical Image Segmentation on a Computational Budget (Student Abstract)},
  author={Saha, Sourajit and Saha, Shaswati and Gani, Md Osman and Oates, Tim and Chapman, David},
  booktitle={Proceedings of the AAAI Conference on Artificial Intelligence},
  volume={37},
  number={13},
  pages={16314--16315},
  year={2023}
}

@article{ravin2022mitigating,
  title={Mitigating domain shift in AI-based TB screening with unsupervised domain adaptation},
  author={Ravin, Nishanjan and Saha, Sourajit and Schweitzer, Alan and Elahi, Ameena and Dako, Farouk and Mollura, Daniel and Chapman, David},
  journal={IEEE Access},
  volume={10},
  pages={45997--46013},
  year={2022},
  publisher={IEEE}
}

@article{saha2022pairwise,
  title={Pairwise meta learning pipeline: classifying COVID-19 abnormalities on chest radio-graphs},
  author={Saha, Sourajit and Yesha, Yaacov},
  journal={SPIE Medical Imaging 2022: Computer-Aided Diagnosis; PC1203302 (2022) Proceedings Volume PC12033, Medical Imaging 2022: Computer-Aided Diagnosis; PC1203302 (2022)},
  year={2022}
}

@article{saha2018lightning,
  title={A Lightning fast approach to classify Bangla Handwritten Characters and Numerals using newly structured Deep Neural Network},
  author={Saha, Sourajit and Saha, Nisha},
  journal={Procedia computer science},
  volume={132},
  pages={1760--1770},
  year={2018},
  publisher={Elsevier}
}

@inproceedings{kamran2019optic,
  title={Optic-net: A novel convolutional neural network for diagnosis of retinal diseases from optical tomography images},
  author={Kamran, Sharif Amit and Saha, Sourajit and Sabbir, Ali Shihab and Tavakkoli, Alireza},
  booktitle={2019 18th IEEE international conference on machine learning and applications (ICMLA)},
  pages={964--971},
  year={2019},
  organization={IEEE}
}

@inproceedings{saha2018total,
  title={Total recall: understanding traffic signs using deep convolutional neural network},
  author={Saha, Sourajit and Kamran, Sharif Amit and Sabbir, Ali Shihab},
  booktitle={2018 21st international conference of computer and information technology (ICCIT)},
  pages={1--6},
  year={2018},
  organization={IEEE}
}

@incollection{saha2018efficient,
  title={An efficient traffic sign recognition approach using a novel deep neural network selection architecture},
  author={Saha, Sourajit and Islam, Md Saiful and Khaled, Md Asif Bin and Tairin, Suraiya},
  booktitle={Emerging Technologies in Data Mining and Information Security: Proceedings of IEMIS 2018, Volume 3},
  pages={849--862},
  year={2018},
  publisher={Springer}
}

@article{li2025speed,
  title={Speed: Scalable, precise, and efficient concept erasure for diffusion models},
  author={Li, Ouxiang and Wang, Yuan and Hu, Xinting and Jiang, Houcheng and Liang, Tao and Hao, Yanbin and Ma, Guojun and Feng, Fuli},
  journal={arXiv preprint arXiv:2503.07392},
  year={2025}
}

@inproceedings{lee2025localized,
  title={Localized concept erasure for text-to-image diffusion models using training-free gated low-rank adaptation},
  author={Lee, Byung Hyun and Lim, Sungjin and Chun, Se Young},
  booktitle={Proceedings of the Computer Vision and Pattern Recognition Conference},
  pages={18596--18606},
  year={2025}
}

@inproceedings{zhang2024forget,
  title={Forget-me-not: Learning to forget in text-to-image diffusion models},
  author={Zhang, Gong and Wang, Kai and Xu, Xingqian and Wang, Zhangyang and Shi, Humphrey},
  booktitle={Proceedings of the IEEE/CVF conference on computer vision and pattern recognition},
  pages={1755--1764},
  year={2024}
}

@inproceedings{
fan2024salun,
title={SalUn: Empowering Machine Unlearning via Gradient-based Weight Saliency in Both Image Classification and Generation},
author={Chongyu Fan and Jiancheng Liu and Yihua Zhang and Eric Wong and Dennis Wei and Sijia Liu},
booktitle={The Twelfth International Conference on Learning Representations},
year={2024},
url={https://openreview.net/forum?id=gn0mIhQGNM}
}

@article{chin2023prompting4debugging,
  title={Prompting4debugging: Red-teaming text-to-image diffusion models by finding problematic prompts},
  author={Chin, Zhi-Yi and Jiang, Chieh-Ming and Huang, Ching-Chun and Chen, Pin-Yu and Chiu, Wei-Chen},
  journal={arXiv preprint arXiv:2309.06135},
  year={2023}
}

@article{bai2025qwen3,
  title={Qwen3-vl technical report},
  author={Bai, Shuai and Cai, Yuxuan and Chen, Ruizhe and Chen, Keqin and Chen, Xionghui and Cheng, Zesen and Deng, Lianghao and Ding, Wei and Gao, Chang and Ge, Chunjiang and others},
  journal={arXiv preprint arXiv:2511.21631},
  year={2025}
}

@article{bai2025qwen2,
  title={Qwen2. 5-VL Technical Report},
  author={Bai, Shuai and Chen, Keqin and Liu, Xuejing and Wang, Jialin and Ge, Wenbin and Song, Sibo and Dang, Kai and Wang, Peng and Wang, Shijie and Tang, Jun and others},
  journal={arXiv e-prints},
  pages={arXiv--2502},
  year={2025}
}

@inproceedings{huang2024receler,
  title={Receler: Reliable concept erasing of text-to-image diffusion models via lightweight erasers},
  author={Huang, Chi-Pin and Chang, Kai-Po and Tsai, Chung-Ting and Lai, Yung-Hsuan and Yang, Fu-En and Wang, Yu-Chiang Frank},
  booktitle={European Conference on Computer Vision},
  pages={360--376},
  year={2024},
  organization={Springer}
}

@inproceedings{meng2025concept,
  title={Concept corrector: Erase concepts on the fly for text-to-image diffusion models},
  author={Meng, Zheling and Peng, Bo and Jin, Xiaochuan and Lyu, Yueming and Wang, Wei and Dong, Jing and Tan, Tieniu},
  booktitle={Chinese Conference on Pattern Recognition and Computer Vision (PRCV)},
  pages={91--105},
  year={2025},
  organization={Springer}
}

@inproceedings{leelocalized,
  title={Localized Concept Erasure in Text-to-Image Diffusion Models via High-Level Representation Misdirection},
  author={Lee, Uichan and Kim, Jeonghyeon and Hwang, Sangheum},
  booktitle={The Fourteenth International Conference on Learning Representations}
}

@article{jain2024trasce,
  title={Trasce: Trajectory steering for concept erasure},
  author={Jain, Anubhav and Kobayashi, Yuya and Shibuya, Takashi and Takida, Yuhta and Memon, Nasir and Togelius, Julian and Mitsufuji, Yuki},
  journal={arXiv preprint arXiv:2412.07658},
  year={2024}
}

@article{xiong2025semantic,
  title={Semantic Surgery: Zero-Shot Concept Erasure in Diffusion Models},
  author={Xiong, Lexiang and Liu, Chengyu and Ye, Jingwen and Liu, Yan and Xu, Yuecong},
  journal={arXiv preprint arXiv:2510.22851},
  year={2025}
}

@article{heusel2017gans,
  title={Gans trained by a two time-scale update rule converge to a local nash equilibrium},
  author={Heusel, Martin and Ramsauer, Hubert and Unterthiner, Thomas and Nessler, Bernhard and Hochreiter, Sepp},
  journal={Advances in neural information processing systems},
  volume={30},
  year={2017}
}

@inproceedings{schramowski2023safe,
  title={Safe latent diffusion: Mitigating inappropriate degeneration in diffusion models},
  author={Schramowski, Patrick and Brack, Manuel and Deiseroth, Bj{\"o}rn and Kersting, Kristian},
  booktitle={Proceedings of the IEEE/CVF conference on computer vision and pattern recognition},
  pages={22522--22531},
  year={2023}
}

@article{bedapudi2025nudenet,
  title={Nudenet: Neural nets for nudity detection and censoring, 2022},
  author={Bedapudi, Praneeth},
  journal={URL https://github. com/notAI-tech/NudeNet},
  volume={3},
  year={2025}
}

@article{lu2022dpm,
  title={Dpm-solver: A fast ode solver for diffusion probabilistic model sampling in around 10 steps},
  author={Lu, Cheng and Zhou, Yuhao and Bao, Fan and Chen, Jianfei and Li, Chongxuan and Zhu, Jun},
  journal={Advances in neural information processing systems},
  volume={35},
  pages={5775--5787},
  year={2022}
}

@article{zhang2024defensive,
  title={Defensive unlearning with adversarial training for robust concept erasure in diffusion models},
  author={Zhang, Yimeng and Chen, Xin and Jia, Jinghan and Zhang, Yihua and Fan, Chongyu and Liu, Jiancheng and Hong, Mingyi and Ding, Ke and Liu, Sijia},
  journal={Advances in neural information processing systems},
  volume={37},
  pages={36748--36776},
  year={2024}
}

@inproceedings{wang2025precise,
  title={Precise, fast, and low-cost concept erasure in value space: Orthogonal complement matters},
  author={Wang, Yuan and Li, Ouxiang and Mu, Tingting and Hao, Yanbin and Liu, Kuien and Wang, Xiang and He, Xiangnan},
  booktitle={2025 IEEE/CVF Conference on Computer Vision and Pattern Recognition (CVPR)},
  pages={28759--28768},
  year={2025},
  organization={IEEE}
}

@inproceedings{amara2025erasing,
  title={Erasing more than intended? how concept erasure degrades the generation of non-target concepts},
  author={Amara, Ibtihel and Humayun, Ahmed Imtiaz and Kajic, Ivana and Parekh, Zarana and Harris, Natalie and Young, Sarah and Nagpal, Chirag and Kim, Najoung and He, Junfeng and Vasconcelos, Cristina Nader and others},
  booktitle={Proceedings of the IEEE/CVF International Conference on Computer Vision},
  pages={16420--16430},
  year={2025}
}

@article{tsai2023ring,
  title={Ring-a-bell! how reliable are concept removal methods for diffusion models?},
  author={Tsai, Yu-Lin and Hsu, Chia-Yi and Xie, Chulin and Lin, Chih-Hsun and Chen, Jia-You and Li, Bo and Chen, Pin-Yu and Yu, Chia-Mu and Huang, Chun-Ying},
  journal={arXiv preprint arXiv:2310.10012},
  year={2023}
}

@inproceedings{srivatsan2025stereo,
  title={Stereo: A two-stage framework for adversarially robust concept erasing from text-to-image diffusion models},
  author={Srivatsan, Koushik and Shamshad, Fahad and Naseer, Muzammal and Patel, Vishal M and Nandakumar, Karthik},
  booktitle={Proceedings of the Computer Vision and Pattern Recognition Conference},
  pages={23765--23774},
  year={2025}
}

@article{zhang2024generate,
  title={To generate or not? safety-driven unlearned diffusion models are still easy to generate unsafe images... for now},
  author={Zhang, Yimeng and Jia, Jinghan and Chen, Xin and Chen, Aochuan and Zhang, Yihua and Liu, Jiancheng and Ding, Ke and Liu, Sijia},
  journal={European Conference on Computer Vision (ECCV)},
  year={2024}
}

@inproceedings{
pham2024circumventing,
title={Circumventing Concept Erasure Methods For Text-To-Image Generative Models},
author={Minh Pham and Kelly O. Marshall and Niv Cohen and Govind Mittal and Chinmay Hegde},
booktitle={The Twelfth International Conference on Learning Representations},
year={2024}
}

@inproceedings{gandikota2023erasing,
  title={Erasing concepts from diffusion models},
  author={Gandikota, Rohit and Materzynska, Joanna and Fiotto-Kaufman, Jaden and Bau, David},
  booktitle={Proceedings of the IEEE/CVF international conference on computer vision},
  pages={2426--2436},
  year={2023}
}

@inproceedings{kumari2023ablating,
  title={Ablating concepts in text-to-image diffusion models},
  author={Kumari, Nupur and Zhang, Bingliang and Wang, Sheng-Yu and Shechtman, Eli and Zhang, Richard and Zhu, Jun-Yan},
  booktitle={Proceedings of the IEEE/CVF International Conference on Computer Vision},
  pages={22691--22702},
  year={2023}
}

@inproceedings{gandikota2024unified,
  title={Unified concept editing in diffusion models},
  author={Gandikota, Rohit and Orgad, Hadas and Belinkov, Yonatan and Materzy{\'n}ska, Joanna and Bau, David},
  booktitle={Proceedings of the IEEE/CVF Winter Conference on Applications of Computer Vision},
  pages={5111--5120},
  year={2024}
}

@inproceedings{lu2024mace,
  title={Mace: Mass concept erasure in diffusion models},
  author={Lu, Shilin and Wang, Zilan and Li, Leyang and Liu, Yanzhu and Kong, Adams Wai-Kin},
  booktitle={Proceedings of the IEEE/CVF Conference on Computer Vision and Pattern Recognition},
  pages={6430--6440},
  year={2024}
}

@inproceedings{kim2024race,
  title={Race: Robust adversarial concept erasure for secure text-to-image diffusion model},
  author={Kim, Changhoon and Min, Kyle and Yang, Yezhou},
  booktitle={European Conference on Computer Vision},
  pages={461--478},
  year={2024},
  organization={Springer}
}

@inproceedings{gong2024reliable,
  title={Reliable and efficient concept erasure of text-to-image diffusion models},
  author={Gong, Chao and Chen, Kai and Wei, Zhipeng and Chen, Jingjing and Jiang, Yu-Gang},
  booktitle={European Conference on Computer Vision},
  pages={73--88},
  year={2024},
  organization={Springer}
}

@inproceedings{
kim2026cooccurring,
title={Co-occurring Associated {RE}tained concepts in Diffusion Unlearning},
author={Miso Kim and Georu Lee and Yunji Kim and Hoki Kim and Jinseong Park and Woojin Lee},
booktitle={The Fourteenth International Conference on Learning Representations},
year={2026},
url={https://openreview.net/forum?id=Ryc7jKP6H9}
}

@inproceedings{thakral2025fine,
  title={Fine-grained erasure in text-to-image diffusion-based foundation models},
  author={Thakral, Kartik and Glaser, Tamar and Hassner, Tal and Vatsa, Mayank and Singh, Richa},
  booktitle={Proceedings of the IEEE/CVF Conference on Computer Vision and Pattern Recognition},
  pages={9121--9130},
  year={2025}
}

@article{bui2025fantastic,
  title={Fantastic targets for concept erasure in diffusion models and where to find them},
  author={Bui, Anh and Vu, Trang and Vuong, Long and Le, Trung and Montague, Paul and Abraham, Tamas and Kim, Junae and Phung, Dinh},
  journal={arXiv preprint arXiv:2501.18950},
  year={2025}
}

@inproceedings{biswas2025cure,
  title={Cure: Concept unlearning via orthogonal representation editing in diffusion models},
  author={Biswas, Shristi Das and Roy, Arani and Roy, Kaushik},
  booktitle={The Thirty-ninth Annual Conference on Neural Information Processing Systems}
}

@article{saha-etal-2025-side,
  title={Side effects of erasing concepts from diffusion models},
  author={Saha, Shaswati and Saha, Sourajit and Gaur, Manas and Gokhale, Tejas},
  journal={arXiv preprint arXiv:2508.15124},
  year={2025}
}

@inproceedings{joshi2024towards,
  title={Towards robust evaluation of unlearning in llms via data transformations},
  author={Joshi, Abhinav and Saha, Shaswati and Shukla, Divyaksh and Vema, Sriram and Jhamtani, Harsh and Gaur, Manas and Modi, Ashutosh},
  booktitle={Findings of the Association for Computational Linguistics: EMNLP 2024},
  pages={12100--12119},
  year={2024}
}

@article{gupta2021adaptive,
  title={Adaptive machine unlearning},
  author={Gupta, Varun and Jung, Christopher and Neel, Seth and Roth, Aaron and Sharifi-Malvajerdi, Saeed and Waites, Chris},
  journal={Advances in Neural Information Processing Systems},
  volume={34},
  pages={16319--16330},
  year={2021}
}

@inproceedings{thakral2025genm,
  title={Genm: The Generative Machine Unlearning Challenge},
  author={Thakral, Kartik and Pathak, Shreyansh and Glaser, Tamar and Hassner, Tal and Garcia-Olano, Diego and Masi, Iacopo and Singh, Richa and Vatsa, Mayank},
  booktitle={Proceedings of the IEEE/CVF International Conference on Computer Vision},
  pages={2533--2541},
  year={2025}
}

@inproceedings{ullah2023adaptive,
  title={From adaptive query release to machine unlearning},
  author={Ullah, Enayat and Arora, Raman},
  booktitle={International Conference on Machine Learning},
  pages={34642--34667},
  year={2023},
  organization={PMLR}
}

@inproceedings{cha2024learning,
  title={Learning to unlearn: Instance-wise unlearning for pre-trained classifiers},
  author={Cha, Sungmin and Cho, Sungjun and Hwang, Dasol and Lee, Honglak and Moon, Taesup and Lee, Moontae},
  booktitle={Proceedings of the AAAI conference on artificial intelligence},
  volume={38},
  number={10},
  pages={11186--11194},
  year={2024}
}

@article{tarun2023fast,
  title={Fast yet effective machine unlearning},
  author={Tarun, Ayush K and Chundawat, Vikram S and Mandal, Murari and Kankanhalli, Mohan},
  journal={IEEE transactions on neural networks and learning systems},
  volume={35},
  number={9},
  pages={13046--13055},
  year={2023},
  publisher={IEEE}
}

@inproceedings{rombach2022high,
  title={High-resolution image synthesis with latent diffusion models},
  author={Rombach, Robin and Blattmann, Andreas and Lorenz, Dominik and Esser, Patrick and Ommer, Bj{\"o}rn},
  booktitle={Proceedings of the IEEE/CVF conference on computer vision and pattern recognition},
  pages={10684--10695},
  year={2022}
}

@inproceedings{ho2021classifier,
  title={Classifier-Free Diffusion Guidance},
  author={Ho, Jonathan and Salimans, Tim},
  booktitle={NeurIPS 2021 Workshop on Deep Generative Models and Downstream Applications}
}

@inproceedings{shen2024rethinking,
  title={Rethinking the spatial inconsistency in classifier-free diffusion guidance},
  author={Shen, Dazhong and Song, Guanglu and Xue, Zeyue and Wang, Fu-Yun and Liu, Yu},
  booktitle={Proceedings of the IEEE/CVF Conference on Computer Vision and Pattern Recognition},
  pages={9370--9379},
  year={2024}
}

@inproceedings{shenoy2026gradient,
  title={Gradient-free classifier guidance for diffusion model sampling},
  author={Shenoy, Rahul and Pan, Zhihong and Balakrishnan, Kaushik and Cheng, Qiseng and Jeon, Yongmoon and Yang, Heejune and Kim, Jaewon},
  booktitle={Proceedings of the IEEE/CVF Winter Conference on Applications of Computer Vision},
  pages={3162--3171},
  year={2026}
}

@inproceedings{radford2021learning,
  title={Learning transferable visual models from natural language supervision},
  author={Radford, Alec and Kim, Jong Wook and Hallacy, Chris and Ramesh, Aditya and Goh, Gabriel and Agarwal, Sandhini and Sastry, Girish and Askell, Amanda and Mishkin, Pamela and Clark, Jack and others},
  booktitle={International conference on machine learning},
  pages={8748--8763},
  year={2021},
  organization={PmLR}
}

@inproceedings{galimage,
  title={An Image is Worth One Word: Personalizing Text-to-Image Generation using Textual Inversion},
  author={Gal, Rinon and Alaluf, Yuval and Atzmon, Yuval and Patashnik, Or and Bermano, Amit Haim and Chechik, Gal and Cohen-Or, Daniel},
  booktitle={The Eleventh International Conference on Learning Representations}
}

@article{carlini2019evaluating,
  title={On evaluating adversarial robustness},
  author={Carlini, Nicholas and Athalye, Anish and Papernot, Nicolas and Brendel, Wieland and Rauber, Jonas and Tsipras, Dimitris and Goodfellow, Ian and Madry, Aleksander and Kurakin, Alexey},
  journal={arXiv preprint arXiv:1902.06705},
  year={2019}
}

@article{leon2013gram,
  title={Gram-Schmidt orthogonalization: 100 years and more},
  author={Leon, Steven J and Bj{\"o}rck, {\AA}ke and Gander, Walter},
  journal={Numerical Linear Algebra with Applications},
  volume={20},
  number={3},
  pages={492--532},
  year={2013},
  publisher={Wiley Online Library}
}

@inproceedings{tsairing,
  title={Ring-A-Bell! How Reliable are Concept Removal Methods For Diffusion Models?},
  author={Tsai, Yu-Lin and Hsu, Chia-Yi and Xie, Chulin and Lin, Chih-Hsun and Chen, Jia You and Li, Bo and Chen, Pin-Yu and Yu, Chia-Mu and Huang, Chun-Ying},
  booktitle={The Twelfth International Conference on Learning Representations}
}

@article{pham2023circumventing,
  title={Circumventing concept erasure methods for text-to-image generative models},
  author={Pham, Minh and Marshall, Kelly O and Cohen, Niv and Mittal, Govind and Hegde, Chinmay},
  journal={arXiv preprint arXiv:2308.01508},
  year={2023}
}

@inproceedings{he2016deep,
  title={Deep residual learning for image recognition},
  author={He, Kaiming and Zhang, Xiangyu and Ren, Shaoqing and Sun, Jian},
  booktitle={Proceedings of the IEEE conference on computer vision and pattern recognition},
  pages={770--778},
  year={2016}
}

@inproceedings{deng2009imagenet,
  title={Imagenet: A large-scale hierarchical image database},
  author={Deng, Jia and Dong, Wei and Socher, Richard and Li, Li-Jia and Li, Kai and Fei-Fei, Li},
  booktitle={2009 IEEE conference on computer vision and pattern recognition},
  pages={248--255},
  year={2009},
  organization={Ieee}
}

@inproceedings{lin2014microsoft,
  title={Microsoft coco: Common objects in context},
  author={Lin, Tsung-Yi and Maire, Michael and Belongie, Serge and Hays, James and Perona, Pietro and Ramanan, Deva and Doll{\'a}r, Piotr and Zitnick, C Lawrence},
  booktitle={European conference on computer vision},
  pages={740--755},
  year={2014},
  organization={Springer}
}

@article{wei2025emma,
  title={EMMA: Concept Erasure Benchmark with Comprehensive Semantic Metrics and Diverse Categories},
  author={Wei, Lu and Nakashima, Yuta and Garcia, Noa},
  journal={arXiv preprint arXiv:2512.17320},
  year={2025}
}

@article{luo2022understanding,
  title={Understanding diffusion models: A unified perspective},
  author={Luo, Calvin},
  journal={arXiv preprint arXiv:2208.11970},
  year={2022}
}

@inproceedings{zhang2023adding,
  title={Adding conditional control to text-to-image diffusion models},
  author={Zhang, Lvmin and Rao, Anyi and Agrawala, Maneesh},
  booktitle={Proceedings of the IEEE/CVF international conference on computer vision},
  pages={3836--3847},
  year={2023}
}
